%% file: main.tex
\documentclass[runningheads]{llncs}

\usepackage{eccv}

\usepackage{xcolor}
\usepackage{subcaption} % 加载 subcaption 宏包
\usepackage{amsmath}
\usepackage{graphicx}
\usepackage{amsmath}
\usepackage{amssymb}
\usepackage{booktabs}
\usepackage{footnote}
\usepackage{url}
\usepackage[flushleft]{threeparttable}
\usepackage{enumitem}
\usepackage{algorithm}
\usepackage{algorithmic}
\usepackage{listings}
\usepackage{tabularx}
\usepackage{pifont} 
\usepackage{bm}
\usepackage{tablefootnote}
\usepackage{balance}
\usepackage{colortbl}
\usepackage{multirow}
\usepackage{multicol}
\definecolor{lightgray}{gray}{0.9}
\definecolor{linecolor}{rgb}{0.82, 0.94, 0.75}

\definecolor{evaunit01green}{RGB}{82,208,83}
\definecolor{lowred}{RGB}{238,18,137}

\definecolor{lowerred}{RGB}{255,110,180}

\newcommand{\dplus}[1]{\fontsize{6pt}{0.1em}\selectfont (\textbf{\textcolor{lowred}{#1}})}
\newcommand{\ddplus}[1]{\fontsize{6pt}{0.1em}\selectfont (\textbf{\textcolor{lowerred}{#1}})}

\definecolor{defaultcolor}{RGB}{12,127,17}

\newcommand{\tp}{TP}

\definecolor{red}{RGB}{255,0,0}
\definecolor{blue}{RGB}{0,0,255}

\newcommand{\blk}[1]{}

\input{preamble}

\usepackage{eccvabbrv}

\usepackage{graphicx}
\usepackage{booktabs}

\usepackage[accsupp]{axessibility}  % Improves PDF readability for those with disabilities.

\usepackage{hyperref}

\usepackage{orcidlink}

\begin{document}

\title{Point Ladder Tuning: Parameter-Efficient Hierarchical Adaptation for 3D Point Cloud Understanding} 

\titlerunning{Point Ladder Tuning}

\author{Junlin Chang\inst{1,2}\orcidlink{0009-0004-0742-3003} \and
Longhao Zou\inst{2}\orcidlink{0000-0002-3477-7438} \and
Rui Li\inst{2}$^{\dagger}$\orcidlink{0000-0001-5332-2167}}

\authorrunning{J. Chang et al.}

\institute{Beihang University, Beijing \and
Pengcheng Laboratory, Shenzhen\\
\email{changjunlin@buaa.edu.cn} \email{\{zoulh,lir01\}@pcl.ac.cn}}

\maketitle

\begingroup
\renewcommand{\thefootnote}{}
\footnotetext{${\dagger}$ Corresponding author.}
\endgroup

\input{sec/0_abstract}
\input{sec/1_intro}

\input{sec/2_relatedwork}
\input{sec/3_method}

\input{sec/4_experiment}
\input{sec/5_conclusion}
% \clearpage  % TODO FINAL: This \clearpage needs to be removed from both review and camera-ready versions.

\section*{Acknowledgements}
% Please insert your acknowledgments here.
This work is supported by  Mobile Information Networks-National Science and Technology Major Project under Grant No.2025ZD1303200.

\clearpage
\appendix
\begin{center}
    {\Large\bfseries Supplementary Material}
\end{center}
\vspace{1.0em}
\input{sec/X_suppl}

\bibliographystyle{splncs04}
\bibliography{main}
\end{document}

%% file: preamble.tex
\definecolor{rowunit}{RGB}{128,128,255}
\definecolor{evaunit01green}{RGB}{54,125,189}
\newcommand{\evagreen}[1]{\textcolor{evaunit01green}{#1}}
\newcommand{\dtplus}[1]{\fontsize{6pt}{0.1em}\selectfont (\textbf{\evagreen{#1}})}

\usepackage{tabularx}

%% file: sec/0_abstract.tex
\begin{abstract}
Fine-tuning pre-trained point-cloud backbones typically updates all parameters, resulting in substantial computation and memory overhead. More importantly, modern point backbones rely on aggressive tokenization and downsampling, which yields compact global tokens but irreversibly discards fine-grained local geometry, an inherent bottleneck for parameter-efficient adaptation. Consequently, existing PEFT methods that operate only on these coarsened tokens can modulate global semantics but struggle to recover the missing multi-scale locality.
We present Point Ladder Tuning (PLT), a locality-aware PEFT framework that performs hierarchical, instance-conditioned adaptation while keeping the backbone frozen. PLT forms a lightweight closed loop: (i) a Hierarchical Ladder Network (HLN) constructs a multi-resolution local feature pyramid directly from raw points; (ii) a Local–Global Fusion (LGF) aligns and fuses local pyramids with intermediate backbone semantics; and (iii) a Dynamic Prompt Generator produces instance-aware multi-scale prompts to modulate the frozen backbone effectively. For dense prediction, we further introduce a lightweight segmentation head that progressively upsamples fused features and leverages backbone priors to refine fine structures.
Extensive experiments on classification and dense prediction show that PLT consistently surpasses prior PEFT baselines with minimal tunable parameters. PLT achieves state-of-the-art performance using only 2.71\% trainable parameters for classification and 7.69\% for dense prediction, and scales favorably to larger backbones, requiring merely 0.36\% parameters on PointGPT-L. The code is released at \url{https://github.com/JunLinChang/ECCV2026-PLT}.

\keywords{3D Point Cloud Understanding \and Parameter-Efficient Fine-Tuning}

\end{abstract}

%% file: sec/1_intro.tex
\section{Introduction}
\label{sec:intro}

The rapid advancement of 3D sensing technologies, such as Light Detection and Ranging (LiDAR), depth cameras (structured light/time-of-flight), photogrammetry, and stereo vision, has led to the widespread availability of point cloud data, driving the development of learning-based 3D imaging and point cloud understanding methods. 
Applications span a variety of domains, including autonomous driving~\cite{yang2024visual,song2024graphbev,chen20203d, liu2026pointtpa}, virtual/augmented reality~\cite{casado2023rendering,garrido2021point}, and robotics~\cite{wang2021trajectory,chen2022direct,christen2023learning}. 
Compared to grid-structured images, point clouds are sparse and unordered, exhibit non-uniform sampling and varying densities, and lack explicit geometric connectivity. These properties demand representations that simultaneously preserve fine-grained local geometry and maintain coherent global context, making naive extensions of 2D architectures insufficient in practice.

To address these challenges, the community has developed point-specific architectures~\cite{qi2017pointnet,qi2017pointnet++,li2018pointcnn,qian2022pointnext,wang2019dynamic,wu2024point} and increasingly relies on self-supervised pre-training~\cite{devlin2018bert,brown2020language,he2020momentum,chen2020improved,pang2022masked,yu2022point,zhang2022point,afham2022crosspoint} to learn transferable 3D features.
However, deploying such pre-trained backbones typically requires full fine-tuning, i.e., updating all parameters for every downstream task.
This paradigm is often impractical due to: (i) overfitting and catastrophic forgetting, (ii) storage overhead from maintaining task-specific model copies, and (iii) substantial compute and memory cost.

\input{tab/motivation}

Parameter-Efficient Fine-Tuning (PEFT) mitigates these issues by freezing pre-trained weights and optimizing only a small set of additional parameters, such as adapters~\cite{houlsby2019parameter}, prompt tuning~\cite{li2021prefix}, and ladder-style tuning~\cite{sung2022lst}. 
While highly effective in NLP and vision, directly applying these techniques to point-cloud backbones often yields sub-optimal results. 
As shown in Tab.~\ref{tab:locality},modern point cloud pretrained backbones rely on aggressive tokenization and downsampling of raw points (e.g., $2048 \rightarrow 128$), which preserve global semantics but inevitably loss fine-grained, multi-scale local geometry. 
Since most PEFT methods modulate only these coarsened tokens, they can adapt global semantics but have limited capacity to recover the missing locality that is critical for recognition and dense prediction. 
A natural idea is to introduce a local branch that extracts geometric details from raw points. 
However, local cues alone are typically insufficient for complex 3D understanding because they lack high-level semantic context. 
Without effectively integrating local geometry with the backbone's global priors, the additional branch brings limited gains.

To this end, we propose \textbf{Point Ladder Tuning (PLT)}, a novel locality-aware PEFT framework specifically designed for point cloud learning. Specially, We decouple the adaptation process into two tightly coupled pathways. First, a lightweight \textbf{Hierarchical Ladder Network (HLN)} constructs a multi-resolution pyramid of geometric features directly from raw points, preserving spatial granularity across scales. In parallel, a dynamic-prompt-based global adaptation pathway converts the constructed multi-scale cues into instance-aware prompts and injects them back into the frozen backbone, enabling task-aware refinement without full retraining. 
These two pathways are coupled through a \textbf{Local--Global Fusion (LGF)} module: LGF selectively fuses the fine-grained local cues extracted by HLN with intermediate pretrained representations from the backbone, and then feeds back the resulting multi-scale signals to modulate self-attention and global feature learning inside the frozen model.

In summary, this work has the following contributions:
\begin{itemize}
	\item We propose Point Ladder Tuning (PLT), a dual-pathway collaborative PEFT framework that adapts a frozen pretrained backbone through a construct-fuse-feedback loop.
	\item We design a lightweight Hierarchical Ladder Network (HLN) and a Local-Global Fusion (LGF) module to construct multi-resolution local geometric features, selectively integrate them with intermediate pretrained semantics, and feed the fused cues back into the backbone for more effective adaptation.
    \item We conduct extensive experiments on multiple point-cloud classification and segmentation benchmarks, demonstrating that PLT achieves competitive performance with only a small number of trainable parameters.
\end{itemize}

%% file: tab/motivation.tex
\begin{table}[t]
\centering
\caption{Locality preservation after tokenization. Cov@2r measures raw-point coverage by FPS token centers within 2r-radius neighborhoods, where r denotes the average nearest-neighbor distance among raw points. Boundary edge collapse is the fraction of cross-label raw KNN edges collapsed into the same token region. Token-Adj. R@k measures whether raw KNN edges remain connected within a token-center KNN graph, where $k$ is the number of nearest token neighbors.}
\label{tab:locality}
\scriptsize
\resizebox{\linewidth}{!}{
\begin{tabular}{lcccc}
\toprule
Dataset & Cov@2r (\%) & Boundary Edge Coll. (\%) & Token-Adj. R@1 (\%) & Token-Adj. R@8 (\%) \\
\midrule
ScanObjectNN & 22.64 & N/A & 58.96 & 99.31 \\
ShapeNetPart & 28.12 & 34.93 & 57.71 & 98.66 \\
S3DIS & 21.26 & 36.72 & 59.06 & 99.67 \\
\bottomrule
\end{tabular}}
\end{table}

%% file: sec/2_relatedwork.tex
\section{Related Work}
\label{sec:relatedwork}

% In this section, we review representative methods in both areas: deep learning techniques for 3D point cloud processing and the evolution of PEFT strategies across domains.

\subsection{Self-Supervised Learning on Point Cloud}

% Deep learning on point clouds has evolved significantly, with various architectures designed to capture the unique characteristics of 3D data. Early methods like PointNet~\cite{qi2017pointnet} and PointNet++~\cite{qi2017pointnet++} introduced novel approaches for processing unordered point sets, employing symmetric functions to aggregate features. These methods laid the groundwork for subsequent architectures, including DGCNN~\cite{wang2019dynamic}, which incorporated edge features and local connectivity, and Point Transformer~\cite{zhao2021point}, which adapted transformer mechanisms for point cloud processing.

Recent self-supervised pre-training has advanced point-cloud understanding by leveraging large unlabeled 3D data before task-specific fine-tuning. Current approaches mainly fall into three categories: contrastive learning, reconstruction-based learning, and hybrid methods combining both.
In contrastive learning, models like PointContrast~\cite{xie2020pointcontrast} and CrossPoint~\cite{afham2022crosspoint} learn view-invariant semantic features by contrasting augmented views of the same cloud. 
Reconstruction-based methods, such as PointBERT~\cite{yu2022point} and PointMAE~\cite{pang2022masked}, draw inspiration from NLP and vision models by employing masked prediction and autoencoding frameworks. PointGPT~\cite{chen2023pointgpt} scales this paradigm into a larger-scale generative foundation model, demonstrating strong generalization to diverse downstream tasks.
To address the scarcity of labeled 3D data, ACT~\cite{dong2022autoencoders} leverages a cross-modal teacher-student framework that transfers semantic priors from a stronger teacher to the 3D learner.

Traditionally, these 3D pre-trained models are fine-tuned for downstream tasks through full parameter updates. However, full fine-tuning can be inefficient, often resulting in the degradation of valuable knowledge gained during pre-training and increasing the risk of catastrophic forgetting. Therefore, this paper investigates more efficient and effective strategies for transferring 3D pre-trained models to downstream tasks while preserving the benefits of pre-training.

\subsection{PEFT for Point Cloud}
Fine-tuning pre-trained models often demands substantial computational and storage resources. To address these limitations, PEFT techniques have been developed, particularly within natural language processing (NLP) and computer vision. PEFT methods aim to transfer knowledge from pre-trained models to downstream tasks with minimal parameter updates. 

In recent years, PEFT methods tailored for 3D point cloud understanding have shown potential in balancing performance and efficiency.
IDPT~\cite{zha2023instance} pioneers PEFT for point clouds by employing DGCNN~\cite{wang2019dynamic} to generate instance-level prompts, effectively adapting conventional prompt-based tuning paradigms to geometric data.
Point-PEFT~\cite{tang2024point} applies a domain-specific memory bank and geometry-aware adapters to encode local geometric priors, while DAPT~\cite{zhou2024dynamic} utilizes dynamic adapters that scale tokens according to task-specific importance and integrate internal prompts for instance-adaptive feature modulation.
PPT~\cite{zhang2024positional} injects trainable positional prompt tokens at sampled centers and refines them through lightweight inter-layer adapters to enhance 3D spatial awareness.
PMA~\cite{zha2025pma} leverages Mamba-based sequence modeling to fuse complementary semantics across modalities, facilitating more comprehensive point cloud reasoning.
PointLoRA~\cite{wang2025pointlora} combines low-rank adaptation (LoRA) with multi-scale token selection, achieving efficient fine-tuning with minimal trainable parameters.
Despite these advances, most PEFT methods freeze the backbone network, which restricts fine-grained local feature learning and hinders local-global context integration capabilities crucial for dense prediction tasks~\cite{armeni20163d}.
To address this, PointGST~\cite{liang2024parameter} incorporates frequency-domain spectral bases derived from raw point clouds into intermediate layers, enriching multi-scale global and local context. STAG~\cite{furuya2026token}, GEM~\cite{tang2026geometry}, and GAPrompt~\cite{ai2025gaprompt} are closer geometry-aware methods. STAG~\cite{furuya2026token} adapts coarsened tokens through a side network, GEM~\cite{tang2026geometry} refines token-level positional/context cues with Spatial/Context Adapters, and GAPrompt~\cite{ai2025gaprompt} learns geometry-aware point prompts.

Most existing methods extract single-resolution features from the heavily downsampled point tokens of the backbone (e.g., from 2048 to 128), which improves efficiency but discards fine-grained local structures vital for point cloud understanding. 
In contrast, our proposed PLT introduces an HLN to extract multi-resolution spatial features directly from the original point clouds and an LGF module to adaptively fuse them with global features from the frozen backbone. This design preserves fine-grained local details while retaining rich global priors, enabling more comprehensive point cloud understanding with fewer trainable parameters and superior performance.

%% file: sec/3_method.tex
\section{Methodology}
\label{sec:methodology}

As shown in Fig.~\ref{fig:overall2}, PLT follows a \emph{construct-fuse-feedback} principle for parameter efficient adaptation with a frozen point-cloud backbone. Specifically, we decouple fine-tuning into two tightly coupled pathways: (1) a lightweight \textbf{Hierarchical Ladder Network (HLN)} that constructs a multi-resolution pyramid of local geometric features directly from raw points (\S\ref{sec:HLN}); and (2) a \textbf{prompt-based global adaptation} pathway that injects the constructed locality back into the frozen Transformer via instance-aware, multi-scale prompts (\S\ref{sec:FT_backbone}). These pathways form a bidirectional closed loop through a \textbf{Local-Global Fusion (LGF)} module: LGF selectively fuses HLN features with intermediate pretrained semantics, and the resulting multi-scale cues are fed back to modulate self-attention within the backbone (\S\ref{sec:LGF}). In this way, PLT allows local geometry to influence global reasoning inside the frozen backbone, rather than being exploited only through late fusion at the task head.

Finally, we employ lightweight task-specific heads (\S\ref{sec:head}) and leverage the hierarchical structure for progressive upsampling to recover spatial resolution. This design is particularly beneficial for dense prediction, where preserving spatial fidelity is critical for accurate outputs.

\input{fig/overall}

\subsection{Hierarchical Ladder Network}
\label{sec:HLN}

While transformer-based point cloud encoders capture global semantics well, they often miss fine-grained spatial details due to weak local geometric biases. Existing PEFT methods (e.g., PointPEFT~\cite{tang2024point}, PointGST~\cite{liang2024parameter}) try to improve locality but still depend heavily on frozen backbone features and fixed resolutions, missing key local details needed for dense prediction.

To solve this, we propose a Hierarchical Ladder Network (HLN) that extracts multi-resolution local features from point clouds. HLN better preserves 3D geometric structure, boosting performance in tasks like segmentation. It has two main components: (1) a Set Abstraction (SA) module and (2) a Local-Global Fusion (LGF) module.

Given an input point cloud $\mathbf{P} \in \mathbb{R}^{N \times 3}$, where $N$ is the number of points, we first apply point embedding to map $\mathbf{P}$ into a high-dimensional space, resulting in a initial point cloud features $\mathbf{F} \in \mathbb{R}^{N \times C}$, where $C$ represents the feature dimension. We then use Set Abstraction (SA) for downsampling, starting with farthest point sampling (FPS) to select a set of center points $\mathbf{C}=\{c_1, \ldots, c_n\}$, followed by k-Nearest Neighbors (kNN) to define local neighborhoods: 

\begin{equation}
    \mathcal{N}(c_i) = \left\{(\mathbf{p}_{c_i}^j, \mathbf{f}_{c_i}^j) \mid j = 1, \dots, k \right\}.
\end{equation}

The local features are aggregated as:
\begin{align}
    \hat{\mathbf{f}}_{c_i}^{j} &= \varphi\left(\left[\mathbf{f}_{c_i}^{j}; (\mathbf{p}_{c_i} - \mathbf{p}_{c_i}^j)\right]\right), \\
    \mathbf{f}_{c_i}^{\prime} &= \mathbf{f}_{c_i} + \rho\left(h_{\boldsymbol{\Theta}}([\hat{\mathbf{f}}_{c_i}^{1}; \dots; \hat{\mathbf{f}}_{c_i}^{k}])\right),
\end{align}
where $\varphi$, $\rho$ are learnable functions, $h_{\boldsymbol{\Theta}}$ denotes the aggregation function (e.g., max pooling) and $[\cdot ; \cdot]$ denotes concatenation. By stacking multiple SA layers, HLN constructs a hierarchical local representation $\mathbf{F}_l$ that complements global transformer outputs.

\subsection{Local-Global Fusion Module}
\label{sec:LGF}

A key challenge in point cloud learning is to effectively integrate fine-grained geometric details with high-level semantic context. Existing fusion approaches, such as naive concatenation or element-wise addition, typically fail to adaptively balance these two information sources.

To address this issue, we propose a novel Local-Global Fusion (LGF) module that performs adaptive feature integration via selective attention. Our design is motivated by two key insights: (1) local geometric features extracted by the HLN pathway are critical for precise spatial reasoning; and (2) global features from the frozen pre-trained backbone encode rich semantic priors. LGF allows the network to dynamically balance and selectively integrate these complementary representations in a content-aware manner, ensuring that both local details and global semantics are effectively leveraged.

Let $\mathbf{T}_l$ and $\mathbf{T}_g$ denote the local and global tokens extracted from the HLN and the frozen backbone, respectively. To align their feature dimensions, the backbone features are first linearly projected using a learnable transformation matrix $W$. We then apply an aggregation operation, similar to SA but without spatial downsampling, to capture cross-level interactions between hierarchical feature representations. This process yields global features $\mathbf{F}_g$, which encode the interplay between $\mathbf{T}_l$ and $\mathbf{T}_g$, while the local branch $\mathbf{F}_l$ retains fine-grained intra-local dependencies within $\mathbf{T}_l$.

To effectively integrate these complementary cues, we introduce a selective attention fusion mechanism that adaptively emphasizes the most informative features from each branch. Specifically, a global aggregation function $\mathcal{A}$ (implemented as average pooling) is used to obtain compact descriptors:

\begin{equation}
	\mathbf{f}_l = \mathcal{A}_l(\mathbf{F}_l), \quad \mathbf{f}_g = \mathcal{A}_g(\mathbf{F}_g).
\end{equation}

The resulting descriptors are passed through lightweight MLPs to produce attention logits $\mathbf{z}_l$ and $\mathbf{z}_g$, from which normalized attention weights are derived via a softmax-like operation. These weights modulate the contribution of each feature source, yielding a fused multi-scale representation obtained through a weighted summation. Finally, the fused features are refined by an MLP and combined with the residual shortcut to form the final output $\mathbf{F}_o$, preserving both global contextual consistency and local structural fidelity.

\subsection{Dynamic Prompt-Based Global Adaption}
\label{sec:FT_backbone}
To enable the pre-trained backbone network to produce task-adaptive global representations, we propose a dynamic multi-scale prompt generation strategy driven by the fused feature representation $\mathbf{F}_{o}$ from the LGF module. This design aims to inject instance-specific geometric cues into the transformer backbone, allowing the model to preserve general semantic priors while adapting flexibly to downstream tasks.

Firstly, $\mathbf{F}_o$ is mean-pooled to obtain a compact representation. This pooled feature is then scaled and shifted using learnable parameters $\gamma$ and $\beta$, generating a multi-scale prompt $\mathbf{p}$. To maximize parameter efficiency, we reuse $W$ from the LGF module to align the dimensionality of the multi-scale prompts with that of the backbone network features. The formula for this process is as follows:
\begin{equation}
	\mathbf{p} = \left(\gamma \cdot \frac{1}{n} \sum_{i=1}^{n} \mathbf{F}_o^i + \beta \right) W^T,
\end{equation}
where $n$ is the number of tokens in $\mathbf{F}_o$. 
% It is worth noting that the internal prompt of DAPT~\cite{zhou2024dynamic} can be regarded as a special case of our multi-scale dynamic prompt, which degenerates into the internal attention of DAPT when only the global branch is activated.

Finally, the multi-scale prompts $\mathbf{P}=[\mathbf{p}_{1}, \ldots, \mathbf{p}_{s}]$ are incorporated into the backbone network during training, where $s$ is corresponds to the number of stages in the HLN pathway. These prompts are inserted at each transformer layer of the frozen backbone to condition the global representation in a scale-aware, instance-sensitive manner:
\begin{equation}
	\mathbf{x}_l = \mathcal{L}_l\left([\mathbf{T}_{cls}; \mathbf{P}; \mathbf{T}_{g}]\right),
\end{equation}
where $\mathcal{L}_l$ is the $l$-th transformer layer. 

The injected prompts participate in the attention mechanism, interacting with the original tokens to enrich the contextual modeling without altering the backbone's architecture. This allows the model to adaptively refine global features based on the special task, while maintaining the semantic richness of large-scale pre-training.

To further enhance adaptability, we adopt the lightweight scale-and-shift tuning strategy proposed in SSF~\cite{lian2022scaling}, applying element-wise affine transformations to the output of each backbone module: $\mathbf{y} = s \cdot \mathbf{x} + t$,
where $s$ and $t$ are learnable  scaling and shifting  vectors that fine-tune the global representation with minimal overhead.

\subsection{Lightweight Segmentation Head}
\label{sec:head}

Unlike classification tasks, semantic segmentation requires per-point predictions, necessitating the recovery of features for all original points. A common approach is to propagate features from downsampled representations back to the original point cloud through feature interpolation. However, most existing parameter-efficient fine-tuning methods~\cite{zha2023instance,zhou2024dynamic,liang2024parameter} for point cloud lack the ability to capture multi-resolution information, leading to a considerable loss of local detail.

To address this issue, we introduce an additional HLN branch to extract fine-grained multi-resolution spatial features.
Inspired by PointNext~\cite{qian2022pointnext}, we adopt an inverse distance-weighted interpolation strategy to progressively recover point-level features at each resolution.
Unlike PointNext~\cite{qian2022pointnext}, our method concatenates global backbone features with HLN's local multi-resolution features during each interpolation step, forming a more comprehensive representation for effective upsampling.

\begin{equation}
\mathbf{T}_{\text{interp}} = [\operatorname{Interp}(\mathbf{T}, \mathbf{T}_g); \operatorname{Interp}(\mathbf{T}, \mathbf{T}_l)],
\end{equation}
where $\mathbf{T}$ denotes the point features at the current layer. The function $\operatorname{Interp}(\cdot)$ performs inverse distance interpolation~\cite{qian2022pointnext}.

At each resolution, we concatenate the interpolated features $\mathbf{T}_{\text{interp}}$ with the corresponding original features $\mathbf{T}$, and then pass them through a MLP to obtain updated features. This process is repeated iteratively until the resolution of the original point cloud is restored. Finally, point-wise classification is performed using a MLP followed by a softmax activation to predict the semantic category of each point. This design not only enhances segmentation performance but also substantially reduces the number of trainable parameters and overall computational complexity.

%% file: fig/overall.tex
\begin{figure*}[t]
	 \centering
    \begin{subfigure}{0.53\textwidth}
        \centering
        \includegraphics[width=1.0\linewidth]{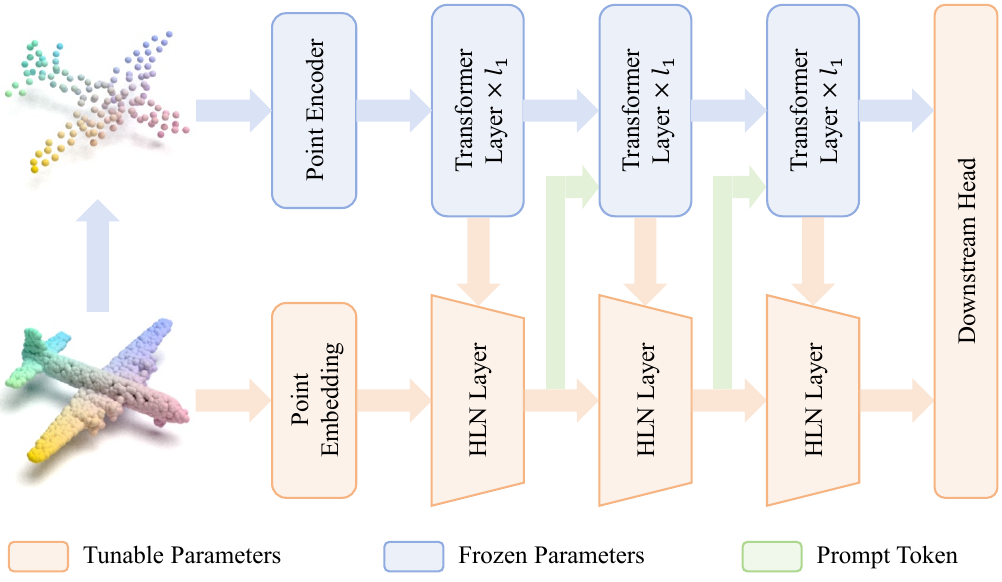}
        \caption{Overall Architecture}
        \label{fig:overall}
    \end{subfigure}
    \hfill
    \begin{subfigure}{0.45\textwidth}
        \centering
        \includegraphics[width=0.9\linewidth]{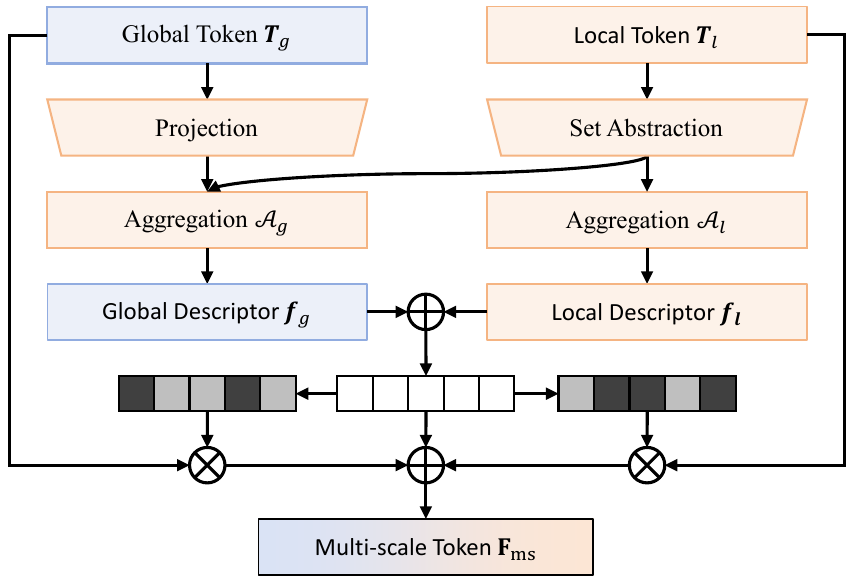}
        \caption{Local-Global Fusion Module (LGF)}
        \label{fig:lgf}
    \end{subfigure}
	\caption{Overview of the proposed Point Ladder Tuning (PLT) framework.  (a) The overall architecture couples a frozen pre-trained backbone with tunable Hierarchical Ladder Network (HLN) layers for adaptive multi-resolution and multi-scale learning. (b) The Local-Global Fusion (LGF) module encodes local tokens via set abstraction (SA) and fuses them with global tokens from the backbone to generate multi-scale representations. This hierarchical design facilitates efficient parameter adaptation, robust representation learning, and scalable performance on downstream point cloud tasks.
	}
    % \vspace{-10pt}
	\label{fig:overall2}
\end{figure*}

%% file: sec/4_experiment.tex
\input{tab/sota}

\section{Experiments}
\label{sec:experiments}

In this section, we comprehensively evaluate PLT on diverse point cloud datasets~\cite{wu20153d, uy2019revisiting, yi2016scalable, armeni20163d, dai2017scannet} and tasks, including 3D object classification, few-shot learning and 3D dense prediction.
Our experiments systematically assess PLT's state-of-the-art performance and parameter efficiency, supported by extensive comparisons with existing fine-tuning approaches and an ablation study validating each key component.

For fair comparison with prior methods~\cite{zha2023instance, zhou2024dynamic}, we adopt identical training configurations. All experiments are conducted on an NVIDIA RTX 3090 GPU, freezing the pre-trained backbone and fine-tuning only the lightweight PLT modules.

To ensure a comprehensive comparison, we employ three state-of-the-art pre-trained models as baselines: Point-BERT~\cite{yu2022point}, Point-MAE~\cite{pang2022masked}, ACT~\cite{dong2022autoencoders} and PointGPT~\cite{chen2023pointgpt}. These models represent diverse pre-training paradigms, shown with detailed discussions in the experiment section.

\input{fig/tsne}

\subsection{3D Object Classification}
\label{sec:classification}

\textbf{Object Classification on Synthetic Dataset.}
We conduct experiments on ModelNet40~\cite{wu20153d} under the same settings as DAPT~\cite{zhou2024dynamic} to ensure a fair and consistent comparison.

As shown in Tab.~\ref{tab:sota}, without voting, PLT achieves 93.8\%, 93.5\%, and 93.6\% on Point-MAE~\cite{pang2022masked}, Point-BERT~\cite{yu2022point}, and ACT~\cite{dong2022autoencoders}, surpassing full fine-tuning by +0.6\%, +0.8\%, +0.6\%.
With voting, PLT further improves Point-BERT by +1.0\%.
These results confirm PLT's effectiveness in point-cloud classification by capturing local structures and leveraging both global and local representations.

\textbf{Object Classification on Real-World Dataset.}
Most pre-trained point cloud models rely on synthetic data like ShapeNet~\cite{chang2015shapenet}, while real-world scans introduce noise, missing points, and background clutter. We evaluate PLT on ScanObjectNN~\cite{uy2019revisiting} (OBJ\_BG, OBJ\_ONLY, PB\_T50\_RS) using Point-BERT~\cite{yu2022point}, Point-MAE~\cite{pang2022masked}, and ACT~\cite{dong2022autoencoders} as baselines to assess robustness in challenging real-world conditions.

Our PLT achieves significant accuracy gains over full fine-tuning with only 2.71\% of parameters, improving Point-BERT~\cite{yu2022point} by +4.14\%, +1.73\% and +3.02\% across ScanObjectNN~\cite{uy2019revisiting} variants.
Under the most challenging PB\_T50\_RS setting, PLT further surpasses PointGST by +0.45\% on Point-BERT~\cite{yu2022point} and +0.24\% on PointMAE~\cite{pang2022masked}, showing strong robustness and efficiency to noisy, occluded data.
Furthermore, it also exhibits the least performance drop on ACT~\cite{dong2022autoencoders}, maintaining stable accuracy where others significantly decline, underscoring its architectural strength in preserving semantic representations during adaptation.

As shown in Fig.~\ref{fig:tsne}, we further visualize the feature manifolds of full fin-tuning, point-cloud-specific PEFT methods and our PLT on the ScanObjectNN PB\_T50\_RS dataset using t-SNE~\cite{van2008visualizing}. Compared with previous methods, Our PLT exhibits significantly greater inter-cluster dispersion and smaller intra-cluster distances, highlighting its superior ability to preserve discriminative features. These results collectively demonstrate that PLT attains an effective balance between representation quality and parameter efficiency, making it well suited for point-cloud learning.

\input{tab/semantic}

\subsection{3D Dense Prediction Task}
For dense prediction tasks, including part and semantic segmentation, we employ a lightweight prediction head to fully leverage multi-resolution features while keeping parameters minimal.

As shown in Tab.~\ref{tab:segmentation}, PLT consistently achieves superior performance across all datasets and backbones with far fewer tunable parameters. On ShapeNetPart~\cite{yi2016scalable}, PLT attains competitive instance-level mIoU (Inst. mIoU) and improves class-level mIoU (Cls. mIoU). Specifically, PLT improves Inst. mIoU on PointBERT~\cite{yu2022point} by 0.5\% over DAPT~\cite{zhou2024dynamic}, demonstrating its effectiveness in capturing fine-grained geometric details.

\input{tab/segmentation}

As shown in Tab.~\ref{tab:semantic_segmentation}, on the large-scale indoor scene benchmarks S3DIS~\cite{armeni20163d} and ScanNetV2~\cite{dai2017scannet}, PLT achieves state-of-the-art results among PEFT methods with only about 2M tunable parameters, far fewer than the 27 M required for full fine-tuning.
With ACT as the backbone on S3DIS~\cite{armeni20163d}, PLT attains 70.6\% mAcc and 61.5\% mIoU, surpassing the closest competitor by 2.3\% in mIoU. On the more challenging ScanNetV2, PLT leads in both voxel- and point-level mIoU (48.2\%/47.8\%) and shows notable gains over DAPT~\cite{zhou2024dynamic} and PointGST~\cite{liang2024parameter} for PointBERT~\cite{yu2022point}.

These results highlight PLT's strong generalization and efficiency in dense 3D scene prediction. Its HLN effectively captures multi-resolution semantic cues from raw points, while the LGF module and dynamic prompts harmonize local details with global priors. A lightweight segmentation head further strengthens dense-prediction accuracy with minimal computational overhead. This synergy enhances fine-grained localization and semantic consistency across varying spatial scales, making PLT a lightweight yet powerful PEFT solution for 3D dense prediction.

\subsection{Model Scaling}
To assess the scalability of the proposed PLT framework, we conduct experiments across point cloud backbones of different model sizes, ranging from regular-scale networks to large-scale foundation models.

As reported in Tab.~\ref{tab:sota} and Tab.~\ref{tab:segmentation}, PLT consistently boosts performance on both classification (ScanObjectNN~\cite{uy2019revisiting}, ModelNet40~\cite{wu20153d}) and segmentation (ShapeNetPart~\cite{yi2016scalable}) benchmarks, regardless of backbone size. In particular, PLT achieves competitive or superior accuracy with only $0.6$M tunable parameters when applied to Point-BERT~\cite{yu2022point}, Point-MAE~\cite{pang2022masked}, and ACT~\cite{dong2022autoencoders}, demonstrating its efficiency in small-scale settings. When scaled to large models such as PointGPT-L, PLT continues to deliver notable improvements (e.g., $+1.94\%$ OA on ScanObjectNN OBJ\_BG and $+1.81\%$ mIoU on ShapeNetPart), while requiring just $1.3$M and $3.85$M trainable parameters for classification and segmentation, corresponding to only $0.36\%$ and $1.13\%$ of the full model size.

These results highlight the strong scalability of PLT: it not only adapts effectively to compact models with minimal overhead, but also remains effective when applied to large-scale foundation models, thereby offering a unified and parameter-efficient tuning paradigm for diverse point cloud architectures.

\subsection{Few-shot Learning}

\label{sec:few_shot}

We further evaluate PLT's transferability on ModelNet40~\cite{wu20153d} dataset for assessing the efficiency of data usage in low-resource settings. Following prior works~\cite{zha2023instance, zhou2024dynamic}, we adopt the standard n-way m-shot protocol, where $n \in \{5, 10\}$ and $m \in \{10, 20\}$.

As shown in Tab.~\ref{tab:fewshot}, PLT consistently surpasses full fine-tuning and leading PEFT methods across most settings and backbones (Point-BERT~\cite{yu2022point}, ACT~\cite{dong2022autoencoders}).
In the 5-way 20-shot setting with Point-BERT~\cite{yu2022point}, PLT reaches 98.8\% accuracy, exceeding full fine-tuning by +2.5\% and PointGST~\cite{liang2024parameter} by +0.9\%.
These results demonstrate PLT's strong generalization in data-scarce scenarios, attributed to its hierarchical local feature extraction, which introduces a robust inductive bias and enhances fine-tuning effectiveness.

In addition to superior accuracy, PLT shows improved stability across few-shot settings, yielding consistently lower or comparable standard deviations than other PEFT baselines.
For example, in the 5-way 20-shot setting with Point-BERT~\cite{yu2022point}, PLT achieves ±1.1\%, outperforming LST~\cite{sung2022lst} (±1.8\%) and PointGST~\cite{liang2024parameter} (±2.0\%).
This consistent trend across configurations highlights PLT's ability to reduce performance variance, which is crucial for robust few-shot learning under limited data.

\input{tab/fewshot}

\subsection{Ablation Study}

\textbf{Comparison Between HLN and PLT.} Fig.~\ref{fig:ablation} shows that \textbf{HLN} alone, while recovering multi-resolution locality from raw points, remains limited (\textbf{80.74\%}) due to the lack of semantic priors from the pre-trained backbone. Coupling HLN with the backbone via \textbf{LGF} raises accuracy to \textbf{83.34\%} (+2.60\%), validating the necessity of \emph{selective} semantic injection. Adding \textbf{Dynamic Prompt} further improves to \textbf{84.52\%} (+1.18 \%) with a comparable tunable budget, highlighting the role of instance-conditioned high-level modulation. Finally, \textbf{SSF} delivers the full \textbf{PLT} performance of \textbf{85.53\%} (+1.01), while removing only DP drops to \textbf{84.66\%}, evidencing clear complementarity between prompt adaptation and layer-wise refinement. Overall, PLT improves over HLN by \textbf{+4.79} \% and even surpasses full fine-tuning (85.18\%) with \textbf{$<3\%$} trainable parameters, underscoring PLT as a co-designed locality-aware PEFT framework for point cloud understanding.

\input{fig/ablation}

\textbf{Fusion Strategy for Local and Global Information.} The integration strategy of local and global information plays a critical role in downstream performance. In Tab.~\ref{tab:fusion_way}, we evaluate several fusion mechanisms. The proposed Local-Global Fusion (LGF) with Softmax achieves the highest accuracy (85.53\%), outperforming additive, concatenative and single-source approaches. Notably, leveraging only local features yields a considerably lower accuracy (82.86\%) than relying solely on global features (84.52\%), reflecting the benefits of pretrained global representation learning.

\input{tab/fusion_way}

\textbf{HLN Source Attribution.}
Tab.~\ref{tab:source} and Tab.~\ref{tab:seghead} isolate the role of the HLN information source in classification and segmentation, respectively. Replacing raw-point HLN with token-center HLN consistently degrades performance, from 85.53\% to 84.73\% OA in classification and from 60.5\% to 59.0\% mIoU under the standard-head setting in segmentation. These results show that PLT's gain is not merely due to additional parameters or an auxiliary branch; rather, it stems from leveraging fine-grained raw-point geometry that complements frozen coarse tokens.

\input{tab/raw_token}

\textbf{Segmentation Head Attribution.} As shown in Tab.~\ref{tab:seghead}, the improvement in dense prediction is not solely attributable to the lightweight segmentation head. Replacing it with the standard segmentation head increases the number of trainable parameters, yet reduces mIoU to 60.5, while still outperforming most 3D PEFT baselines.

\input{tab/seg_ablation}

%% file: tab/sota.tex
\begin{table}[ht]
    \centering
    \scriptsize
    \caption{
Classification results on three variants of ScanObjectNN~\cite{uy2019revisiting} and ModelNet40~\cite{wu20153d}, including the number of trainable parameters and overall accuracy (OA). All methods apply default data augmentation as used in~\cite{zhou2024dynamic}. \textcolor{red}{$^*$} indicates reproduced results. $^{\dagger}$ indicates training with a post–pre-training checkpoint and the use of generated point clouds as additional supervision for the reconstruction loss. For ScanObjectNN~\cite{uy2019revisiting}, results are reported without voting. For ModelNet40~\cite{wu20153d}, results are shown w/o and w/ voting as (-/-). TP represents the tunable parameters.
}
    \label{tab:sota}

\resizebox{1.0\textwidth}{!}{
\begin{tabular}{lccccccc}
\toprule
\multirow{2.3}{*}{Method} &
\multirow{2.3}{*}{Reference} &
\multirow{2.3}{*}{TP (M)} &
\multirow{2.3}{*}{FLOPs (G)} &
\multicolumn{3}{c}{ScanObjectNN} &
\multirow{2.3}{*}{ModelNet40} \\
\cmidrule(r){5-7}
& & & & OBJ\_BG & OBJ\_ONLY & PB\_T50\_RS \\
\midrule

Point-BERT~\cite{yu2022point} & CVPR 22 & 22.1 (100\%) & 4.9 & 87.43 & 88.12 & 83.07 & {92.7} / {\color{gray}{93.2}} \\
+ IDPT~\cite{zha2023instance} & ICCV 23 & 1.7 (7.69\%) & 7.3 & {88.12}\dplus{+0.69} & {88.30}\dplus{+0.18} & {83.69}\dplus{+0.62} & {92.6}{\dtplus{-0.1}} / {\color{gray}{{93.4}}}{\color{gray}{\ddplus{+0.2}}} \\
+ DAPT~\cite{zhou2024dynamic} & CVPR 24 & 1.1 (4.97\%) & 5.1 & {91.05}\dplus{+3.62} & {89.67}\dplus{+1.55} & {85.43}\dplus{+2.36} & {93.1}{\dplus{+0.4}} / {\color{gray}{{93.6}}}{\color{gray}{\ddplus{+0.4}}} \\
+ PMA~\cite{zha2025pma} & CVPR 25 & 1.1 (4.97\%) & - & 91.39\dplus{+3.96} & \textbf{91.05}\dplus{+2.93} & 85.50\dplus{+2.43} & \textbf{93.7}\dplus{+1.0} / ~~-~~~~~~~~~~~~ \\
+ PointGST~\cite{liang2024parameter} & TPAMI 25 & 0.62 (2.81\%) & 5.0 & {91.39}\dplus{+3.96} & {89.67}\dplus{+1.55} & {85.64}\dplus{+2.57} & {93.4}{\dplus{+0.7}} / {\color{gray}{{93.8}}}{\color{gray}{\ddplus{+0.6}}} \\
+ LST~\cite{sung2022lst} & NeurIPS 22 & 0.8 (3.38\%) & 5.0 & {89.15}\dplus{+2.72} & {89.50}\dplus{+1.38} & {83.17}\dplus{+0.10} & {92.9}{\dplus{+0.2}} / {\color{gray}{{93.3}}}{\color{gray}{\ddplus{+0.1}}} \\
\rowcolor{linecolor!40}+ PLT ({ours}) & - & \textbf{0.60} (\textbf{2.71}\%) & 5.0 & \textbf{91.57}\dplus{+4.14} & 89.85\dplus{+1.73} & \textbf{86.09}\dplus{+3.02} & 93.5{\dplus{+0.8}} / {\color{gray}{\textbf{94.2}}}{\color{gray}{\ddplus{+1.0}}} \\
\midrule

Point-MAE~\cite{pang2022masked} & ECCV 22 & 22.1 (100\%) & 4.9 & 90.02 & 88.29 & 85.18 & 93.2 / {\color{gray}{93.8}} \\
+ IDPT~\cite{zha2023instance} & ICCV 23 & 1.7 (7.69\%) & 7.3 & {91.22}\dplus{+1.20} & {90.02}\dplus{+1.73} & {84.94}\dtplus{-0.24} & {93.3}{\dplus{+0.1}} / {\color{gray}{\textbf{94.4}}}{\color{gray}{\ddplus{+0.6}}} \\
+ Point-PEFT\textcolor{red}{$^*$}~\cite{tang2024point} & AAAI 24 & 0.7 (3.13\%) & - & {90.19}\dplus{+0.17} & {89.50}\dplus{+1.21} & {84.35}\dtplus{-0.83} & \textbf{94.2}{\dplus{+1.0}} / ~~~-~~~~~~~~~~ \\
+ DAPT~\cite{zhou2024dynamic} & CVPR 24 & 1.1 (4.97\%) & 5.1 & 90.88\dplus{+0.86} & 90.19\dplus{+1.90} & {85.08}\dtplus{-0.10} & {93.5}{\dplus{+0.3}} / {\color{gray}{{94.0}}}{\color{gray}{\ddplus{+0.2}}} \\
+ PPT\textcolor{red}{$^*$}~\cite{zhang2024positional} & ACM MM 25 & 1.1 (4.97\%) & - & {89.50}\dtplus{-0.52} & {89.50}\dplus{+1.21} & {84.91}\dtplus{-0.27} & {93.7}{\dplus{+0.5}} / ~~~-~~~~~~~~~~ \\
+ PMA~\cite{zha2025pma} & CVPR 25 & 1.1 (4.97\%) & - & 91.05\dplus{+1.03} & \textbf{90.89}\dplus{+2.60} & \textbf{86.43}\dplus{+1.25} & {94.0}{\dplus{+0.2}} / ~~~-~~~~~~~~~~ \\
+ PointGST~\cite{liang2024parameter} & TPAMI 25 & 0.62 (2.81\%) & 5.0 & \textbf{91.74}\dplus{+1.72} & {90.19}\dplus{+1.90} & {85.29}\dplus{+0.11} & {93.5}{\dplus{+0.3}} / {\color{gray}{{94.0}}}{\color{gray}{\ddplus{+0.2}}} \\
+ PointLoRA~\cite{wang2025pointlora} & CVPR 25 & 0.77 (3.43\%) & - & 90.71\dplus{+0.69} & 89.33\dplus{+1.04} & 85.53\dplus{+0.35} & 93.3\dplus{+0.1} / ~~-~~~~~~~~~~~~ \\
+ LST~\cite{sung2022lst} & NeurIPS 22 & 0.8 (3.38\%) & 5.0 & {89.67}\dtplus{-0.35} & {89.67}\dplus{+1.38} & {82.75}\dtplus{-2.43} & {93.2}{\dplus{+0.0}} / {\color{gray}{{93.8}}}{\color{gray}{\ddplus{+0.0}}} \\
\rowcolor{linecolor!40}+ PLT ({ours}) & - & \textbf{0.60} (\textbf{2.71}\%) & 5.0 & 90.88\dplus{+0.86} & {90.02}\dplus{+1.73} & 85.53\dplus{+0.35} & {93.8}{\dplus{+0.6}} / {\color{gray}{{94.0}}}{\color{gray}{\ddplus{+0.2}}} \\
\midrule

ACT~\cite{dong2022autoencoders} & ICLR 23 & 22.1 (100\%) & 4.9 & 91.22 & 89.16 & 85.81 & 93.0 / {\color{gray}{93.7}} \\
+ IDPT~\cite{zha2023instance} & ICCV 23 & 1.7 (7.69\%) & 7.3 & {89.50}\dtplus{-1.72} & {89.33}\dplus{+0.17} & {84.42}\dtplus{-1.39} & {92.9}{\dtplus{-0.1}} / {\color{gray}{{93.6}}}{\color{gray}{\dtplus{-0.1}}} \\
+ Point-PEFT~\cite{tang2024point} & AAAI 24 & 0.7 (3.13\%) & - & {89.33}\dtplus{-1.89} & \textbf{90.88}\dplus{+1.72} & {84.80}\dtplus{-1.01} & \textbf{94.0}{\dplus{+1.0}} / ~~-~~~~~~~~~ \\
+ DAPT~\cite{zhou2024dynamic} & CVPR 24 & 1.1 (4.97\%) & 5.1 & {89.33}\dtplus{-1.89} & {88.30}\dtplus{-0.86} & {83.80}\dtplus{-2.01} & {92.8}{\dtplus{-0.2}} / {\color{gray}{{93.4}}}{\color{gray}{\dtplus{-0.3}}} \\
+ PPT~\cite{zhang2024positional} & ACM MM 25 & 1.1 (4.97\%) & - & {88.98}\dtplus{-2.24} & {88.98}\dtplus{+0.18} & {85.05}\dtplus{-0.76} & {92.9}{\dtplus{-0.1}} / {\color{gray}{93.8}}{\color{gray}{\ddplus{+0.1}}} \\
+ PointGST~\cite{liang2024parameter} & TPAMI 25 & 0.62 (2.81\%) & 5.0 & {88.81}\dtplus{-2.41} & {89.33}\dplus{+0.17} & {84.28}\dtplus{-1.53} & {93.2}{\dplus{+0.2}} / {\color{gray}{{93.5}}}{\color{gray}{\dtplus{-0.2}}} \\
+ LST~\cite{sung2022lst} & NeurIPS 22 & 0.8 (3.38\%) & 5.0 & {90.02}\dtplus{-1.20} & {89.50}\dplus{+0.34} & {82.41}\dtplus{-3.40} & {93.4}{\dplus{+0.4}} / {\color{gray}{{93.6}}}{\color{gray}{\dtplus{-0.1}}} \\
\rowcolor{linecolor!40}+ PLT ({ours}) & - & \textbf{0.60} (\textbf{2.71}\%) & 5.0 & \textbf{90.71}\dtplus{-0.51} & {90.71}\dplus{+1.55} & \textbf{85.39}\dtplus{-0.42} & {93.6}{\dplus{+0.6}} / {\color{gray}{\textbf{94.0}}}{\color{gray}{\ddplus{+0.3}}} \\
\midrule

PointGPT-L$^{\dagger}$~\cite{chen2023pointgpt} & NeurIPS 23 & 360.5 (100\%) & 67.7 & 97.2 & 96.6 & 93.4 & 94.1 / {\color{gray}{94.7}} \\
+ IDPT~\cite{zha2023instance} & ICCV 23 & 10.0 (2.77\%) & 75.2 & {98.11}\dplus{+0.91} & {96.04}\dtplus{-0.56} & {92.99}\dtplus{-0.41} & {93.4}{\dtplus{-0.7}} / {\color{gray}{{94.6}}}{\color{gray}{\dtplus{-0.1}}} \\
+ DAPT~\cite{zhou2024dynamic} & CVPR 24 & 4.2 (1.17\%) & 71.6 & {98.11}\dplus{+0.91} & {96.21}\dtplus{-0.39} & {93.02}\dtplus{-0.38} & {94.2}{\dplus{+0.1}} / {\color{gray}{{94.9}}}{\color{gray}{\ddplus{+0.2}}} \\
+ PMA~\cite{zha2025pma} & CVPR 25 & 4.9 (1.36\%) & - & 98.97\dplus{+1.77} & 96.73\dplus{+0.13} & 95.18\dplus{+1.78} & \textbf{94.9}{\dplus{+0.8}} / {\color{gray}{\textbf{95.4}}}{\color{gray}{\ddplus{+0.7}}} \\
+ PointGST~\cite{liang2024parameter} & TPAMI 25 & 2.4 (0.67\%) & 68.1 & {98.97}\dplus{+1.77} & \textbf{97.59}\dplus{+0.99} & {94.83}\dplus{+1.43} & {94.8}{\dplus{+0.7}} / {\color{gray}{{95.3}}}{\color{gray}{\ddplus{+0.6}}} \\
+ LST~\cite{sung2022lst} & NeurIPS 22 & 1.7 (0.47\%) & 67.9 & {97.76}\dplus{+0.56} & {95.53}\dtplus{-1.07} & {92.75}\dtplus{-0.65} & {93.4}{\dtplus{-0.7}} / {\color{gray}{{94.5}}}{\color{gray}{\dtplus{-0.2}}} \\
\rowcolor{linecolor!40}+ PLT ({ours}) & - & \textbf{1.3} (\textbf{0.36}\%) & 68.2 & \textbf{99.14}\dplus{+1.94} & {97.25}\dplus{+0.65} & \textbf{95.21}\dplus{+1.81} & {94.5}{\dplus{+0.4}} / {\color{gray}{95.0}}{\color{gray}{\ddplus{+0.3}}} \\

\bottomrule
\end{tabular}%
}
\end{table}%

%% file: fig/tsne.tex
\begin{figure}[ht]
    \centering
    \scriptsize
    \begin{subfigure}{0.18\textwidth}
        \centering
        \includegraphics[width=\linewidth]{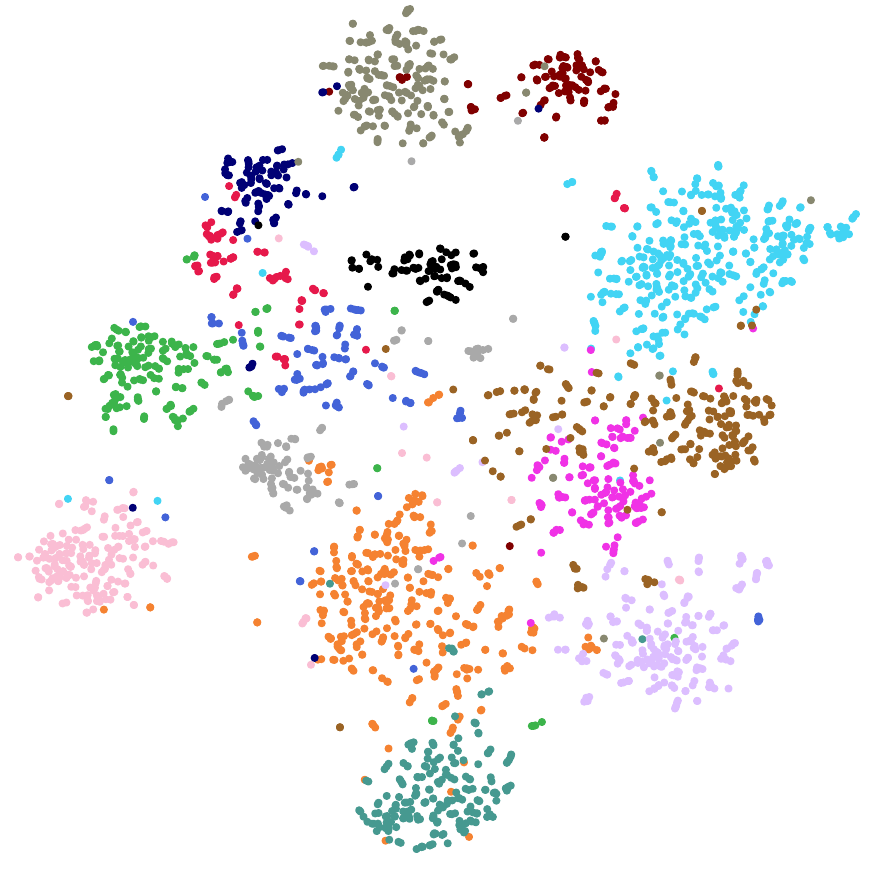}
        % \subcaption*{\textbf{TP}:22.1M \\ \textbf{OA}:85.18}
        \caption{Full}
        \label{fig:sub1}
    \end{subfigure}
    \hfill
    \begin{subfigure}{0.18\textwidth}
        \centering
        \includegraphics[width=\linewidth]{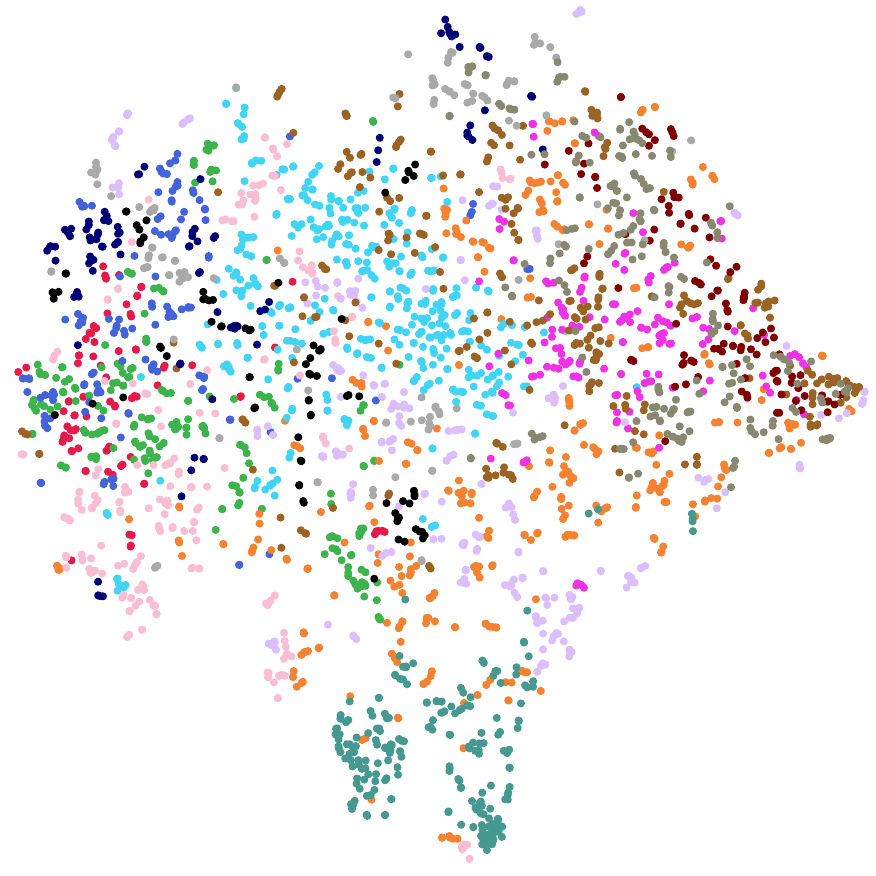}
        % \subcaption*{\textbf{TP}:0.3M \\ \textbf{OA}:75.99}
        \caption{LP}
        \label{fig:sub2}
    \end{subfigure}
    \hfill
    \begin{subfigure}{0.18\textwidth}
        \centering
        \includegraphics[width=\linewidth]{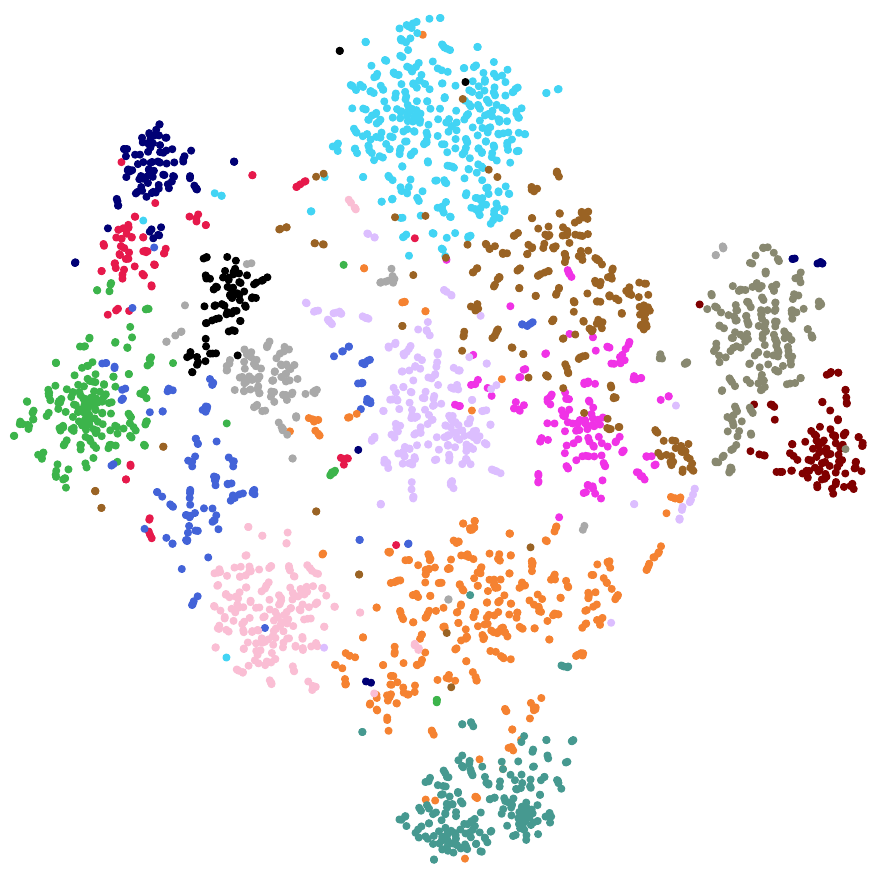}
        % \subcaption*{\textbf{TP}:1.7M \\ \textbf{OA}:84.94}
        \caption{IDPT\cite{zha2023instance}}
        \label{fig:sub3}
    \end{subfigure}
    \hfill
    \begin{subfigure}{0.18\textwidth}
        \centering
        \includegraphics[width=\linewidth]{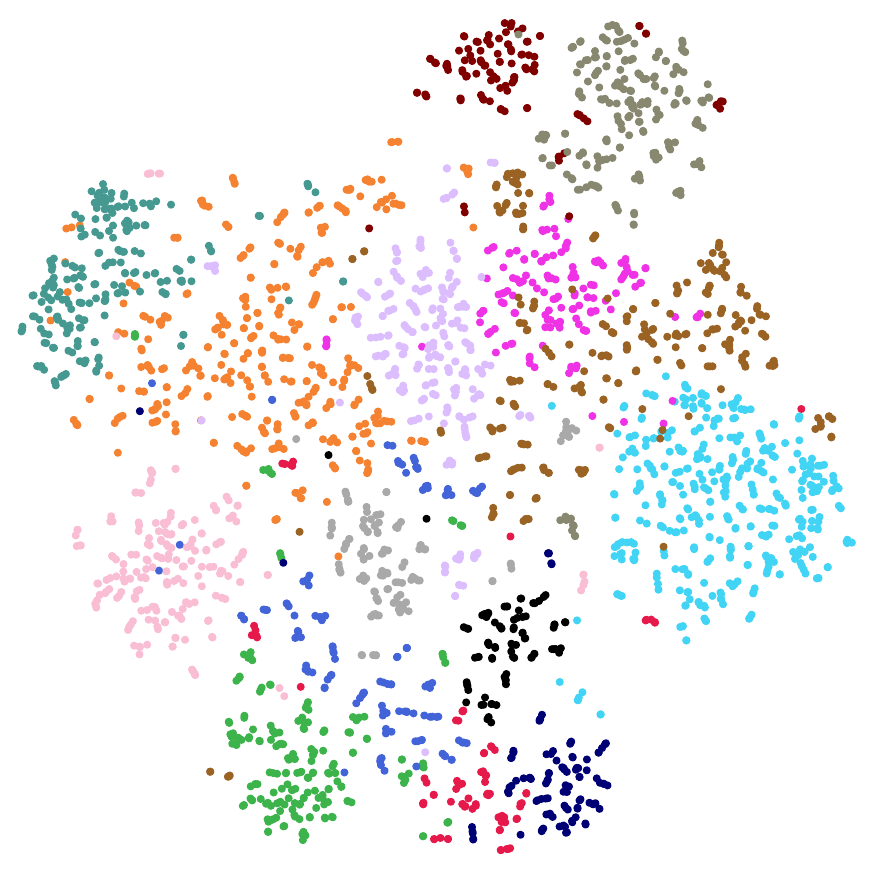}
        % \subcaption*{\textbf{TP}:1.1M \\ \textbf{OA}:85.08}
        \caption{DAPT\cite{zhou2024dynamic}}
        \label{fig:sub4}
    \end{subfigure}
    \hfill
    \begin{subfigure}{0.18\textwidth}
        \centering
        \includegraphics[width=\linewidth]{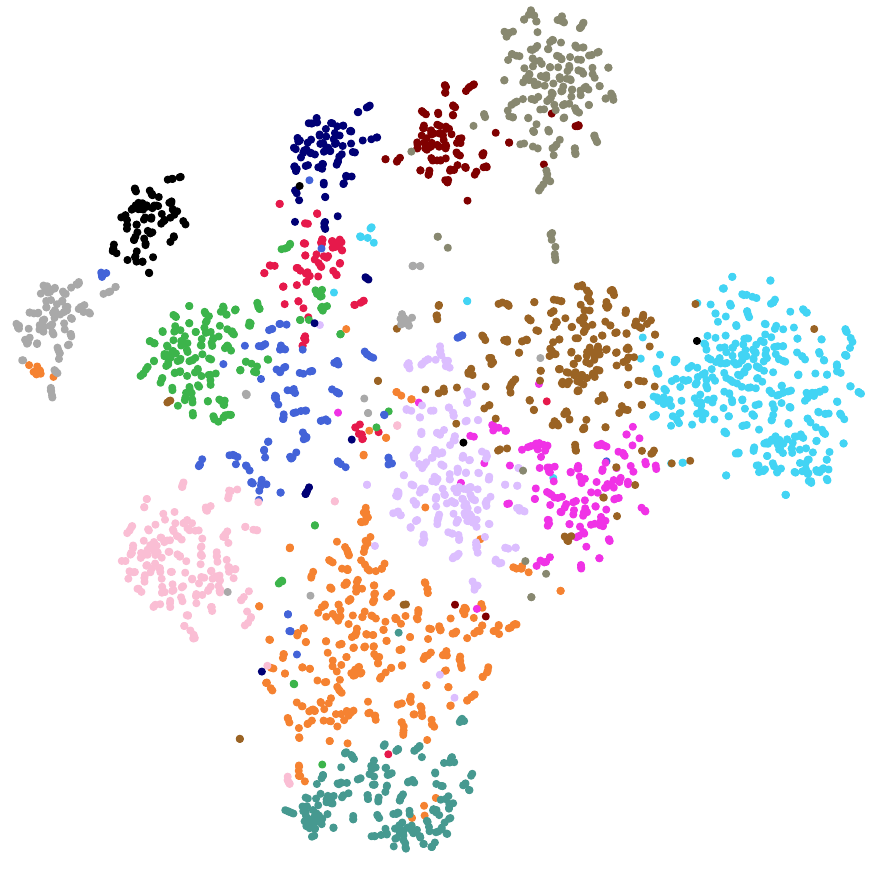}
        % \subcaption*{\textbf{TP}:0.6M \\ \textbf{OA}:85.53}
        \caption{PLT (Ours)}
        \label{fig:sub8}
    \end{subfigure}
    \caption{The visualization of the t-SNE~\cite{van2008visualizing} from the test sets of ScanObjectNN~\cite{uy2019revisiting} (PB\_T50\_RS) by using a pre-trained PointMAE~\cite{pang2022masked} with various fine-tuning strategies. We extract the final classification features from the top linear layer for t-SNE visualizations. LP represents Linear Probing.}
    \label{fig:tsne}
\end{figure}

%% file: tab/semantic.tex
\begin{table}[ht]
  \centering
  \scriptsize
  \caption{Semantic segmentation on the S3DIS~\cite{armeni20163d} and ScanNetV2~\cite{dai2017scannet}. The
    mean accuracy (mAcc) and mean IoU (mIoU) are reported on the S3DIS. VmIoU and PmIoU represents Voxel mIoU and Point mIoU respectively.
    TP represents the tunable parameters.}
    \begin{tabular}{lcccccc}
    \toprule
    Methods & Reference & TP (M)& \multicolumn{2}{c}{S3DIS} & \multicolumn{2}{c}{ScanNetV2} \\
    & & & mAcc & mIoU & VmIoU & PmIoU \\
    \midrule
    Point-BERT~\cite{yu2022point} &  CVPR 22 & 27.02 (100\%) & 69.7 & 62.1 & 49.9 & 49.6 \\ 
    % + Linear probing & - & 5.20  & 65.9  & 56.0 & 30.3 & 30.1  \\
    + IDPT~\cite{zha2023instance} & ICCV 23 & 5.64 (20.9\%)  & 66.9  & 57.7 & 33.9 & 33.6  \\
    + DAPT~\cite{zhou2024dynamic} & CVPR 24 & 5.61 (20.8\%)  & 68.3 & 58.9 & 43.6 & 43.3 \\
    + PPT~\cite{zhang2024positional} & ACM MM  25 & 5.58 (20.7\%) & 68.8 & 60.5 & 46.7 & 46.3 \\
    + PointGST~\cite{liang2024parameter} & TPAMI 25 & 5.55 (20.5\%)  & 68.5 & 59.5 & 46.3 & 45.8  \\
    + LST~\cite{sung2022lst} & NeurIPS 22 & 5.65 (20.9\%) & 67.4 & 58.6 & 34.5 & 34.2  \\
    \rowcolor{linecolor!40}+ PLT (\textbf{ours})& - & \textbf{2.04} (\textbf{7.55\%}) & \textbf{69.6} & \textbf{61.1} & \textbf{48.2} & \textbf{47.8}  \\
    \midrule
    Point-MAE~\cite{pang2022masked} &  ECCV 22 & 27.02 (100\%) & 69.9 & 60.8 & 51.2 & 50.8 \\ 
    % + Linear probing & - & 5.20  & 63.4  & 52.5 & 30.9 & 30.7 \\
    + IDPT~\cite{zha2023instance} & ICCV 23 & 5.64 (20.9\%)  & 66.6  & 57.6 & 33.5 & 33.2  \\
    + Point-PEFT~\cite{tang2024point} & AAAI 24 & 5.58 (20.7\%)  & 66.5  & 56.0 & - & -  \\
    + DAPT~\cite{zhou2024dynamic} & CVPR 24 & 5.61 (20.8\%)  & 68.2 & 59.3 & 41.1 & 40.7 \\
    + PPT~\cite{zhang2024positional} & ACM MM  25 & 5.58 (20.7\%) & 68.7 & 60.4 & 46.2 & 45.8\\
    + PointGST~\cite{liang2024parameter} & TPAMI 25 & 5.55 (20.5\%)  & 68.4 & 58.6 & 44.6 & 44.2\\
    + LST~\cite{sung2022lst} & NeurIPS 22 & 5.65 (20.9\%) & 66.9 & 57.7 & 35.2 & 34.9 \\
    \rowcolor{linecolor!40}+ PLT (\textbf{ours})& - & \textbf{2.04} (\textbf{7.55\%}) & \textbf{70.5} & \textbf{61.5} & \textbf{47.2} & \textbf{46.8} \\
    \midrule
    ACT~\cite{dong2022autoencoders} &  ICLR 23 & 27.02 (100\%) & 71.1 & 61.2 & 50.9 & 50.5 \\ 
    % + Linear probing & - & 5.20  & 64.1  & 52.0 & 27.2 & 27.0 \\
    + IDPT~\cite{zha2023instance} & ICCV 23 & 5.64 (20.9\%)  & 64.9  & 55.2 & 32.1 & 31.9 \\
    + Point-PEFT~\cite{tang2024point} & AAAI 24 & 5.58 (20.7\%) & 66.0  & 54.6 & - & - \\
    + DAPT~\cite{zhou2024dynamic} & CVPR 24 & 5.61 (20.8\%)  & 67.9 & 58.5 & 40.3 & 40.0 \\
    + PPT~\cite{zhang2024positional} & ACM MM  25 & 5.58 (20.7\%) & 69.1 & 59.2 & 45.5 & 45.2\\
    + PointGST~\cite{liang2024parameter} &TPAMI 25 & 5.55 (20.5\%)  & 67.6 & 57.4 & 44.2 & 43.8 \\
    + LST~\cite{sung2022lst} & NeurIPS 22 & 5.65 (20.9\%) & 67.6 & 58.3 & 34.0 & 33.7 \\
    \rowcolor{linecolor!40}+ PLT (\textbf{ours})& - & \textbf{2.04} (\textbf{7.55\%}) & \textbf{70.6} & \textbf{61.5} & \textbf{46.9} & \textbf{46.6} \\
    \bottomrule
    \end{tabular}
  \label{tab:semantic_segmentation}
  % \vspace{-10pt}
\end{table}

%% file: tab/segmentation.tex
\begin{table}[ht]
  \centering
  \scriptsize
  \caption{Part segmentation on the ShapeNetPart~\cite{yi2016scalable}. The mIoU for all classes (Cls.) and for all instances (Inst.) are reported. TP represents the tunable parameters. \textcolor{red}{$^*$} denotes reproduced results, while $^{\dagger}$ indicates training initialized from a post-pre-training checkpoint.}
    \begin{tabular}{lcccc}
    \toprule
    Methods & Reference & TP (M)& Cls. mIoU (\%) & Inst. mIoU (\%) \\
    \midrule
    Point-BERT~\cite{yu2022point} &  CVPR 22 & 27.06 (100\%) & 84.11 & 85.6 \\ 
    + IDPT\textcolor{red}{$^*$}~\cite{zha2023instance} & ICCV 23 & 5.69 (21.0\%) & 83.50  & 85.3  \\
    + DAPT~\cite{zhou2024dynamic} & CVPR 24 & 5.65 (20.9\%)  & 83.83 & 85.5 \\
    + PMA~\cite{zha2025pma} & CVPR 25 & 5.64 (20.8\%) & \textbf{83.96} & \textbf{86.1} \\
    + PointGST~\cite{liang2024parameter} & TPAMI 25 & 5.58 (20.6\%)  & 83.87 & 85.7 \\
    + LST~\cite{sung2022lst} & NeurIPS 22 & 5.69 (21.0\%) & 83.38 & 85.2 \\
    \rowcolor{linecolor!40}+ PLT (\textbf{ours})& - & \textbf{2.08} (\textbf{7.69\%})  & 83.85 & 86.0 \\
    \midrule
    Point-MAE~\cite{pang2022masked} &  ECCV 22 & 27.06 (100\%) & 84.19 & 86.1 \\ 
    + IDPT~\cite{zha2023instance} & ICCV 23 & 5.69 (21.0\%)  & 83.79  & 85.7  \\
    + DAPT~\cite{zhou2024dynamic} & CVPR 24 & 5.65 (20.9\%)  & 84.01 & 85.7 \\
    + PPT~\cite{zhang2024positional} & ACM MM  25 & 5.62 (20.6\%)  & \textbf{84.07} & 85.7 \\
    + PMA~\cite{zha2025pma} & CVPR 25 & 5.64 (20.8\%) & 84.01 & \textbf{86.1} \\
    + PointGST~\cite{liang2024parameter} & TPAMI 25 & 5.59 (21.0\%)  & 83.98 & 85.8 \\
    + LST~\cite{sung2022lst} & NeurIPS 22 & 5.69 (21.0\%) & 83.78 & 85.5\\
    \rowcolor{linecolor!40}+ PLT (\textbf{ours})& - & \textbf{2.08} (\textbf{7.69\%}) & 83.90 & 85.9 \\
    \midrule
    ACT~\cite{dong2022autoencoders}  & ICLR 23 &  27.06 (100\%) & 84.66 & 86.1 \\
    + IDPT~\cite{zha2023instance} & ICCV 23 & 5.69 (21.0\%)  & 83.80  & 85.7  \\
    + DAPT~\cite{zhou2024dynamic} & CVPR 24 & 5.65 (20.9\%)  & 83.56 & 85.7 \\
    + PPT~\cite{zhang2024positional} & ACM MM  25 & 5.62 (20.6\%)  & 83.83 & 85.7 \\
    + PointGST~\cite{liang2024parameter} & TPAMI 25 & 5.59 (21.0\%)  & 83.80 & 85.6 \\
    + LST~\cite{sung2022lst} & NeurIPS 22 & 5.69 (21.0\%) & 83.70 & 85.6\\
    \rowcolor{linecolor!40}+ PLT (\textbf{ours})& - & \textbf{2.08} (\textbf{7.69\%}) & \textbf{83.88} & \textbf{86.0} \\
    \midrule
    \rowcolor{gray!40}\color{gray}{PointGPT-L$^{\dagger}$}~\cite{chen2023pointgpt}  & \color{gray}{NeurIPS 23} &  \color{gray}{339.4 (100\%)} & \color{gray}{84.8} & \color{gray}{86.6} \\
    PointGPT-L$^{\dagger}$\textcolor{red}{$^*$}~\cite{chen2023pointgpt} & NeurIPS 23 &  339.4 (100\%) & 84.16 & 86.1 \\
    + IDPT~\cite{zha2023instance} & ICCV 23 & 33.17 (9.77\%)  & 83.14  & 85.3  \\
    + DAPT~\cite{zhou2024dynamic} & CVPR 24 & 33.61 (9.90\%)  & 83.17 & 85.3 \\
    + PointGST~\cite{liang2024parameter} & TPAMI 25 & 31.84 (9.38\%)  & 83.69 & 85.8 \\
    + LST~\cite{sung2022lst} & NeurIPS 22 & 32.06 (9.45\%) & 82.75 & 85.2\\
    \rowcolor{linecolor!40}+ PLT (\textbf{ours})& - & \textbf{3.85} (\textbf{1.13\%}) & \textbf{84.07} & \textbf{86.2} \\
    \bottomrule
    \end{tabular}
  \label{tab:segmentation}
  % \vspace{-10pt}
\end{table}

%% file: tab/fewshot.tex
\begin{table}[h]
  \centering
  \scriptsize
  \caption{Few-shot learning on ModelNet40~\cite{wu20153d}, Overall accuracy (\%)$\pm$ standard deviation (\%)  w/o voting is reported.}
    \begin{tabular}{lccccc}
    \toprule
   \multirow{2.3}{*}{Methods}&\multirow{2.3}{*}{Reference} & \multicolumn{2}{c}{5-way} & \multicolumn{2}{c}{10-way} \\
\cmidrule{3-6}  &        & 10-shot & 20-shot & 10-shot & 20-shot \\
    \midrule
   Point-BERT~\cite{yu2022point} (baseline) & CVPR 22 &94.6$\pm$3.1 & 96.3$\pm$2.7 & 91.0$\pm$5.4 & 92.7$\pm$5.1 \\
   + IDPT~\cite{zha2023instance} & ICCV 23    & 96.0$\pm$\textbf{1.7}& 97.2$\pm$2.6& 91.9$\pm$4.4& 93.6$\pm$3.5\\
   + DAPT~\cite{zhou2024dynamic} & CVPR 24&95.8$\pm$2.1 &97.3$\pm$1.3 &92.2$\pm$4.3 &94.2$\pm$3.4 \\
   + PointGST~\cite{liang2024parameter}  & TPAMI 25&96.5$\pm$2.4 &97.9$\pm$2.0 &92.7$\pm$4.2 &95.0$\pm$\textbf{2.8} \\
   + LST~\cite{sung2022lst}  & NIPS 22&94.3$\pm$2.6 &97.1$\pm$1.8 &90.6$\pm$4.7 &93.7$\pm$3.7 \\
   \rowcolor{linecolor!40}+ PLT (\textbf{ours}) & -&\textbf{96.9}$\pm$2.0 &\textbf{98.8}$\pm$\textbf{1.1}&\textbf{93.3}$\pm$\textbf{4.0} &\textbf{95.5}$\pm$3.1 \\
    \midrule
   ACT~\cite{dong2022autoencoders} (baseline) & ICLR 23 & 96.8$\pm$2.3 & 98.0$\pm$1.4 & 93.3$\pm$4.0 & 95.6$\pm$2.8\\
   + IDPT~\cite{zha2023instance} &   ICCV 23    & 96.8$\pm$1.9& 98.3$\pm$1.2& 92.5$\pm$4.1& 95.3$\pm$3.3\\
   + DAPT~\cite{zhou2024dynamic} & CVPR 24 & 95.3$\pm$2.8  &  97.1$\pm$1.7 & 89.8$\pm$4.8 & 94.1$\pm$3.6  \\
   + PPT~\cite{zhang2024positional}  & ACM MM 25 & \textbf{97.1}$\pm$2.3 &98.1$\pm$1.8 &91.8$\pm$4.3 &94.9$\pm$3.4 \\
   + PointGST~\cite{liang2024parameter} & TPAMI 25 & 97.2$\pm$1.9  &  98.3$\pm$1.3 & 92.9$\pm$4.2 & 95.7$\pm$\textbf{2.6}  \\
   + LST~\cite{sung2022lst}  & NIPS 22&96.2$\pm$2.6 &98.0$\pm$1.9 &92.6$\pm$4.4 &94.9$\pm$3.4 \\
   \rowcolor{linecolor!40}+ PLT (\textbf{ours}) & - & 96.9$\pm$\textbf{1.8}  &  \textbf{98.9}$\pm$\textbf{1.0} & \textbf{93.4}$\pm$\textbf{4.0} & \textbf{95.9}$\pm$3.1  \\
    \bottomrule
    \end{tabular}%
  \label{tab:fewshot}%
  % \vspace{-10pt}
\end{table}%

%% file: fig/ablation.tex
\begin{figure}[h]
    \centering
    \includegraphics[width=0.85\linewidth]{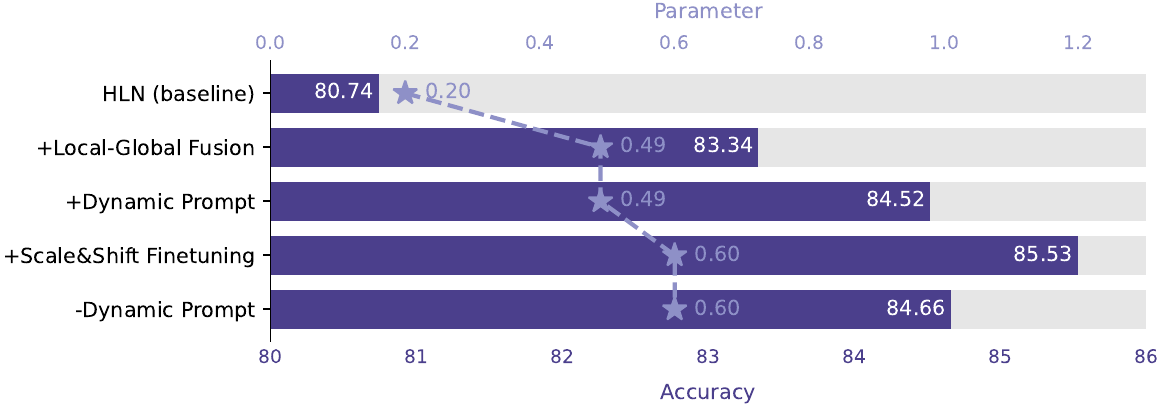}
    \caption{Ablation study demonstrating the necessity and synergy of each PLT component.}
    \label{fig:ablation}
    % \vspace{-10pt}
\end{figure}

%% file: tab/fusion_way.tex
\begin{table}[h]
\scriptsize
\setlength{\tabcolsep}{7.3mm}
\centering
\caption{Fusion way for local and global information in point clouds on PointMAE~\cite{pang2022masked}. The tunable parameters (TP) and the overall accuracy (\%) on the hardest variant of ScanObjectNN\cite{uy2019revisiting} are reported.}
\label{tab:fusion_way}

\begin{tabular}{ccc}
\toprule
Fusion Way & TP (M) & PB\_T50\_RS \\
\midrule
Add & 0.58 & 85.01 \\
Concat & 0.62 & 84.63 \\
Only Global & 0.56 & 84.52 \\
Only Local & 0.56 & 82.86 \\
LGF with Sigmoid & 0.59 & 84.59 \\
\rowcolor{linecolor!40}LGF with Softmax & 0.60 & \textbf{85.53} \\
\bottomrule
\end{tabular}
% \vspace{-10pt}
\end{table}

%% file: tab/raw_token.tex
\begin{table}[h]
\centering
\caption{Raw-point versus token-center/random-center HLN under the same parameter
budget with PointMAE~\cite{pang2022masked} on ScanObjectNN PB\_T50\_RS~\cite{uy2019revisiting}. Token-center HLN changes the HLN source from raw points to backbone token
centers; Random center uses randomly sampled centers.}
\label{tab:source}
\scriptsize
\setlength{\tabcolsep}{5mm}
\begin{tabular}{lccc}
\toprule
Variant & \textbf{Ours} & Token-center HLN & Random center \\
\midrule
\tp(M) & 0.60 & 0.60 & 0.60 \\
OA(\%) & \textbf{85.53} & 84.73 & 85.01 \\
\bottomrule
\end{tabular}
\end{table}

%% file: tab/seg_ablation.tex
\begin{table}[h]
\centering
\caption{Segmentation head and HLN source controls with ACT~\cite{dong2022autoencoders} on S3DIS~\cite{armeni20163d}. Ours + Standard Head replaces the lightweight head with the standard head. Token-center HLN further changes the HLN source from raw points to backbone token centers.}
\label{tab:seghead}
\scriptsize
\setlength{\tabcolsep}{2.5mm}
\begin{tabular}{lccc}
\toprule
Variant & \textbf{Ours} & Ours + Standard Head & Token-center HLN + Standard Head \\
\midrule
\tp(M) & 2.04 & 5.86 & 5.86 \\
mIoU(\%) & \textbf{61.5} & 60.5 & 59.0 \\
\bottomrule
\end{tabular}
\end{table}

%% file: sec/5_conclusion.tex
\section{Conclusion}
\label{sec:conclusion&limitation}

We proposed Point Ladder Tuning (PLT), a parameter-efficient fine-tuning framework for point-cloud analysis. PLT freezes the pre-trained backbone and employs a Hierarchical Ladder Network (HLN) to extract local cues, integrated with global priors through Local-Global Fusion (LGF). A dynamic prompt generator further guides the frozen backbone toward task-aware adaptation, while a lightweight segmentation head boosts dense-prediction accuracy with minimal overhead.  Experiments show PLT achieves strong results in classification and segmentation with minimal added parameters.

% Acknowledgment section
% \section{Acknowledgment}
% This work was supported in part by the National Science and Technology Major Project under Grant 2025ZD1303200.

%The authors would like to thank the anonymous reviewers for their insightful comments and suggestions, which have significantly improved the quality of this paper.

%% file: sec/X_suppl.tex
\setcounter{page}{1}
%\appendix

\section{Dataset and Training Detail}

All experiments were conducted using a single GeForce RTX 3090 GPU. During the training phase, the weights of the pre-trained model were kept fixed, and only a limited subset of parameters associated with the added modules were fine-tuned. The detailed hyperparameter configurations of the model are presented in Tab.~\ref{tab:paramas}

\subsection{ModelNet40}
The ModelNet40 dataset~\cite{wu20153d} is a canonical benchmark for 3D object classification. It contains 12,311 clean CAD models uniformly sampled into point clouds across 40 object categories, with a standard split of 9,843 training and 2,468 testing samples. Following the protocol in DAPT~\cite{zhou2024dynamic}, we employ random scaling and random translation as data augmentation to enhance robustness. Training utilizes the AdamW optimizer with an initial learning rate of $5 \times 10^{-4}$, weight decay of $5 \times 10^{-2}$, and a cosine learning rate scheduler with a 10-epoch warm-up. Models are trained for 300 epochs using a batch size of 32.

\subsection{ScanObjectNN}
To assess performance under real-world conditions, we adopt the ScanObjectNNdataset~\cite{uy2019revisiting}, which includes approximately 15k point cloud samples across 15 categories captured from indoor scenes. These data introduce realistic challenges such as background clutter, occlusion, and non-uniform sampling. Training settings remain consistent with those of ModelNet40 to ensure comparability.

\subsection{Few-Shot Learning on ModelNet40}
We further evaluate the transferability of PLT in low-resource scenarios using a few-shot protocol on ModelNet40. Following~\cite{zha2023instance, zhou2024dynamic}, we adopt the standard $n$-way $m$-shot evaluation where $n \in \{5,10\}$ and $m \in \{10,20\}$. The training configuration mirrors the classification setup, except that the number of training epochs is reduced to 150 to align with few-shot learning conventions.

\subsection{ShapeNetPart}
For part-level segmentation, we use the ShapeNetPart dataset~\cite{yi2016scalable}, which consists of 16,881 3D objects spanning 16 object classes and annotated with 50 part categories. Training adopts the AdamW optimizer with a learning rate of $2 \times 10^{-4}$, weight decay of $5 \times 10^{-2}$, and a cosine scheduler with a 10-epoch warm-up. All models are trained for 300 epochs using a batch size of 16.

\subsection{S3DIS and ScanNetV2}
For scene-level semantic segmentation, we employ two large-scale indoor benchmarks:
\paragraph{S3DIS}~\cite{armeni20163d} is a large-scale indoor scene dataset comprising six areas with a total of 273 million points annotated across 13 categories. Following established practices~\cite{dong2022autoencoders}, Area 5 is recommended for evaluation to provide a more reliable and standardized assessment of performance in semantic segmentation tasks. The model is trained using a cosine learning rate scheduler with an initial learning rate of $2 \times 10^{-4}$. Training is performed over 60 epochs with a batch size of 32, while other configurations are consistent with those used for ShapeNetPart~\cite{yi2016scalable}.
\paragraph{ScanNetV2}~\cite{dai2017scannet} is a large-scale benchmark dataset for indoor 3D scene understanding, encompassing a diverse range of environments, from compact residential and office spaces to expansive public and commercial buildings. The dataset comprises 1,513 RGB-D scanned scenes with annotations for 20 semantic categories. Following standard practice~\cite{dai2017scannet}, 1,201 scenes are used for training and 312 scenes for testing. For training, we adopt a cosine annealing learning rate scheduler, with an initial learning rate set to $5 \times 10^{-3}$. The models are trained for 500 epochs using a batch size of 32.

\input{tab/appendix/parames}

\section{Additional Experiments}

% \subsection{Few-Shot Learning}
% \label{sec:few_shot}

% We further evaluate PLT's transferability on ModelNet40~\cite{wu20153d} dataset for assessing the efficiency of data usage in low-resource settings. Following prior works~\cite{zha2023instance, zhou2024dynamic}, we adopt the standard n-way m-shot protocol, where $n \in \{5, 10\}$ and $m \in \{10, 20\}$.

% As shown in Tab.~\ref{tab:fewshot}, PLT consistently surpasses full fine-tuning and leading PEFT methods across most settings and backbones (Point-BERT~\cite{yu2022point}, Point-MAE~\cite{pang2022masked}, ACT~\cite{dong2022autoencoders}).
% In the 5-way 20-shot setting with Point-BERT~\cite{yu2022point}, PLT reaches 98.8\% accuracy, exceeding full fine-tuning by +2.5\% and PointGST~\cite{liang2024parameter} by +0.9\%.
% These results demonstrate PLT's strong generalization in data-scarce scenarios, attributed to its hierarchical local feature extraction, which introduces a robust inductive bias and enhances fine-tuning effectiveness.

% In addition to superior accuracy, PLT shows improved stability across few-shot settings, yielding consistently lower or comparable standard deviations than other PEFT baselines.
% For example, in the 5-way 20-shot setting with Point-BERT~\cite{yu2022point}, PLT achieves ±1.1\%, outperforming LST~\cite{sung2022lst} (±1.8\%) and PointGST~\cite{liang2024parameter} (±2.0\%).
% This consistent trend across configurations highlights PLT's ability to reduce performance variance, which is crucial for robust few-shot learning under limited data.

% \input{tab/fewshot}

\subsection{Cross-Domain PEFT Comparison}
\label{sec:compare}

We compare our PLT with a broad range of PEFT methods originally proposed for NLP and 2D vision, adapting them to point cloud scenarios. As shown in Tab.~\ref{tab:origin_finetuning}, methods like VPT~\cite{jia2022visual} and Adapter~\cite{houlsby2019parameter} suffer significant accuracy drops when transferred to the 3D domain. For example, VPT~\cite{jia2022visual} leads to a 4.09\% decrease on the challenging PB\_T50\_RS variant compared to full fine-tuning. Similarly, although Adapter~\cite{houlsby2019parameter} achieves moderate gains on OBJ\_ONLY, it performs poorly on more complex tasks.

LST~\cite{sung2022lst} provides competitive performance in specific settings, but its generalization across variants is limited. In contrast, PLT achieves consistently strong results across various challenging scenarios, while tuning only 0.6M parameters, a fraction of the full model size.

A comparison with open-source alternatives in Tab.~\ref{tab:compare} further highlights our PLT's superiority under parameter-efficient fine-tuning (PEFT) settings. It outperforms all other PEFT methods, including the strongest 3D-specific methods like PointGST~\cite{liang2024parameter} and DAPT~\cite{zhou2024dynamic}, especially on the most difficult variant PB\_T50\_RS.

These findings highlight a fundamental limitation of many cross-domain PEFT methods: they often underperform in 3D settings due to domain gaps and architectural mismatches. Specifically, point cloud data is inherently unordered, sparse, and irregular in structure, characteristics not well captured by PEFT techniques originally designed for grid-like data in 2D vision or sequential data in NLP.

By contrast, our PLT is designed to explicitly address these 3D-specific challenges. It integrates hierarchical local feature extraction through HLN and incorporates both global and multi-scale local cues via adaptive fusion, enabling it to maintain high performance even in data-scarce or structurally complex scenarios. This design not only ensures strong generalization but also introduces a beneficial inductive bias tailored to 3D spatial structures. As a result, PLT emerges as a unified and effective PEFT solution for point cloud classification, striking an optimal balance between efficiency and accuracy.

\input{tab/compare}

\input{tab/origin_finetuning}

\subsection{Performance Analysis}
As summarized in Tab.~\ref{tab:performance}, PLT demonstrates strong performance across all dimensions, achieving top-tier accuracy with minimal parameters, competitive FLOPs, efficient inference and training speeds, and modest memory consumption. This highlights its suitability for real-world deployment where both effectiveness and efficiency are essential. Importantly, PLT achieves this without sacrificing the backbone architecture or requiring extensive architectural re-design, offering a plug-and-play solution compatible with powerful pretrained models like Point-MAE~\cite{pang2022masked}.

\input{tab/performance}

\subsection{Qualitative Analysis}

\paragraph{Attention Score Visualization}
Fig.~\ref{fig:attention_score} visualizes local attention weights $s_l$ across categories and layers, revealing three key insights: (1) Scores remain below 0.5, confirming dependence on pretrained global priors; (2) Early layers show strong category-specific variation that diminishes in deeper layers, indicating hierarchical fusion from local details to global semantics; (3) The decreasing attention from first to third layers demonstrates the network's progressive shift from local patterns to abstract concepts.
% These findings collectively validate the design of PLT as a lightweight yet powerful PEFT framework for 3D transfer learning, wherein carefully balanced architectural and fusion strategies contribute to superior accuracy and efficiency.

\input{fig/supplement/attention_score}

\paragraph{t-SNE Visualizations}  
To complement the t-SNE results discussed in the main text, Fig.~\ref{fig:tsne1} provides a comprehensive visualization of the feature manifolds learned by various fine-tuning strategies on the ScanObjectNN PB\_T50\_RS benchmark\cite{uy2019revisiting}, using t-SNE~\cite{van2008visualizing}. The evaluated strategies include full fine-tuning, linear probing, point-cloud-specific adaptation methods, and our proposed PLT approach. We extract the final-layer classification features and project them into a 2D space using t-SNE to facilitate visual interpretation. As illustrated in Fig.~\ref{fig:tsne1}, the features produced by PLT form more compact and well-separated clusters than those generated by both conventional and recent PEFT baselines. This suggests stronger class discrimination and improved feature organization. Importantly, PLT achieves these results with substantially fewer trainable parameters than full fine-tuning, underscoring its efficiency and effectiveness in adapting pre-trained models to downstream 3D recognition tasks.

\input{fig/tsne1}

\paragraph{Part Segmentation Visualizations}  
Fig.~\ref{fig:part_segmentation} provides a visual demonstration of the part segmentation results obtained using our proposed PLT, based on the PointMAE~\cite{pang2022masked} baseline. We select five representative samples from various categories and render three distinct perspectives for each sample. The figure clearly illustrates that our method delivers exceptional segmentation performance. Notably, the PLT method achieves high accuracy in distinguishing object parts with remarkable precision, while maintaining efficiency by using a minimal set of parameters.

\paragraph{Semantic Segmentation Visualizations}  
To complement the semantic segmentation visualization results discussed in the main text, Fig.~\ref{fig:s3dis_1} presents qualitative comparisons between our proposed PLT method and several existing fine-tuning approaches on Area 5 of the S3DIS dataset\cite{armeni20163d}. These results highlight PLT’s effectiveness in addressing key challenges in point cloud segmentation, particularly in preserving structural boundaries and reducing misclassifications. From the first row of the figure, it is evident that while full fine-tuning is resource-intensive, it struggles to accurately segment object boundaries. This often results in poor boundary delineation and misclassification in continuous regions, primarily due to insufficient capture of local point cloud information. A more detailed comparison in the office\_9 scene reveals that methods such as full fine-tuning, IDPT~\cite{zha2023instance}, and DAPT~\cite{zhou2024dynamic} incorrectly classify walls as beams and cluttered regions as ceilings. In contrast, our PLT method successfully segments clutter regions while significantly reducing the misclassification of walls. Similarly, in the office\_35 scene, PLT outperforms other methods in segmenting challenging areas such as the cluttered region and the board. Whereas methods like full fine-tuning, IDPT~\cite{zha2023instance}, and PPT~\cite{zhang2024positional} fail to accurately recognize or segment these regions. In the wc\_2 scene, most methods misclassify cluttered region into door, but our PLT can accurately distinguish them. These results emphasize the ability of the PLT method to deliver outstanding segmentation performance while using far fewer fine-tuning parameters. Compared to other approaches, PLT not only provides clearer segmentation boundaries but also significantly reduces misclassifications, further validating its efficiency and effectiveness for point cloud semantic segmentation tasks.

\subsection{Ablation Study}
%\subsubsection{Hyperparameter Ablation for Hierarchical Ladder Network}
\input{tab/ablation}
\paragraph{Hyperparameter Ablation for HLN}
We conduct an ablation study on key HLN hyperparameters: the number of neighbors $K$, feature dimensions $d$, and the number of hierarchical layers. As shown in Tab.~\ref{tab:k},~\ref{tab:dim} and~\ref{tab:layer}, using $K=[16,8,4]$, $d=[16,32,64,128]$ and a three-layer architecture yields the best performance, reaching the highest accuracy (85.53\%) with only 0.60M tunable parameters. This design reflects a hierarchical principle: as layers deepen and points are downsampled, decreasing $K$ preserves locality and avoids diluting fine-grained local cues and introducing noise from distant points. This adaptive design improves the model's ability to learn multi-resolution features across hierarchical levels, leading to superior performance in downstream tasks.

\paragraph{Token Selections for Head Inputs}  
To better understand the role of different token representations in downstream classification, we conduct an ablation study that systematically evaluates five input configurations for the classification head. Specifically, we investigate the impact of incorporating the class token $T_{cls}$, prompt tokens, patch tokens from the frozen backbone, and tokens from the HLN output. As illustrated in Fig.~\ref{fig:head}, each configuration corresponds to a different combination of these token sources.

Our results show that the best performance (85.53\% overall accuracy) is achieved when all four token types are utilized together (Fig.~\ref{fig:head}(a)), demonstrating the complementary nature of global semantics and local geometric cues. Interestingly, pooling the prompt tokens and fusing them with the class token, as opposed to directly concatenating all tokens, yields a performance improvement of 0.55\% (comparing Fig.~\ref{fig:head}(a) and (b)), while also reducing the number of trainable parameters, suggesting more efficient use of contextual information.

Furthermore, we observe that removing patch tokens leads to the most significant performance degradation (Fig.~\ref{fig:head}(e)), compared to the exclusion of either prompt tokens or HLN tokens. This finding underscores the crucial role of backbone-derived features, which encapsulate strong prior knowledge learned from large-scale pretraining, in ensuring robust downstream performance.

\input{fig/head}

\input{fig/supplement/part_segmentation}

\input{fig/supplement/semantic_segmentation}

%% file: tab/appendix/parames.tex
\begin{table}
\footnotesize
\captionof{table}{Training Details for Downstream Fine-tuning.}
% \vspace{-5pt}
\resizebox{\textwidth}{!}{
\begin{tabular}{lcccccc}
\toprule
    \multirow{2}{*}{Configuration}  &\multicolumn{3}{c}{Classification} & \multicolumn{2}{c}{Segmentation}\\
		\cmidrule(r){2-4} \cmidrule(r){5-6}
	 &ScanObjectNN & ModelNet & ModelNet Few-shot & ShapeNetPart & S3DIS     \\
    \midrule
 Optimizer & AdamW & AdamW & AdamW & AdamW & AdamW \\
 Learning rate & $5 \times 10^{-4}$ & $5 \times 10^{-4}$ & $5 \times 10^{-4}$  & $2 \times 10^{-4}$ & $2 \times 10^{-4}$ \\
 Weight decay & $5 \times 10^{-2}$ & $5 \times 10^{-2}$ & $5 \times 10^{-2}$ & $5 \times 10^{-2}$  & $5 \times 10^{-2}$ \\
 Learning rate scheduler & cosine & cosine & cosine & cosine & cosine \\
 Training epochs  & 300 & 300 & 150 & 300 & 60 \\
 Warmup epochs& 10 & 10& 10 & 10 &10 \\
 Batch size & 32 & 32& 32 & 16 & 32 \\
 \midrule
 Number of points  & 2048 & 1024& 1024 & 2048 & 2048 \\
 Number of point patches & 128 & 64 & 64 & 128 & 128\\
 Point patch size  & 32 & 32 & 32  & 32 & 32 \\
 \midrule
 number of layers & 3 & 3 & 3 & 3 & 3\\
 HLN dims & [16, 32, 64, 128] & [16, 32, 64, 128] & [16, 32, 64, 128] & [32, 64, 128, 256] & [32, 64, 128, 256]\\
 HLN strides & [4, 4, 4] & [4, 4, 4] & [4, 4, 4] & [4, 2, 2] & [4, 2, 2]\\
 number of neighbors in HLN & [16, 8, 4] & [16, 8, 4] & [16, 8, 4] & [32, 16, 8] & [32, 16, 8] \\
\bottomrule
% \vspace{10pt}
\end{tabular}
}
\label{tab:paramas}
\end{table}

%% file: tab/compare.tex
\begin{table}
\scriptsize
\centering
\caption{Comparisons of parameter efficient transfer learning methods from NLP and 2D Vision on the hardest variant of ScanObjectNN~\cite{uy2019revisiting}. Overall accuracy (\%) w/o voting is reported. \#TP represents the tunable parameters. \textcolor{red}{$^*$} denotes reproduced results.}
\label{tab:compare}
\begin{tabular}{ lcccc }
\toprule
 Method &Reference& Design for &\#TP (M) & PB\_T50\_RS \\
\midrule
 Point-MAE~\cite{pang2022masked}  &ECCV 22 & - & 22.1 & 85.18  \\
 Linear probing &- & - & 0.3& 75.99\\
 \midrule
  + Adapter~\cite{houlsby2019parameter}&ICML 19 & NLP & 0.9 & 83.93 \\
  + Perfix tuning~\cite{li2021prefix}& ACL 21 & NLP &0.7 & 77.72  \\
  + BitFit~\cite{zaken2022bitfit} & ACL 21 & NLP &0.3 & 82.62    \\
  + LST~\cite{sung2022lst} & NeurIPS 22 & NLP & 0.8 & 82.75\\
  + LoRA~\cite{hu2021lora} & ICLR 22 & NLP & 0.9&  81.74   \\
  % + DEPT~\cite{shi2024dept} & ICLR 24 & NLP & 0.3 & 79.70\\
  + FourierFT~\cite{Gao2024Fourier} & ICML 24 & NLP &0.3 & 78.57\\
  \midrule
  + VPT-Deep~\cite{jia2022visual}&ECCV 22 & 2D &0.4 &  81.09 \\
  + AdaptFormer~\cite{chen2022adaptformer} &NeurIPS 22 & 2D &0.9  & 83.45 \\
  + SSF~\cite{lian2022scaling} & NeurIPS 22 & 2D &0.4  & 82.58\\
  + FacT~\cite{jie2023fact} & AAAI 23 & 2D & 0.5 & 78.76\\
  % + BI-AdaptFormer~\cite{jie2023revisiting} & ICCV 23 & 2D & 0.4 & 83.66\\
  + SCT~\cite{zhao2024sct} & IJCV 24 & 2D & 0.3 & 80.40\\
  \midrule
  + IDPT~\cite{zha2023instance} &ICCV 23 & 3D & 1.7 &84.94\\
  + Point-PEFT\textcolor{red}{$^*$}~\cite{tang2024point} & AAAI 24 & 3D & 0.7 &84.35\\
  + DAPT~\cite{zhou2024dynamic} & CVPR 24 & 3D & 1.1 & 85.08 \\
  + PPT\textcolor{red}{$^*$}~\cite{zhang2024positional} & ACM MM 25 & 3D & 1.1 & 84.91\\
  %+ PMA~\cite{zha2025pma} & CVPR 25 & 3D & 1.1 & \textbf{86.43}\\
  + PointGST~\cite{liang2024parameter} & TPAMI 25 & 3D & 0.6 & 85.29\\
  + PointLoRA~\cite{wang2025pointlora} & CVPR 25 & 3D & 0.8 & 85.53\\
  \rowcolor{linecolor!40}+ PLT (\textbf{ours}) & - & 3D & 0.6 & \textbf{85.53} \\
\bottomrule
\end{tabular}
\end{table}

%% file: tab/origin_finetuning.tex
\begin{table}
\footnotesize
\setlength{\tabcolsep}{1.6mm}
\centering
\caption{The overall accuracy (\%) for classical fine-tuning strategies on three variants of ScanObjectNN~\cite{uy2019revisiting} is reported. \#TP means the number of tunable parameters. Linear probing indicates head-tuned only.}
\label{tab:origin_finetuning}
\begin{tabular}{ lcccc }
\toprule
Tuning Strategy & \#TP(M) &OBJ\_BG &OBJ\_ONLY & PB\_T50\_RS \\
\midrule
Point-MAE~\cite{pang2022masked} & 22.1 & 90.02 & 88.29 & 85.18 \\
Linear probing & 0.3  & 87.26\dtplus{-2.76} & 84.85\dtplus{-3.44} & 75.99\dtplus{-9.19}\\
\midrule
+ Adapter~\cite{houlsby2019parameter} & 0.9  & 89.50\dtplus{-0.52} & 88.64\dplus{+0.35} & 83.93\dtplus{-1.25}\\
+ VPT~\cite{jia2022visual} & 0.4  & 87.26\dtplus{-2.76} & 87.09\dtplus{-1.20} &81.09\dtplus{-4.09} \\
+ LST~\cite{sung2022lst} & 0.8  & 89.67\dtplus{-0.25} & 89.67\dplus{+1.38} &82.75\dtplus{-2.43} \\
\bottomrule
\end{tabular}
\vspace{-12pt}
\end{table}

%% file: tab/performance.tex
\begin{table*}
  \centering
  \scriptsize
  \caption{Comparison of performance between our PLT and previous methods. We conduct experiments on the hardest variant (i.e., PB\_T50\_RS) of ScanObjectNN~\cite{uy2019revisiting} with Point-MAE baseline~\cite{pang2022masked} when the batch size is 32. Throughput (samples/s) is measured on single RTX 3090 GPU.}
  \resizebox{\textwidth}{!}{
    \begin{tabular}{lccccccc}
    \toprule
    Methods & Parameter (M) & FLOPs (G) & Train Throughput (samples/s) & Train Memory (G) & Infer Speed (samples/s) & Infer Memory (G) & Acc. (\%) \\
    \midrule
    PointMAE~\cite{pang2022masked} & 22.09 & 4.93 & 151.2 & 5.24 & 232.7 & 0.85 & 85.18 \\
    +IDPT~\cite{zha2023instance} & 1.70 & 7.27 & 126.3 & 5.10 & 204.3 & 0.96 & 84.94 \\
    +DAPT~\cite{zhou2024dynamic} & 1.09 & 5.14 & 169.0 & 2.26 & 218.7 & 0.85 & 85.08 \\
    +PointPEFT~\cite{tang2024point} & 0.77 & - & 48.5 & 8.64 & 65.6 & 1.12 & 84.35 \\
    +PointGST~\cite{liang2024parameter} & 0.62 & 4.99 & 146.5 & 1.02 & 173.7 & 0.85 & 85.29 \\
    +LST~\cite{sung2022lst} & 0.76 & 4.98 & 192.1 & 1.06 & 231.4 & 0.85 & 82.75 \\
    \rowcolor{linecolor!40}+PLT (\textbf{ours}) & 0.60 & 5.02 & 165.4 & 2.26 & 210.8 & 0.85 & \textbf{85.53}\\
    \bottomrule
    \end{tabular}
    }
  \label{tab:performance}
\end{table*}

%% file: fig/supplement/attention_score.tex
\begin{figure}
    \centering
    \includegraphics[width=0.8\linewidth]{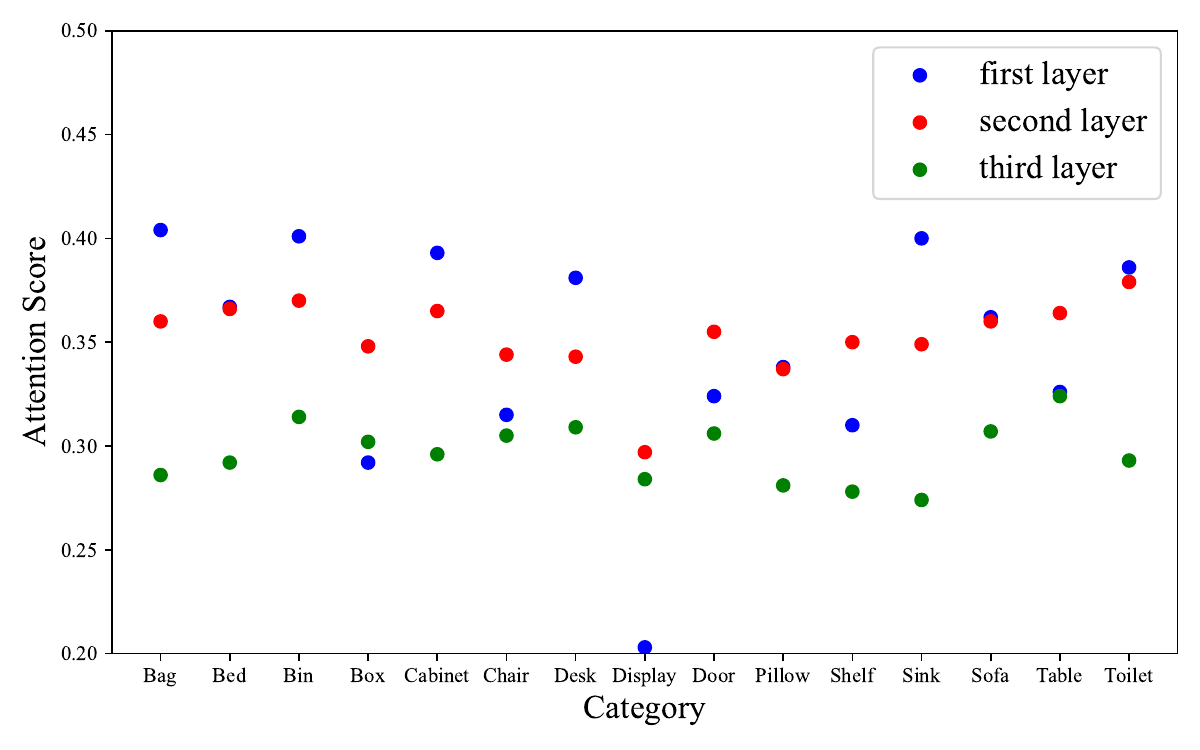}
    \caption{Local information attention score $s_l$ of LGF. We conduct experiments on ScanObjectNN (PB T50 RS)~\cite{uy2019revisiting} and calculate the average attention score of local information for each category.
}
    \label{fig:attention_score}
    % \vspace{-10pt}
\end{figure}

%% file: fig/tsne1.tex
\begin{figure}[ht]
    \centering
    \begin{subfigure}{0.24\textwidth}
        \centering
        \includegraphics[width=\linewidth]{fig/tsne/point_mae.pdf}
        \caption*{\textbf{\#TP}:22M \textbf{\#OA}:85.18}
        \caption{Full fine-tuning}
        \label{fig:sub1_1}
    \end{subfigure}
    \hfill
    \begin{subfigure}{0.24\textwidth}
        \centering
        \includegraphics[width=\linewidth]{fig/tsne/LP.pdf}
        \caption*{\textbf{\#TP}:0.3M \textbf{\#OA}:75.99}
        \caption{Linear Probing}
        \label{fig:sub1_2}
    \end{subfigure}
    \hfill
    \begin{subfigure}{0.24\textwidth}
        \centering
        \includegraphics[width=\linewidth]{fig/tsne/idpt.pdf}
        \caption*{\textbf{\#TP}:1.7M \textbf{\#OA}:84.94}
        \caption{IDPT}
        \label{fig:sub1_3}
    \end{subfigure}
    \hfill
    \begin{subfigure}{0.24\textwidth}
        \centering
        \includegraphics[width=\linewidth]{fig/tsne/dapt.pdf}
        \caption*{\textbf{\#TP}:1.1M \textbf{\#OA}:85.08}
        \caption{DAPT}
        \label{fig:sub1_4}
    \end{subfigure}
    \hfill
    \begin{subfigure}{0.24\textwidth}
        \centering
        \includegraphics[width=\linewidth]{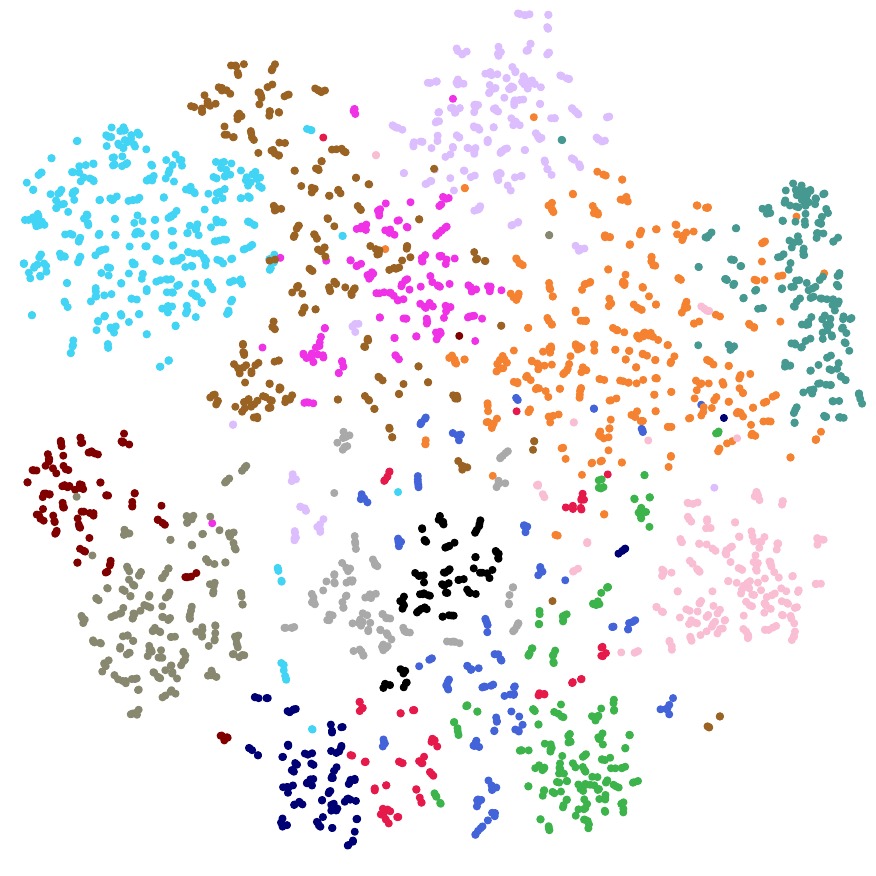}
        \caption*{\textbf{\#TP}:0.6M \textbf{\#OA}:85.29}
        \caption{PointGST}
        \label{fig:sub1_5}
    \end{subfigure}
    \hfill
    \begin{subfigure}{0.24\textwidth}
        \centering
        \includegraphics[width=\linewidth]{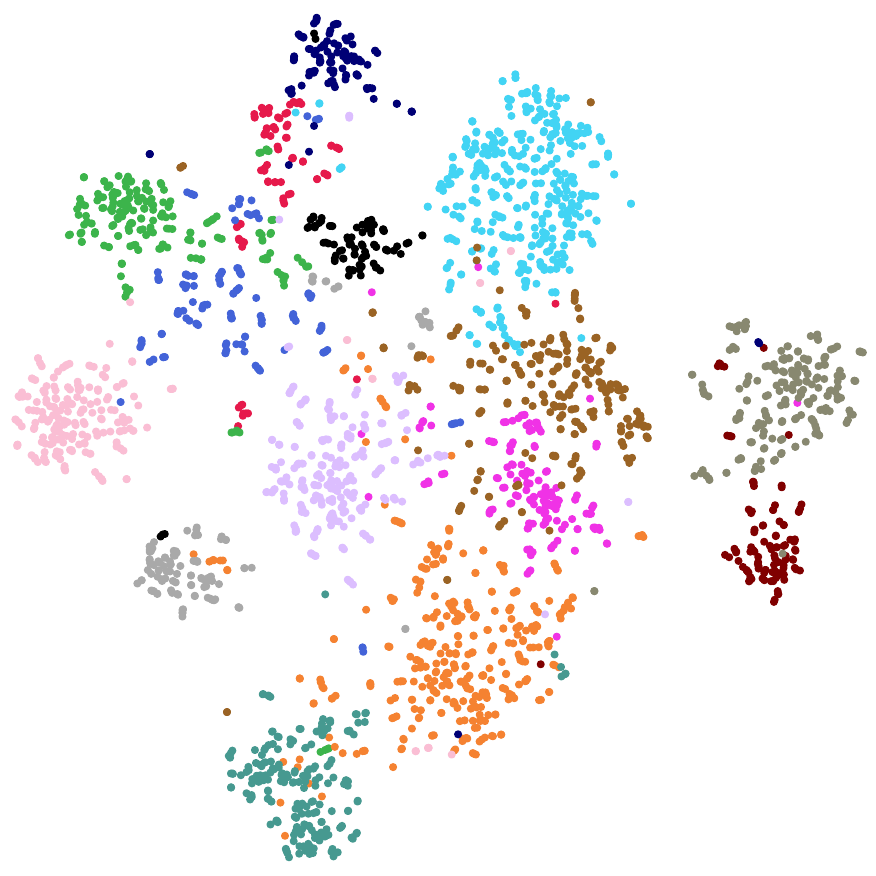}
        \caption*{\textbf{\#TP}:1.1M \textbf{\#OA}:84.91}
        \caption{PPT}
        \label{fig:sub1_6}
    \end{subfigure}
    \hfill
    \begin{subfigure}{0.24\textwidth}
        \centering
        \includegraphics[width=\linewidth]{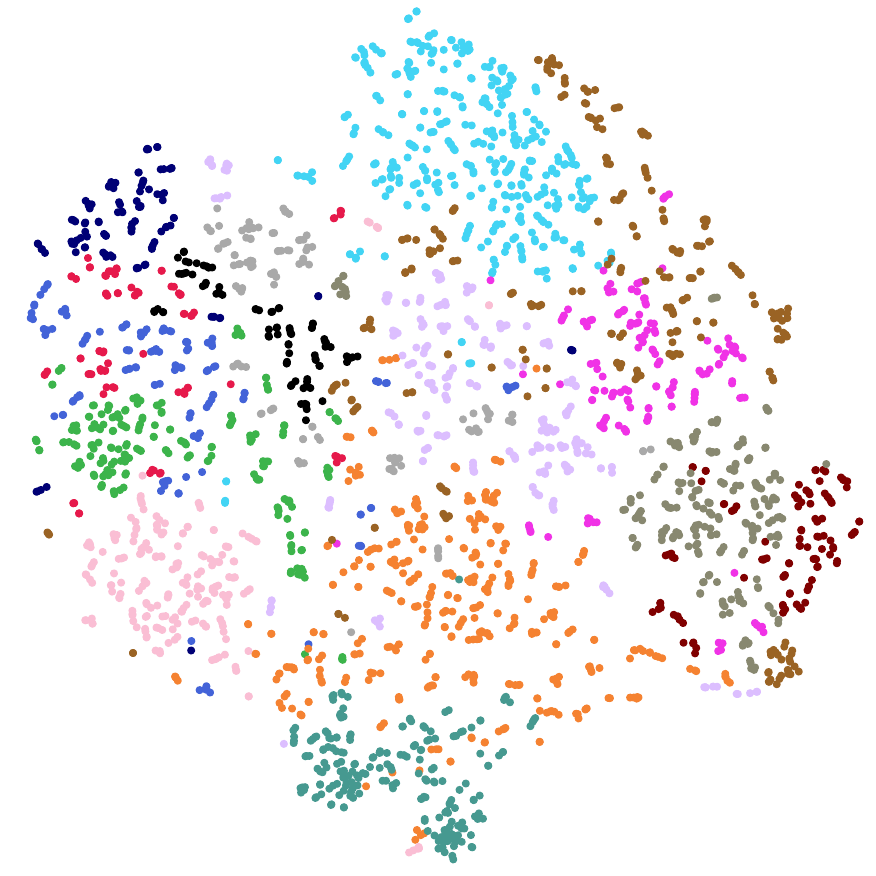}
        \caption*{\textbf{\#TP}:0.8M \textbf{\#OA}:82.75}
        \caption{LST}
        \label{fig:sub1_7}
    \end{subfigure}
    \hfill
    \begin{subfigure}{0.24\textwidth}
        \centering
        \includegraphics[width=\linewidth]{fig/tsne/point_ladder.pdf}
        \caption*{\textbf{\#TP}:0.6M \textbf{\#OA}:85.53}
        \caption{PLT (Ours)}
        \label{fig:sub1_8}
    \end{subfigure}
    \caption{The visualization of the t-SNE~\cite{van2008visualizing} from the test sets of ScanObjectNN~\cite{uy2019revisiting} (PB\_T50\_RS) by using a pre-trained PointMAE~\cite{pang2022masked} with various fine-tuning strategies. We extract the final classification features from the top linear layer for t-SNE visualizations.}
    \label{fig:tsne1}
\end{figure}

%% file: tab/ablation.tex
\begin{table}[h]
    \centering
    \caption{Ablation on $K$ in HLN.}
    \label{tab:k}
    \begin{tabular}{ccc}
       \toprule
        $K$ & \#TP (M) & PB\_T50\_RS \\
        \midrule
        \text{[16, 16, 16]} & 0.60 & 84.46 \\
        \text{[4, 4, 4]} & 0.60 & 84.66 \\
        \text{[4, 8, 16]} & 0.60 & 84.91 \\
        \rowcolor{linecolor!40} \text{[16, 8, 4]} & 0.60 & \textbf{85.53} \\
        \bottomrule
    \end{tabular}
\end{table}

\vspace{-10pt}

\begin{table}[h]
    \centering
    \caption{Ablation on $d$ in HLN.}
    \label{tab:dim}
    \begin{tabular}{ccc}
        \toprule
        Feature Dim & \#TP (M) & PB\_T50\_RS \\
        \midrule
        \text{[8, 16, 32, 64]} & 0.46 & 83.17 \\
        \rowcolor{linecolor!40} \text{[16, 32, 64, 128]} & 0.60 & \textbf{85.53} \\
        \text{[32, 64, 128, 256]} & 1.00 & 84.25 \\
        \bottomrule
    \end{tabular}
\end{table}

\vspace{-10pt}

\begin{table}[h]
    \centering
    \caption{Ablation on the num of layers.}
     \begin{tabular}{ccc}
        \toprule
        Layer num & \#TP (M) & PB\_T50\_RS \\
        \midrule
        1 & 0.40 & 83.55 \\
        2 & 0.45 & 84.84 \\
        \rowcolor{linecolor!40}3 & 0.60 & \textbf{85.53} \\
        4 & 1.09 & 85.05 \\
        \bottomrule
    \end{tabular}
    \label{tab:layer}
\end{table}

%% file: fig/head.tex
\begin{figure*}
    \centering
    \includegraphics[width=1.02\linewidth]{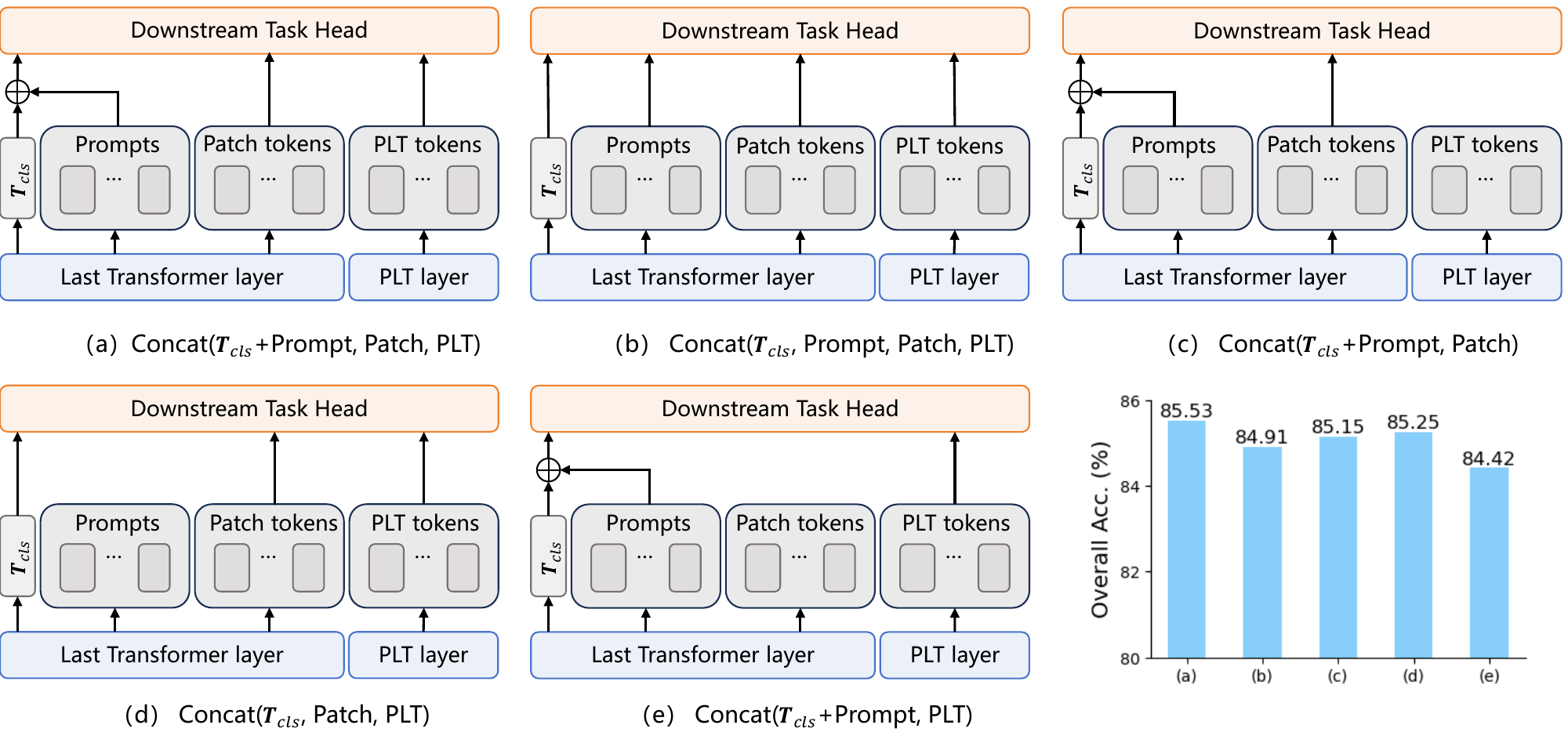}
    \caption{The Impact of Input Variations on Downstream Classification Performance. We conduct experiments on the hardest variant (i.e., PB\_T50\_RS) of ScanObjectNN~\cite{uy2019revisiting} with Point-MAE baseline~\cite{pang2022masked}.}
    \label{fig:head}
\end{figure*}

%% file: fig/supplement/part_segmentation.tex
\begin{figure*}[htbp]
    \centering

    % 第一行左侧的竖排标签
    \begin{minipage}{0.1\textwidth}
        \centering
        {airplane}
        % \rotatebox{90}{14}
    \end{minipage}
    \hfill
    % 第一行图片
    \begin{minipage}{0.25\textwidth}
        \centering
        \includegraphics[width=\textwidth]{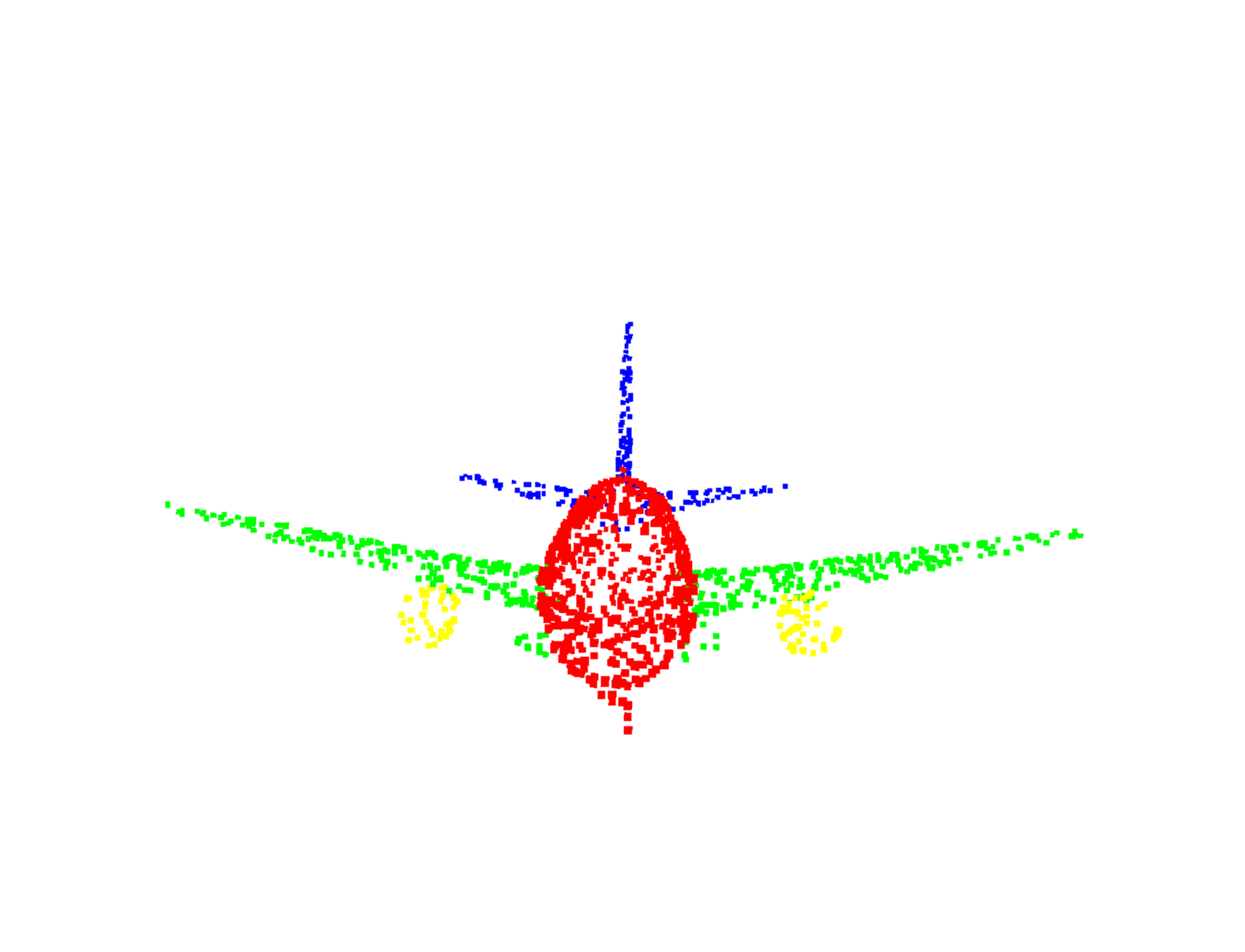} % 替换为你的图片路径
    \end{minipage}
    \hfill
    \begin{minipage}{0.25\textwidth}
        \centering
        \includegraphics[width=\textwidth]{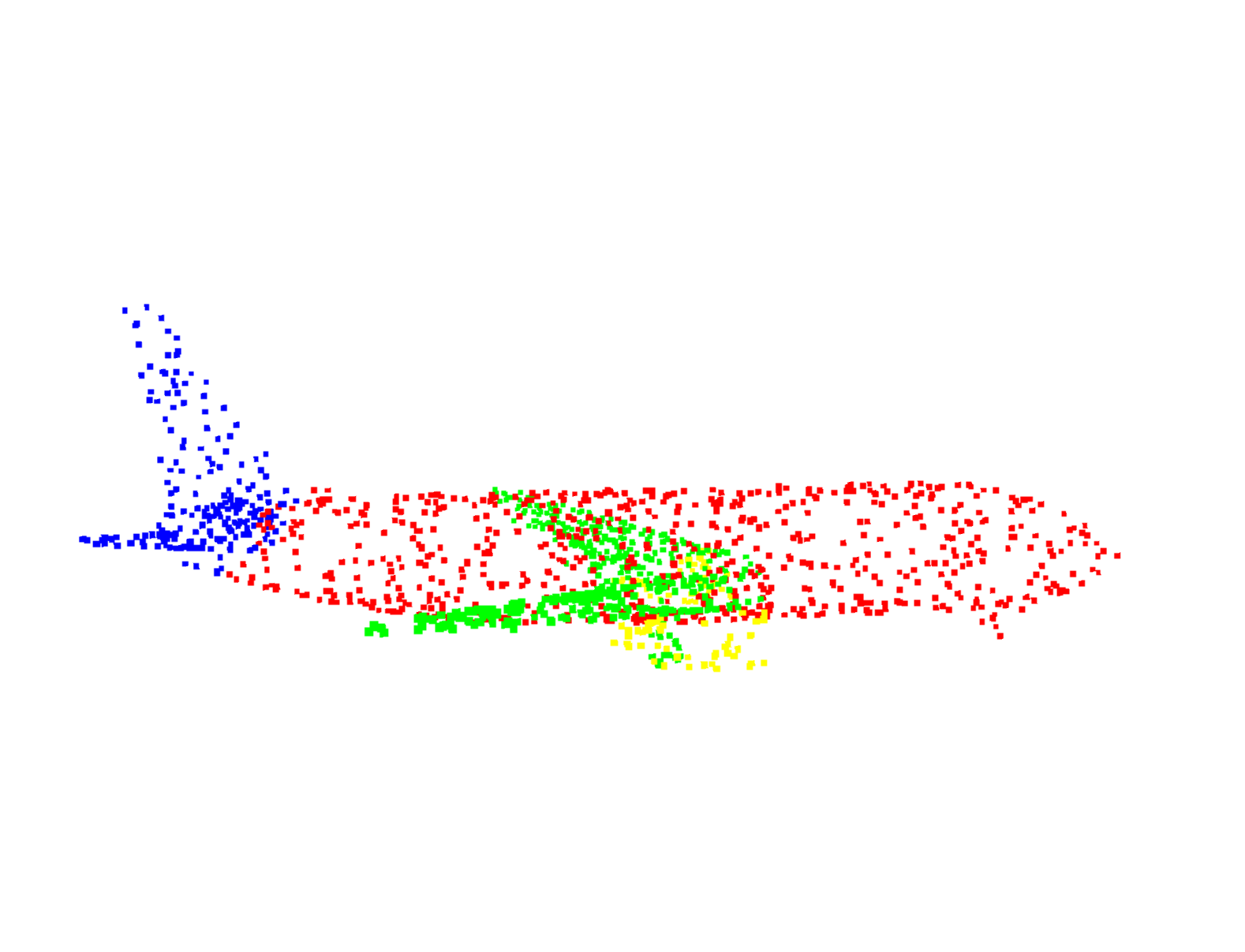}
    \end{minipage}
    \hfill
    \begin{minipage}{0.25\textwidth}
        \centering
        \includegraphics[width=\textwidth]{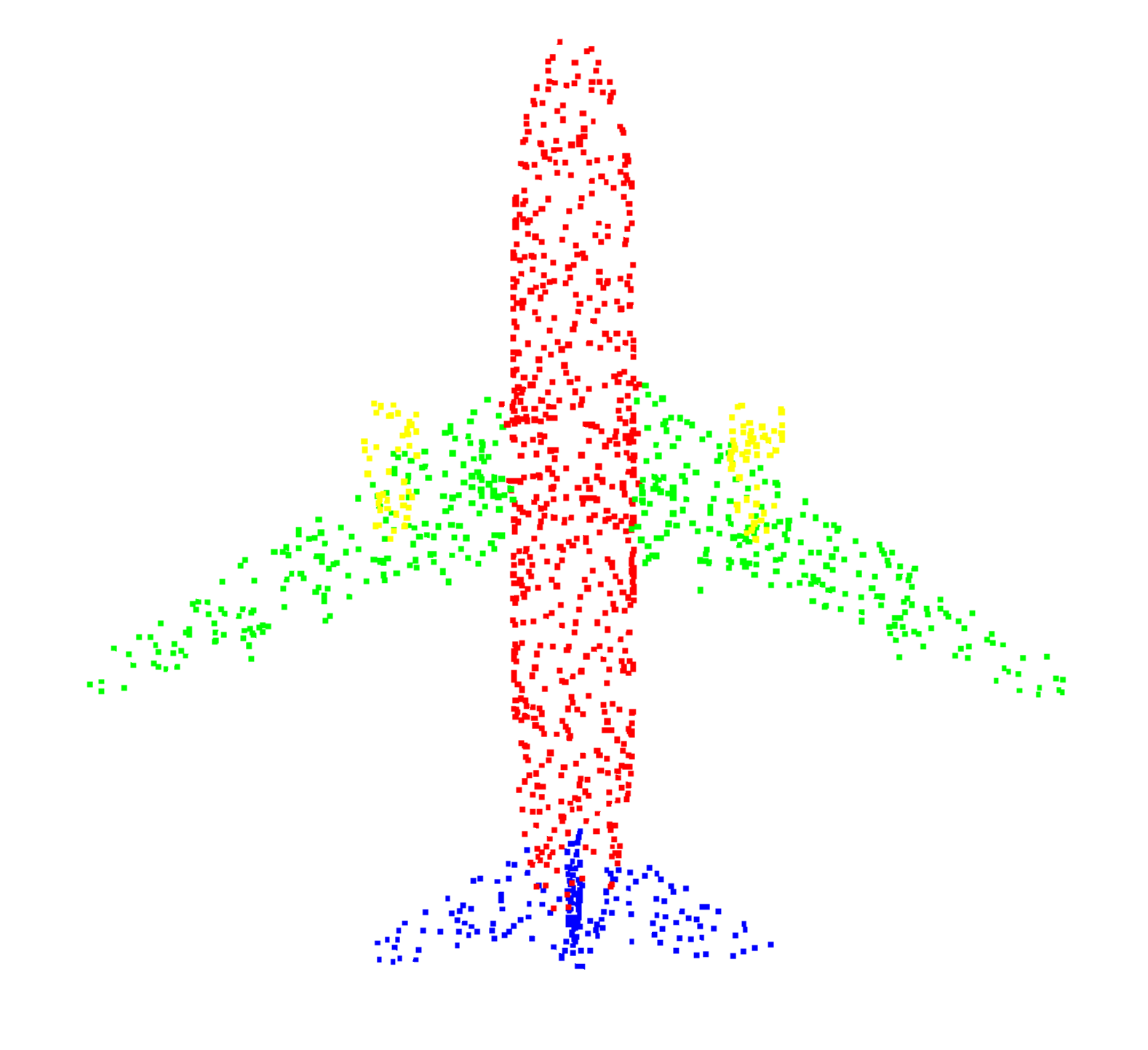}
    \end{minipage}
    \hfill

    % 换行
    \vspace{0.5em}

    % 第二行左侧的竖排标签
    \begin{minipage}{0.1\textwidth}
        \centering
        {chair}
    \end{minipage}
    \hfill
    % 第二行图片
    \begin{minipage}{0.25\textwidth}
        \centering
        \includegraphics[width=\textwidth]{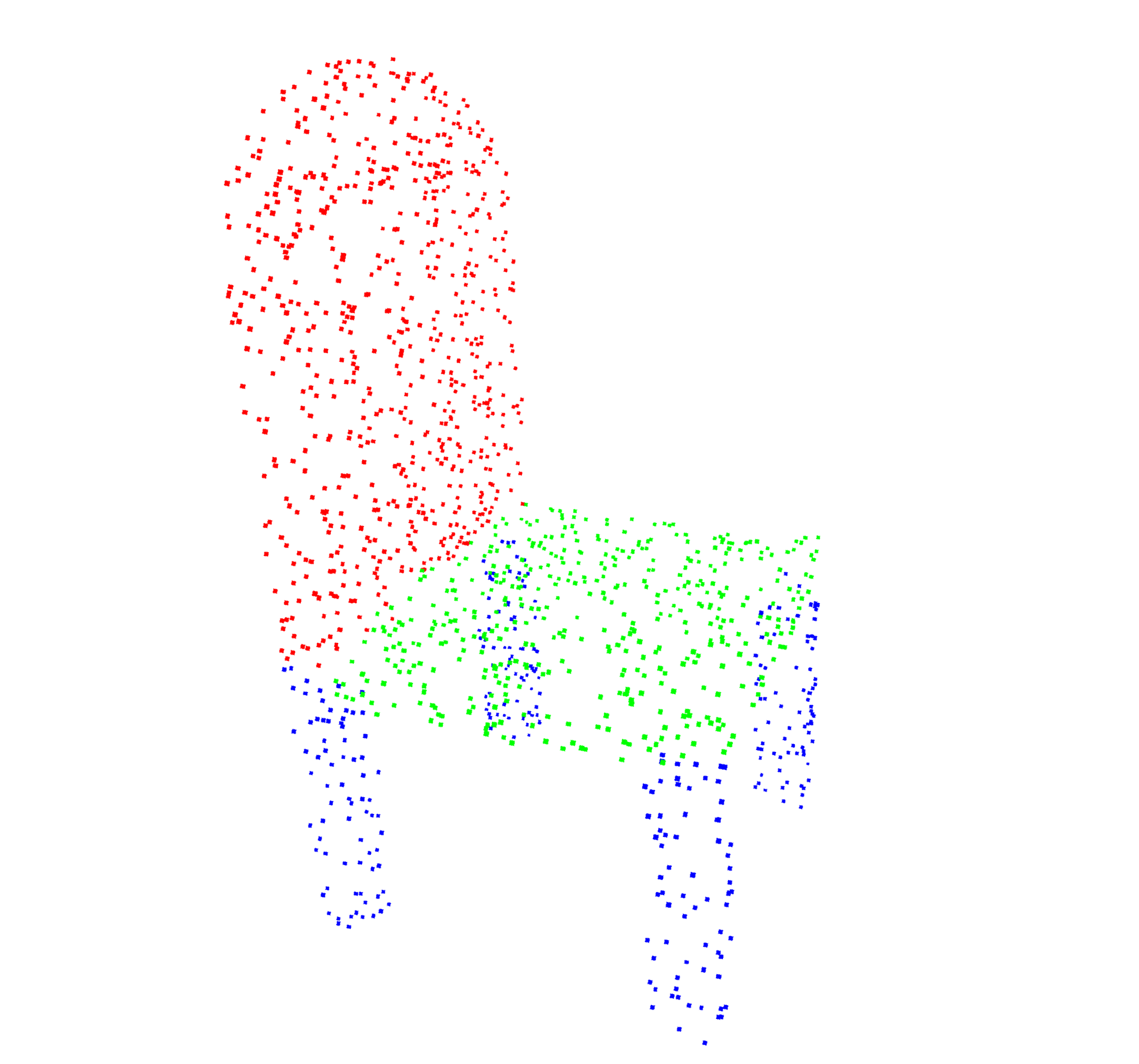}
    \end{minipage}
    \hfill
    \begin{minipage}{0.25\textwidth}
        \centering
        \includegraphics[width=\textwidth]{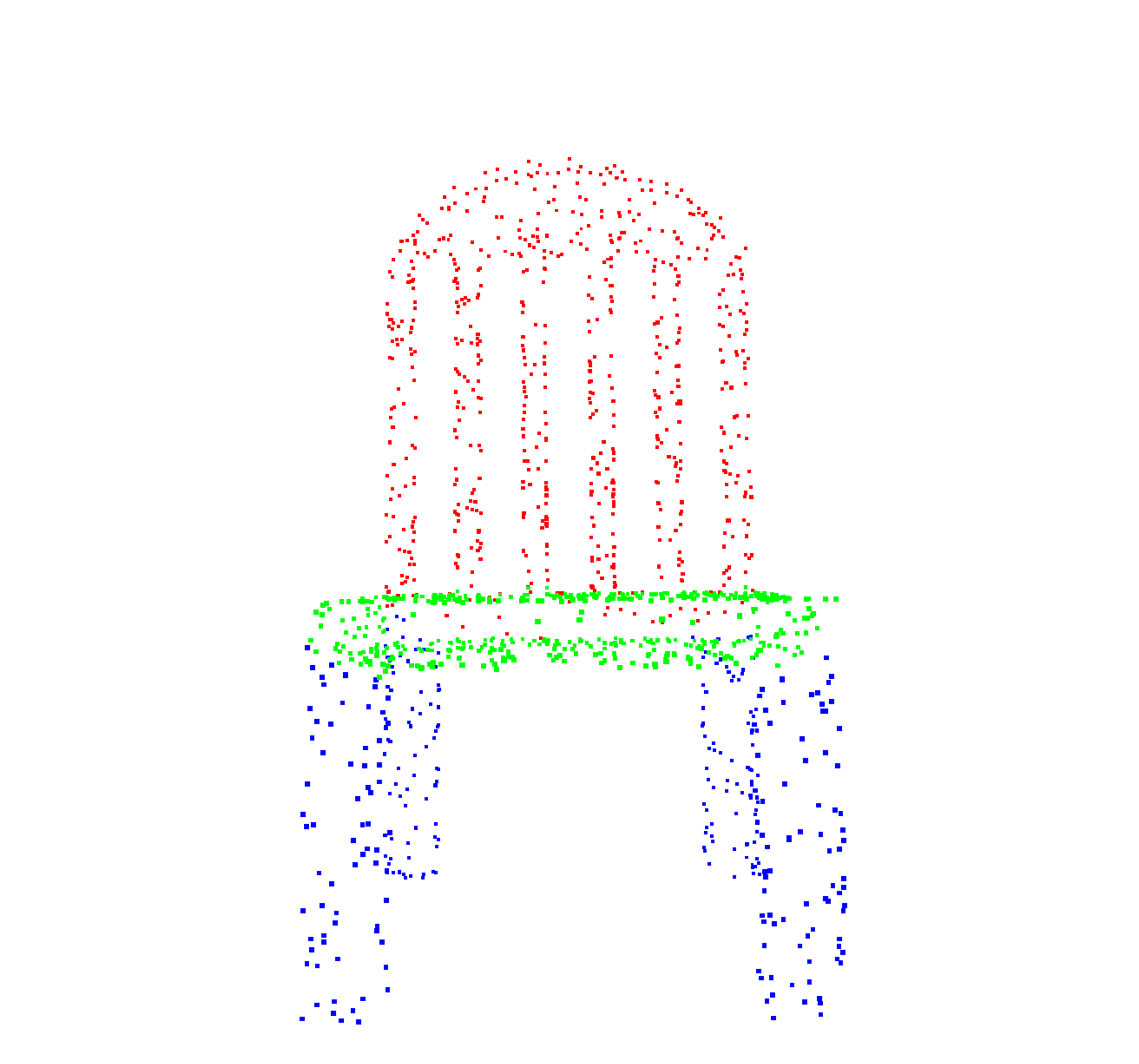}
    \end{minipage}
    \hfill
    \begin{minipage}{0.25\textwidth}
        \centering
        \includegraphics[width=\textwidth]{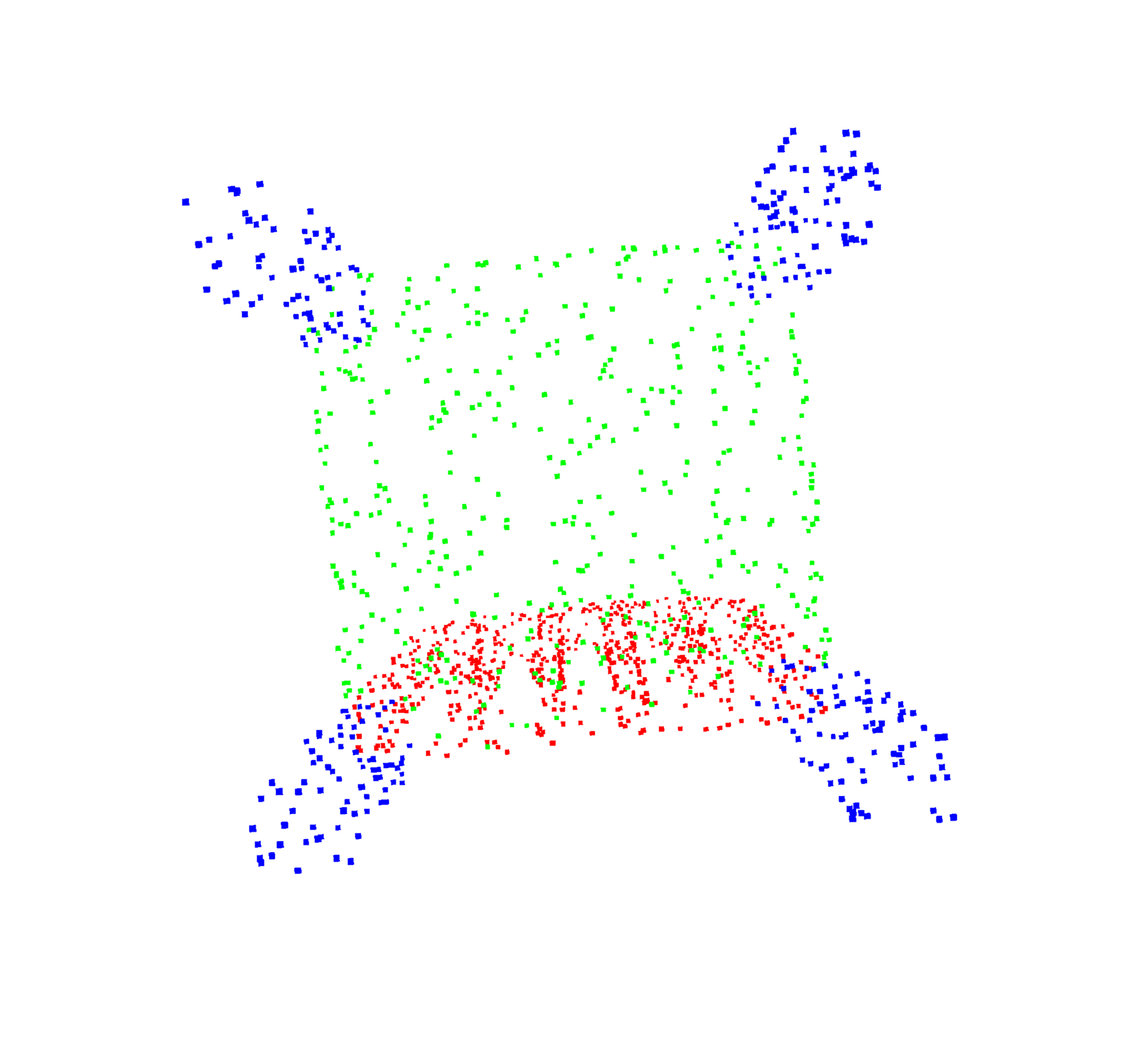}
    \end{minipage}
    \hfill

    % 换行
    \vspace{0.5em}

    % 第三行左侧的竖排标签
    \begin{minipage}{0.1\textwidth}
        \centering
        {lamp}
    \end{minipage}
    \hfill
    % 第三行图片
    \begin{minipage}{0.25\textwidth}
        \centering
        \includegraphics[width=\textwidth]{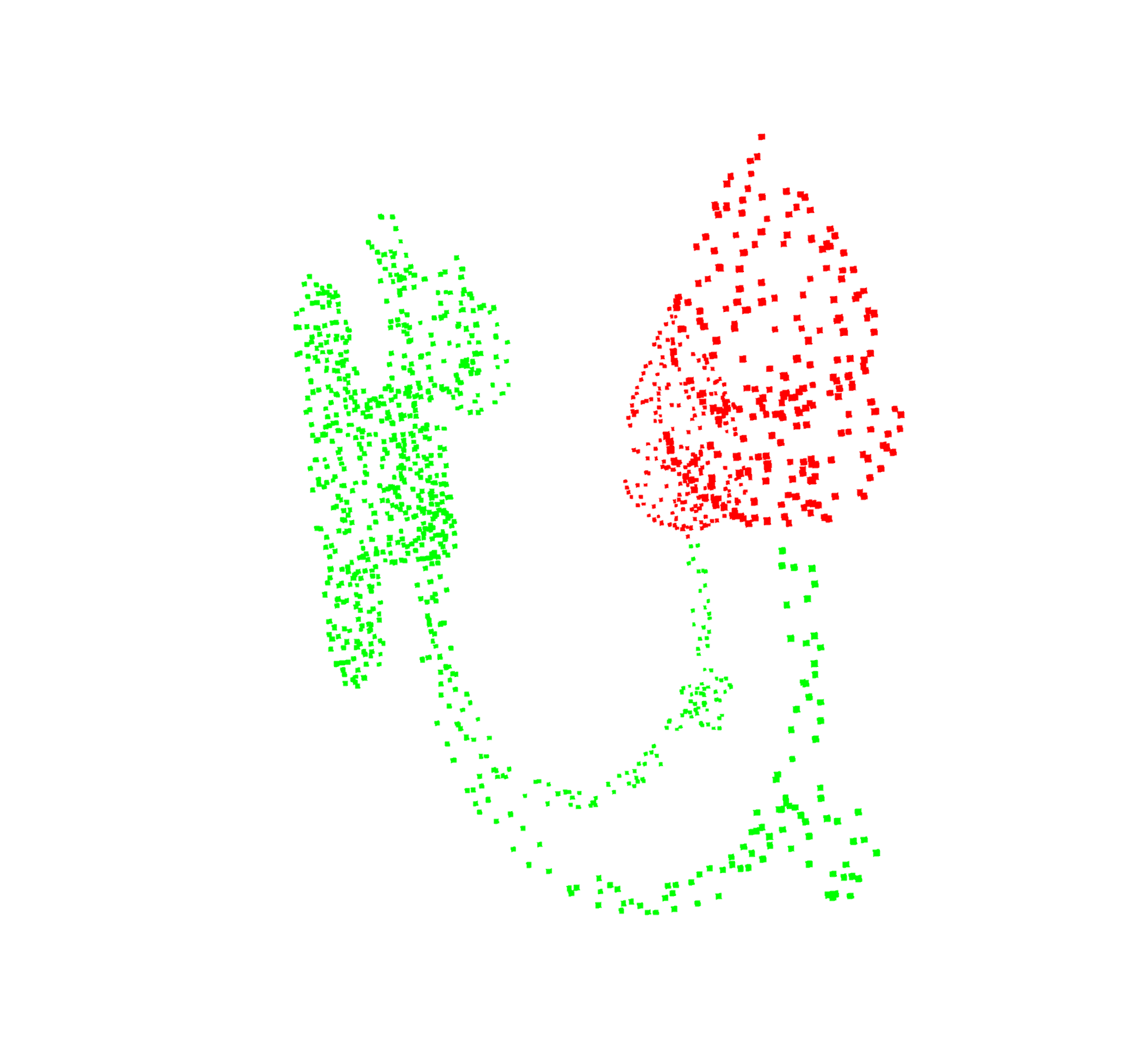}
    \end{minipage}
    \hfill
    \begin{minipage}{0.25\textwidth}
        \centering
        \includegraphics[width=\textwidth]{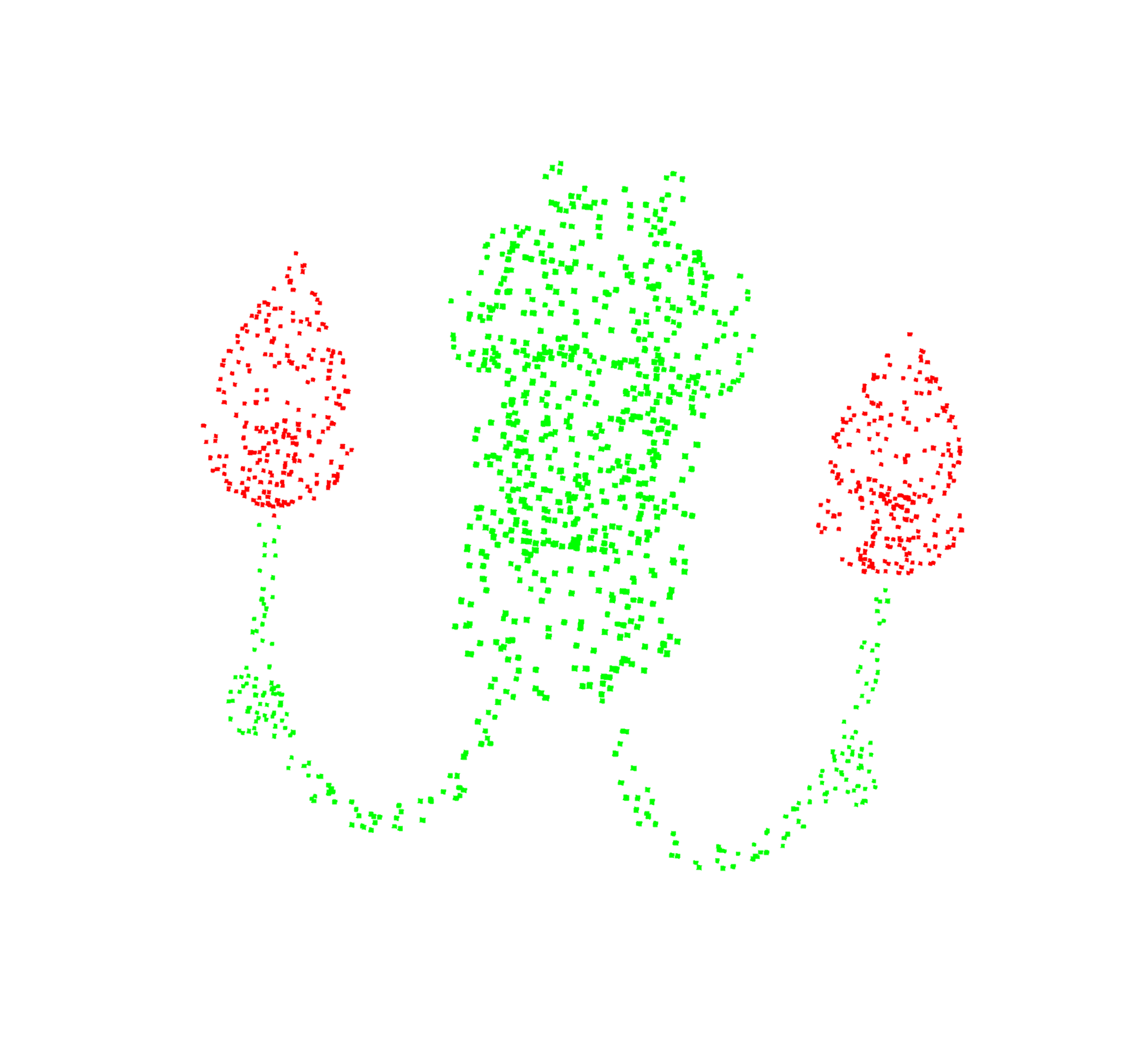}
    \end{minipage}
    \hfill
    \begin{minipage}{0.25\textwidth}
        \centering
        \includegraphics[width=\textwidth]{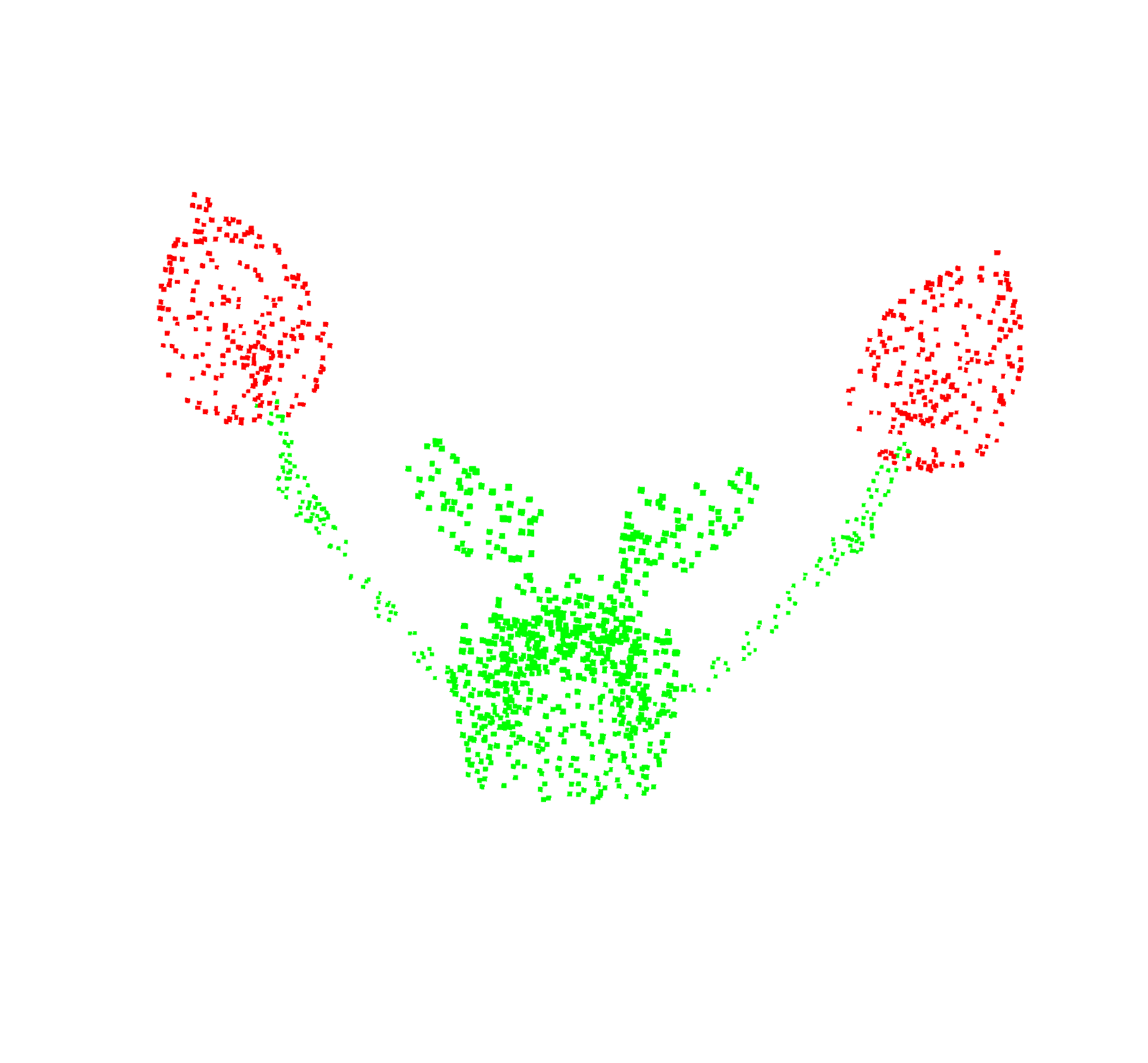}
    \end{minipage}
    \hfill

    % 换行
    \vspace{0.5em}

    % 第四行左侧的竖排标签
    \begin{minipage}{0.1\textwidth}
        \centering
        {skateboard}
    \end{minipage}
    \hfill
    % 第四行图片
    \begin{minipage}{0.25\textwidth}
        \centering
        \includegraphics[width=\textwidth]{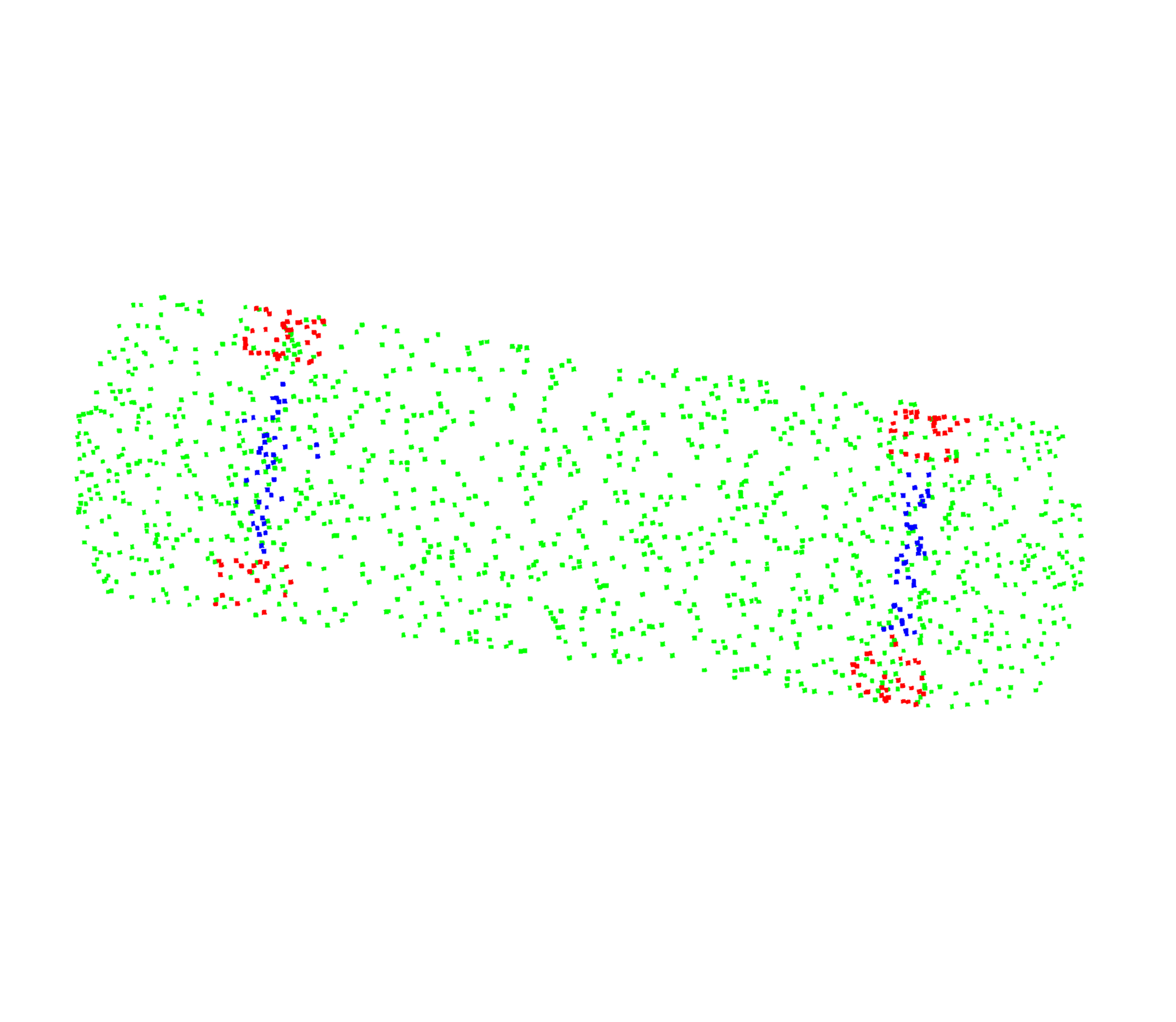}
    \end{minipage}
    \hfill
    \begin{minipage}{0.25\textwidth}
        \centering
        \includegraphics[width=\textwidth]{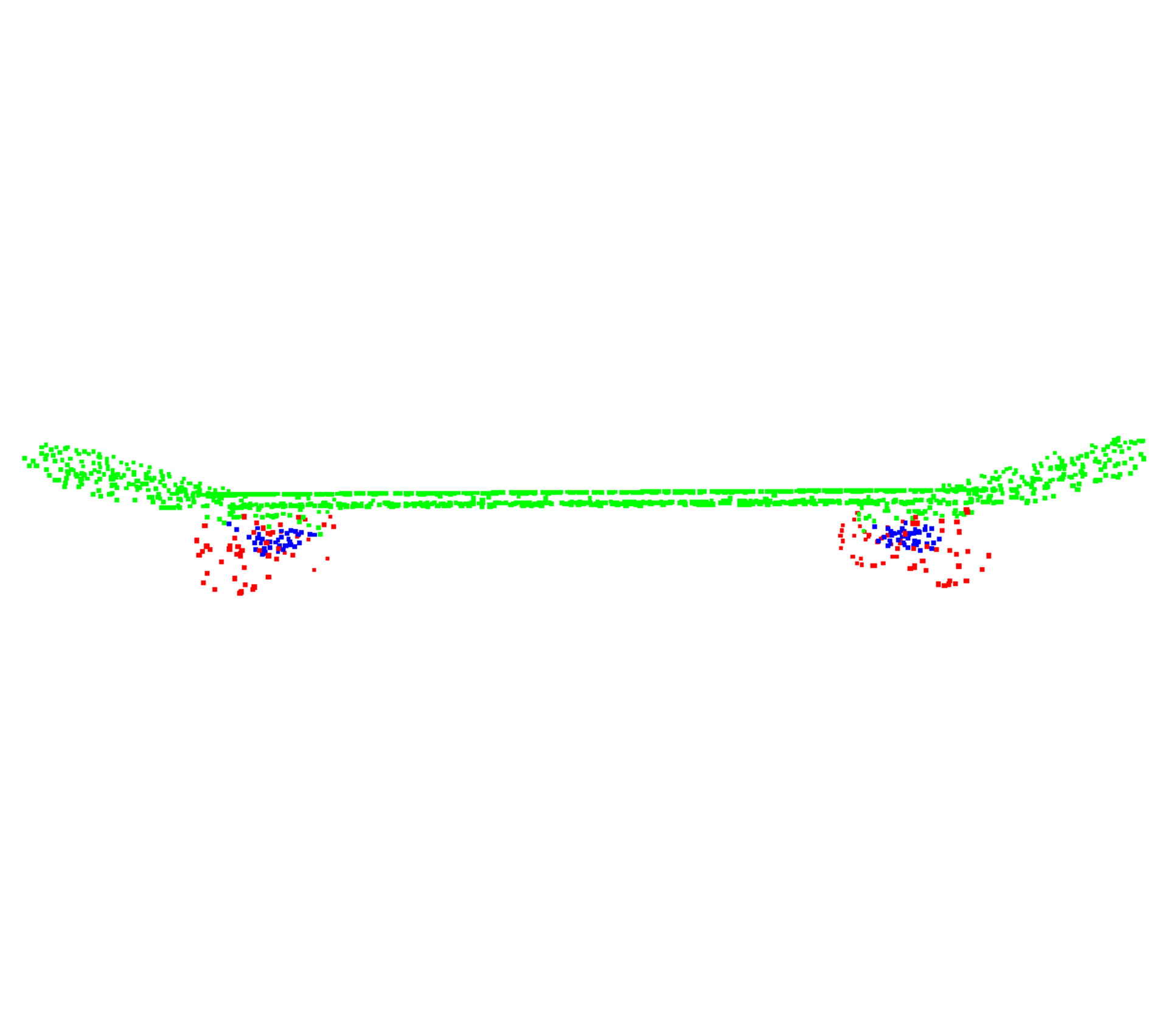}
    \end{minipage}
    \hfill
    \begin{minipage}{0.25\textwidth}
        \centering
        \includegraphics[width=\textwidth]{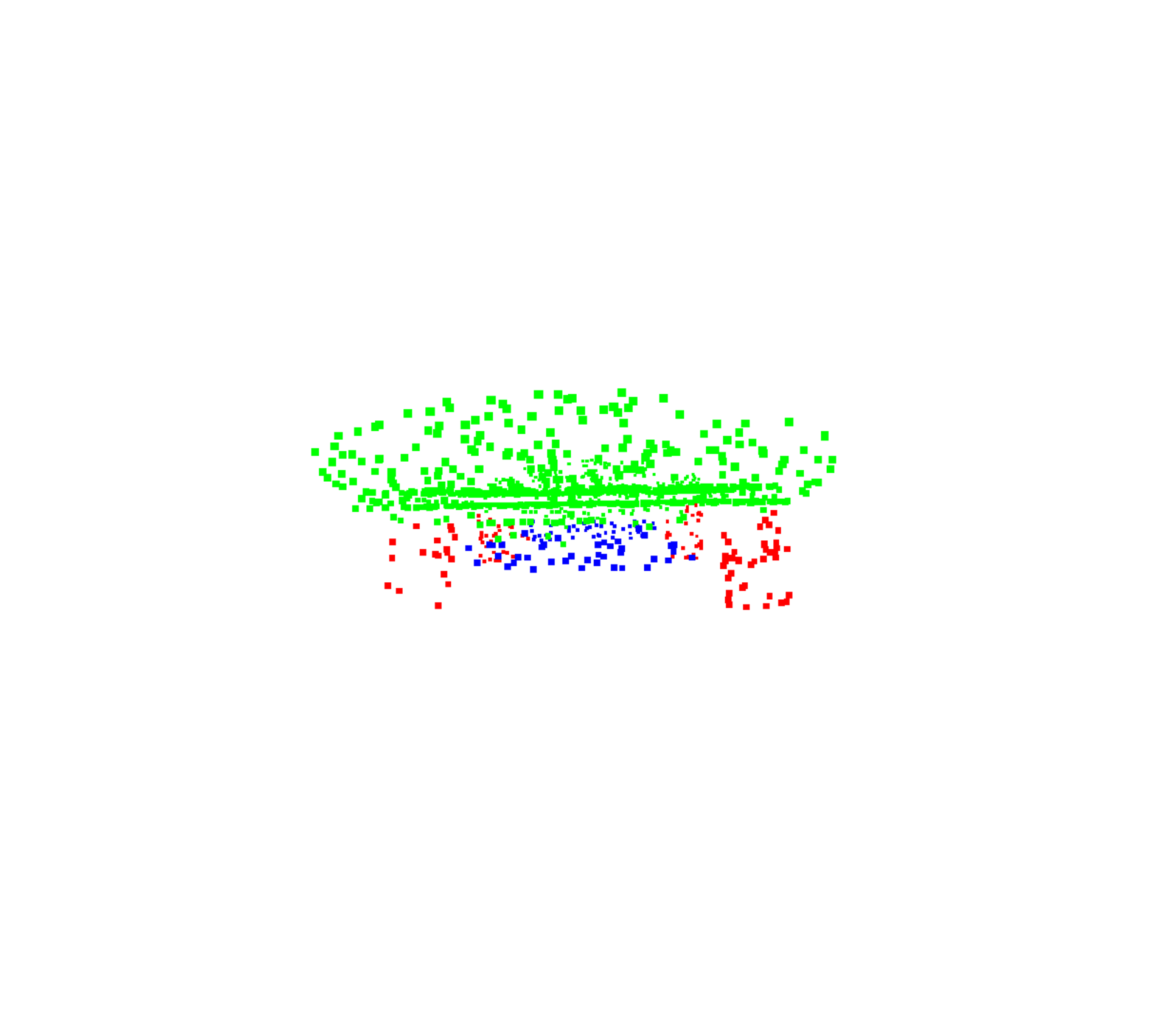}
    \end{minipage}
    \hfill

    % 换行
    \vspace{0.5em}

    % 第五行左侧的竖排标签
    \begin{minipage}{0.1\textwidth}
        \centering
        {table}
    \end{minipage}
    \hfill
    % 第五行图片
    \begin{minipage}{0.25\textwidth}
        \centering
        \includegraphics[width=\textwidth]{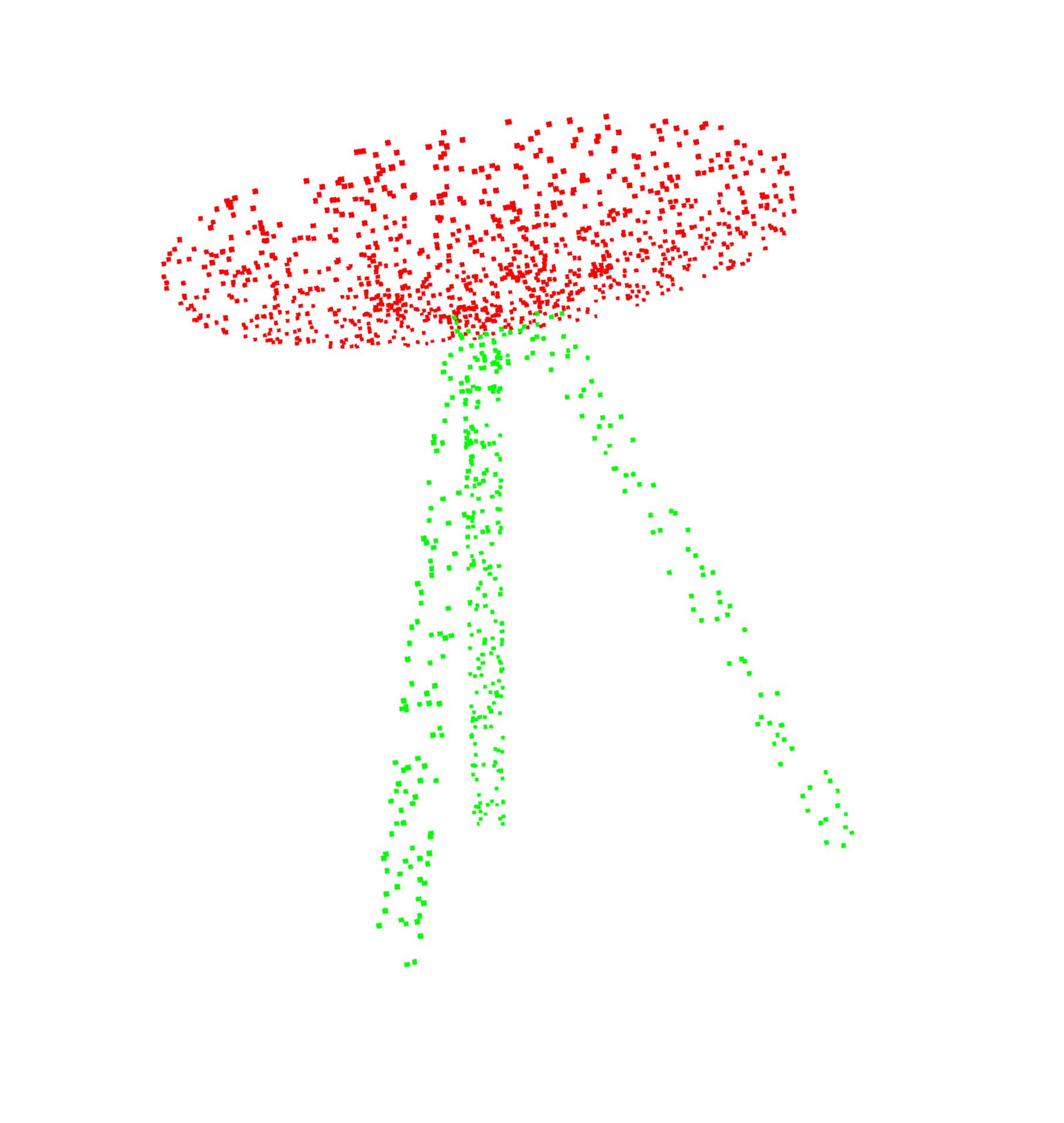}
    \end{minipage}
    \hfill
    \begin{minipage}{0.25\textwidth}
        \centering
        \includegraphics[width=\textwidth]{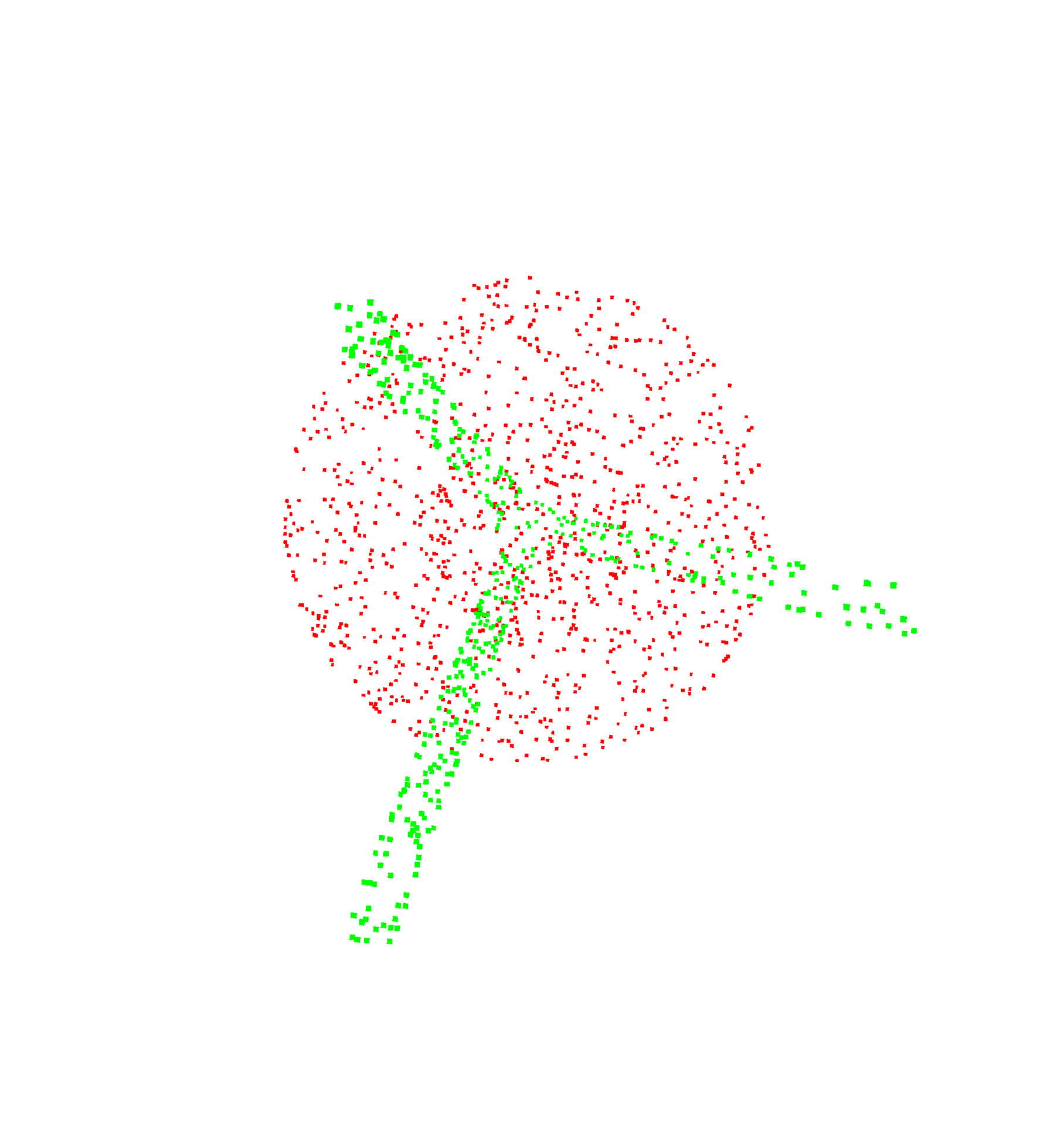}
    \end{minipage}
    \hfill
    \begin{minipage}{0.25\textwidth}
        \centering
        \includegraphics[width=\textwidth]{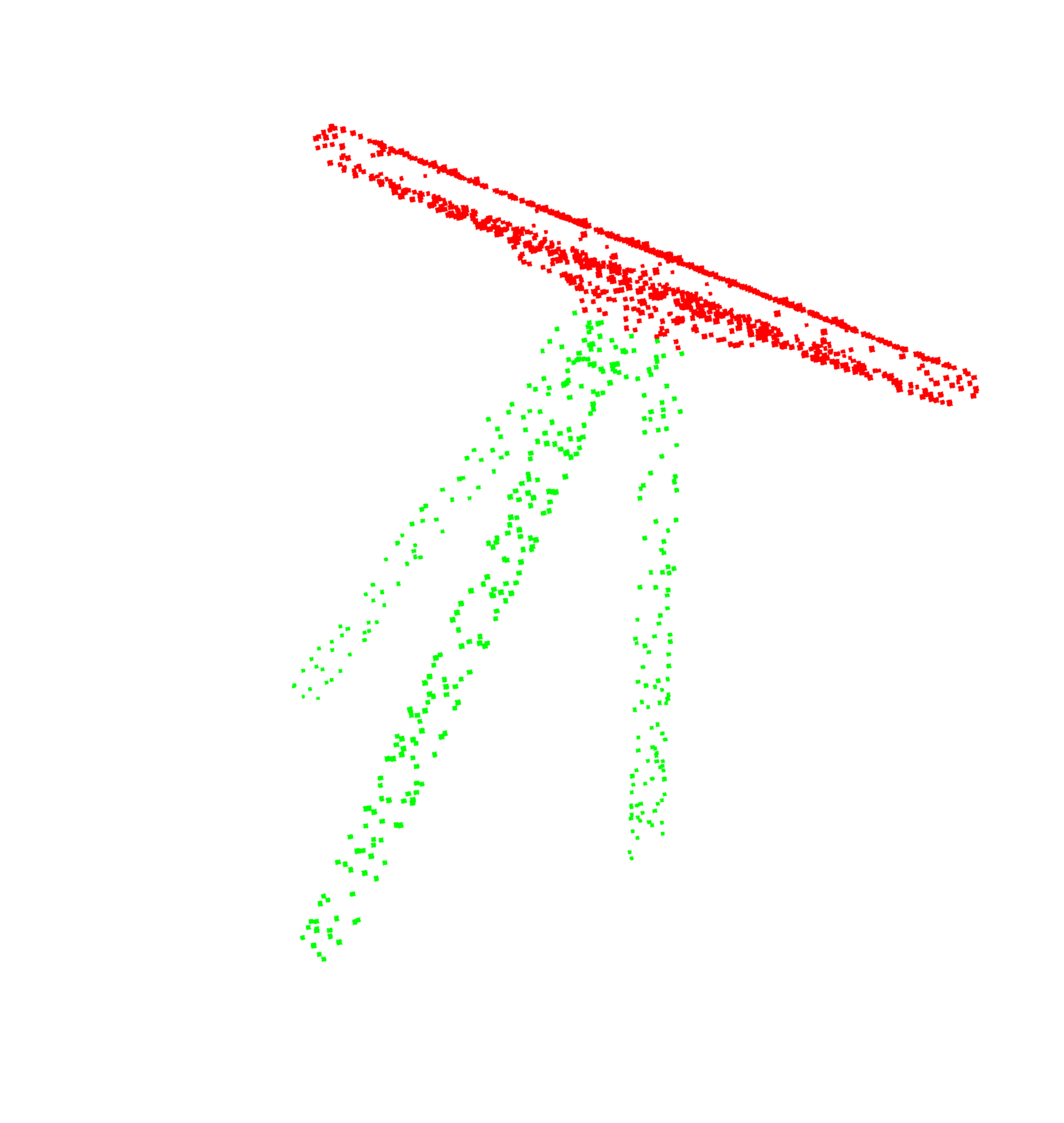}
    \end{minipage}
    \hfill
    \caption{Qualitative Results for Part Segmentation. We visualize our predicted projection images from three different viewpoints.}
    \label{fig:part_segmentation}
\end{figure*}

%% file: fig/supplement/semantic_segmentation.tex
\begin{figure*}[htbp]
    \centering
    \scriptsize
    % 第一行左侧的竖排标签
    \begin{minipage}{0.09\textwidth}
        \centering
        Full
        Fine-tuning
    \end{minipage}
    \hfill
    % 第一行图片
    \begin{minipage}{0.22\textwidth}
        \centering
        \includegraphics[width=\textwidth]{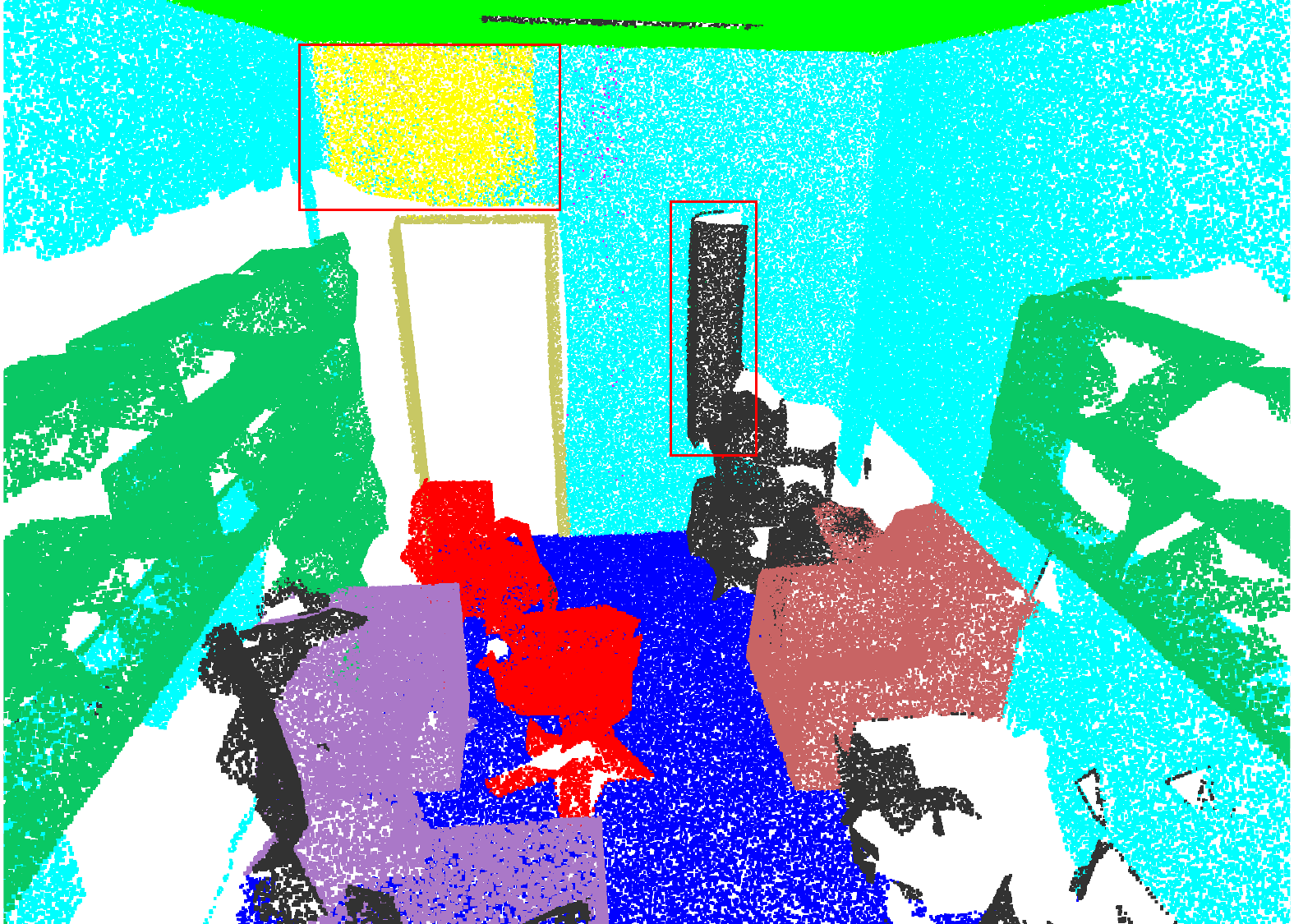}
    \end{minipage}
    \hfill
    \begin{minipage}{0.22\textwidth}
        \centering
        \includegraphics[width=\textwidth]{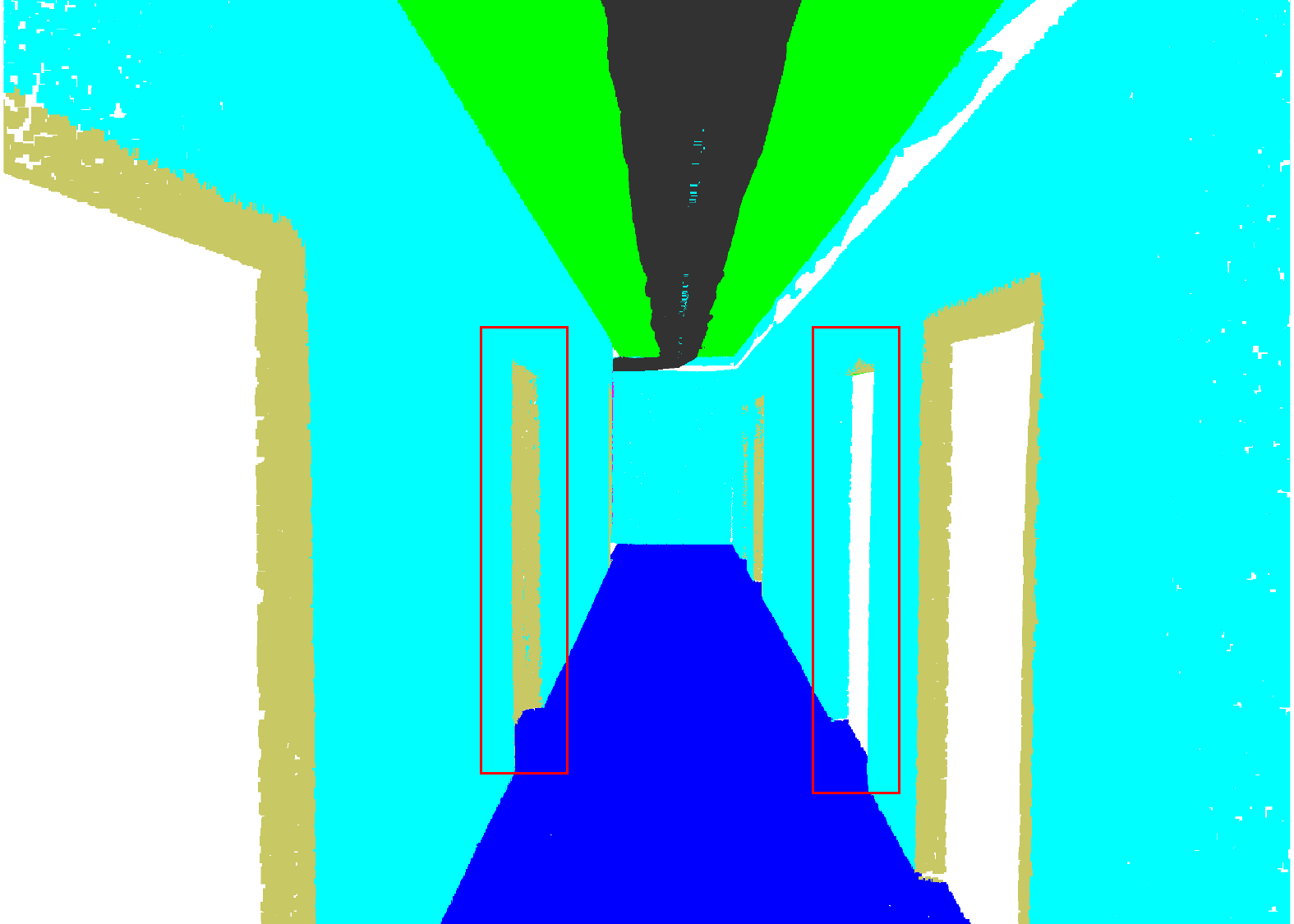} % 替换为你的图片路径
    \end{minipage}
    \hfill
    \begin{minipage}{0.22\textwidth}
        \centering
        \includegraphics[width=\textwidth]{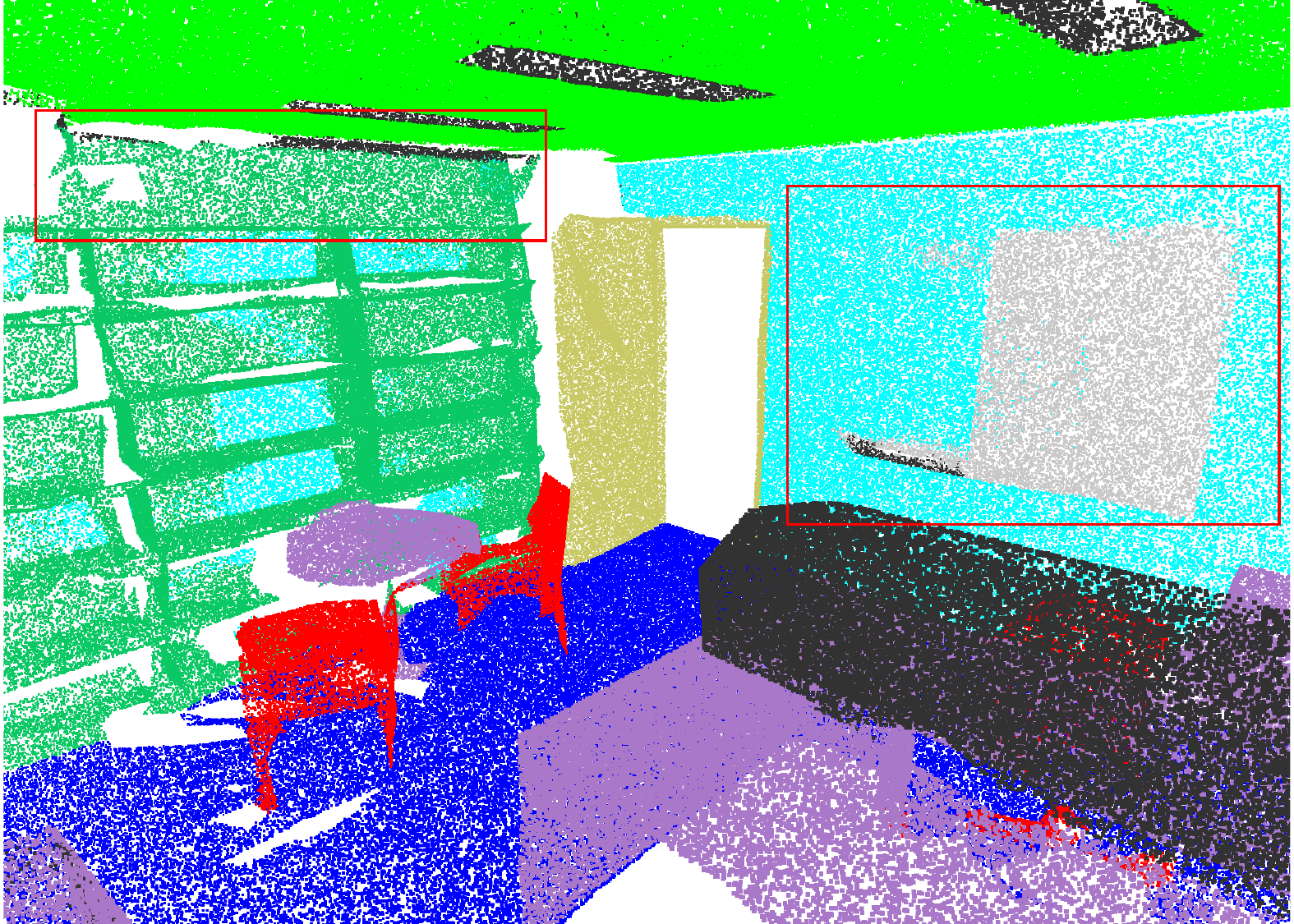}
    \end{minipage}
    \hfill
    \begin{minipage}{0.22\textwidth}
        \centering
        \includegraphics[width=\textwidth]{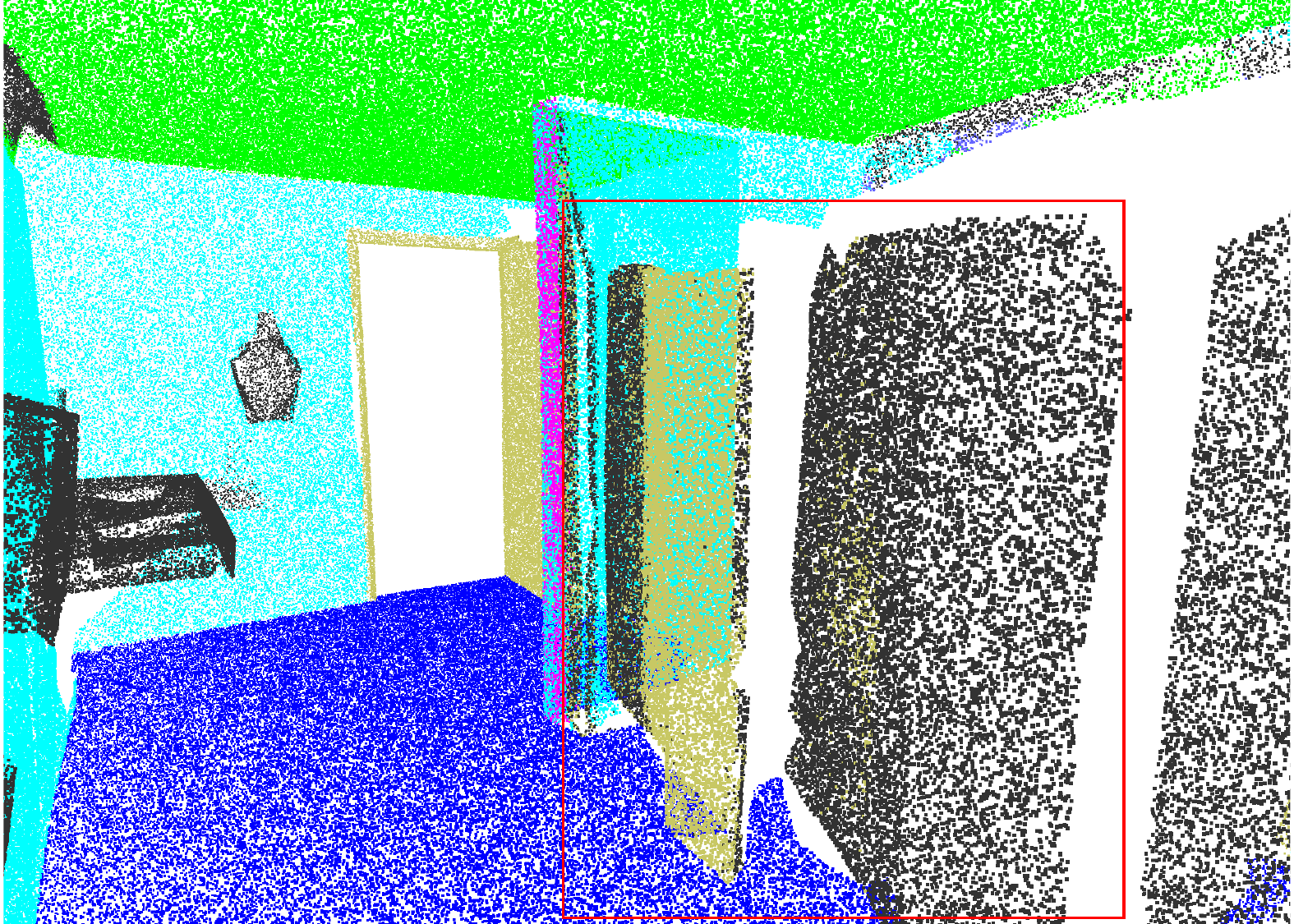}
    \end{minipage}
    \hfill

    % 换行
    \vspace{0.2em}

    % 第二行左侧的竖排标签
    \begin{minipage}{0.09\textwidth}
        \centering
        DAPT
    \end{minipage}
    \hfill
    % 第二行图片
    \begin{minipage}{0.22\textwidth}
        \centering
        \includegraphics[width=\textwidth]{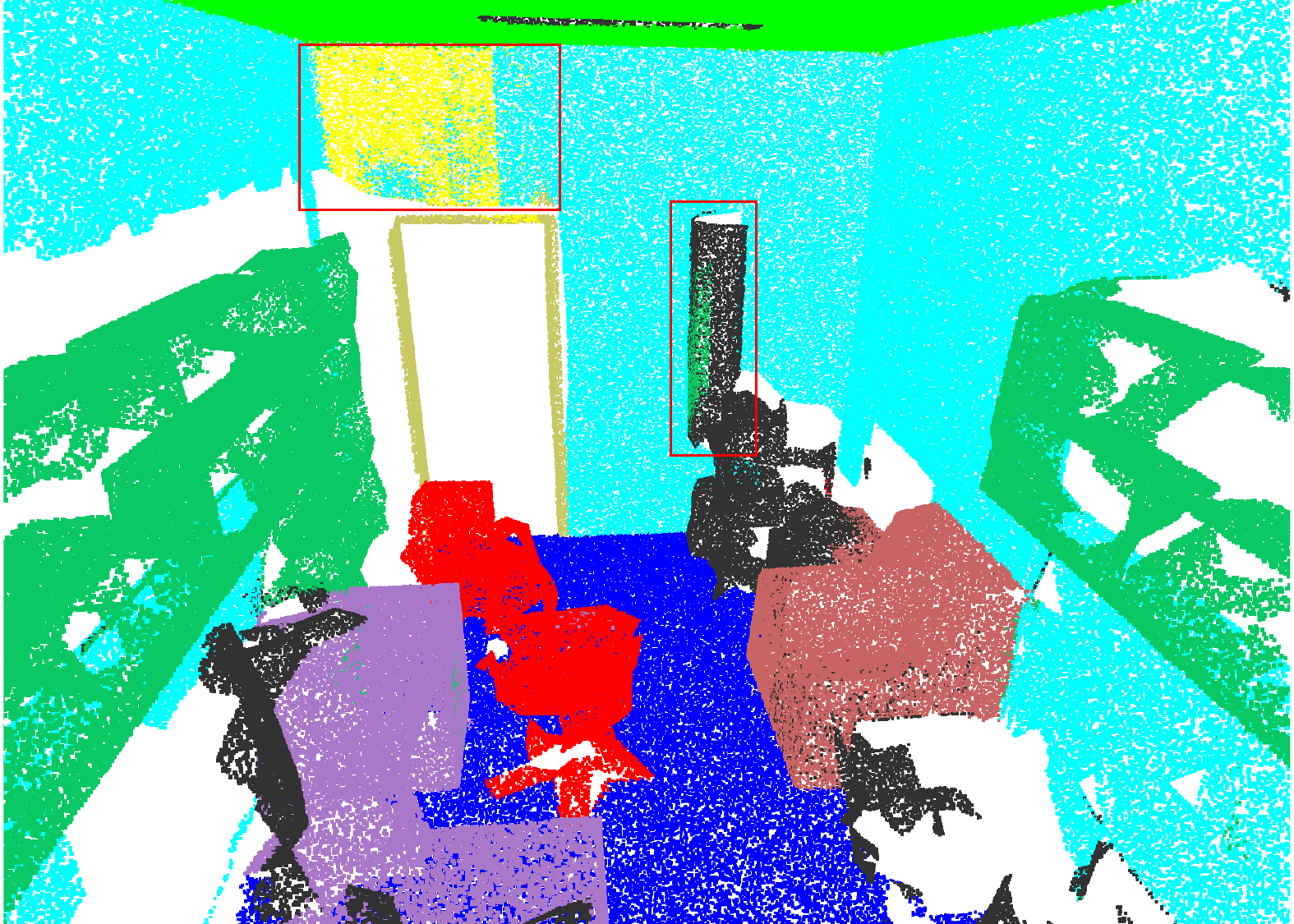}
    \end{minipage}
    \hfill
    \begin{minipage}{0.22\textwidth}
        \centering
        \includegraphics[width=\textwidth]{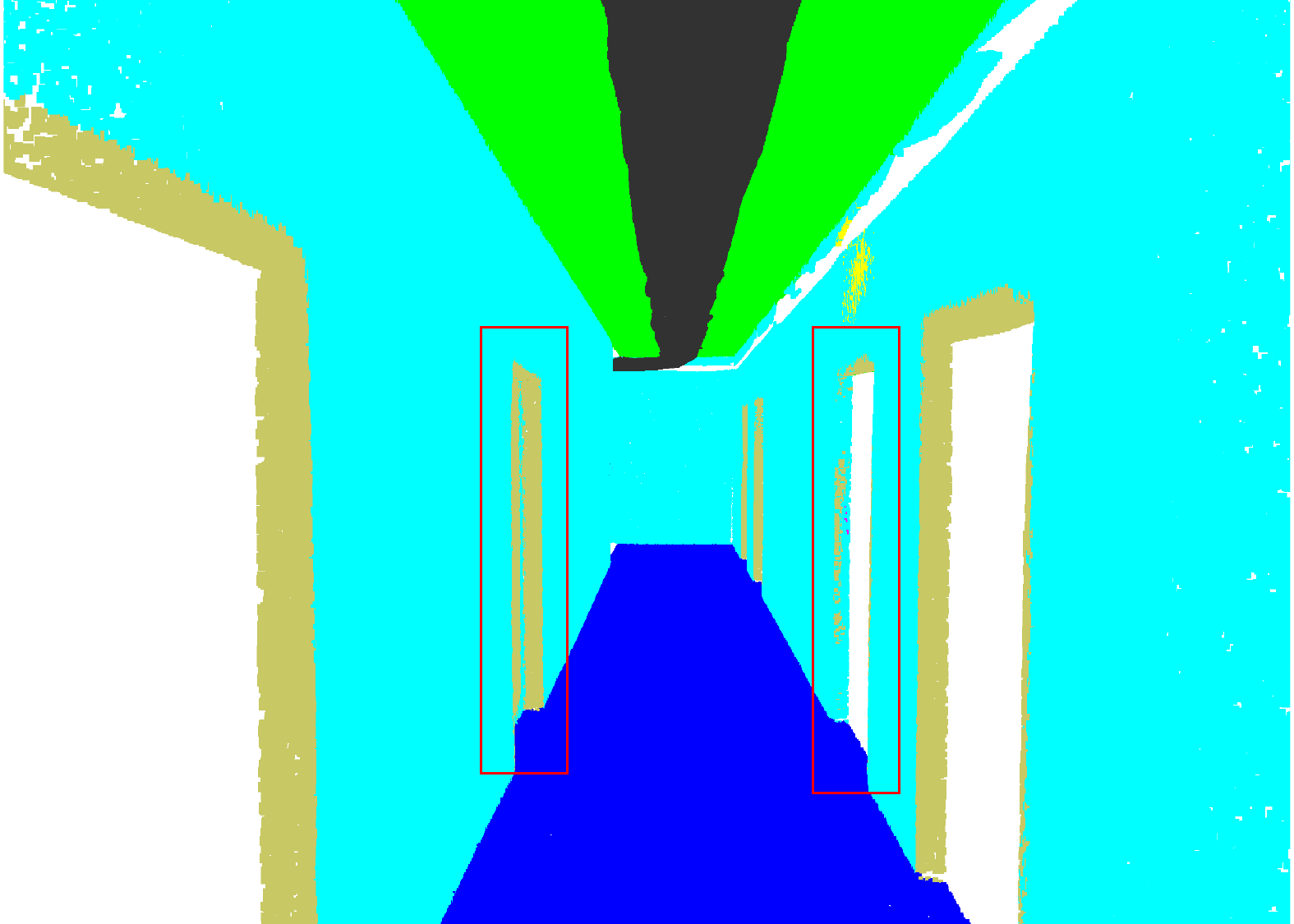}
    \end{minipage}
    \hfill
    \begin{minipage}{0.22\textwidth}
        \centering
        \includegraphics[width=\textwidth]{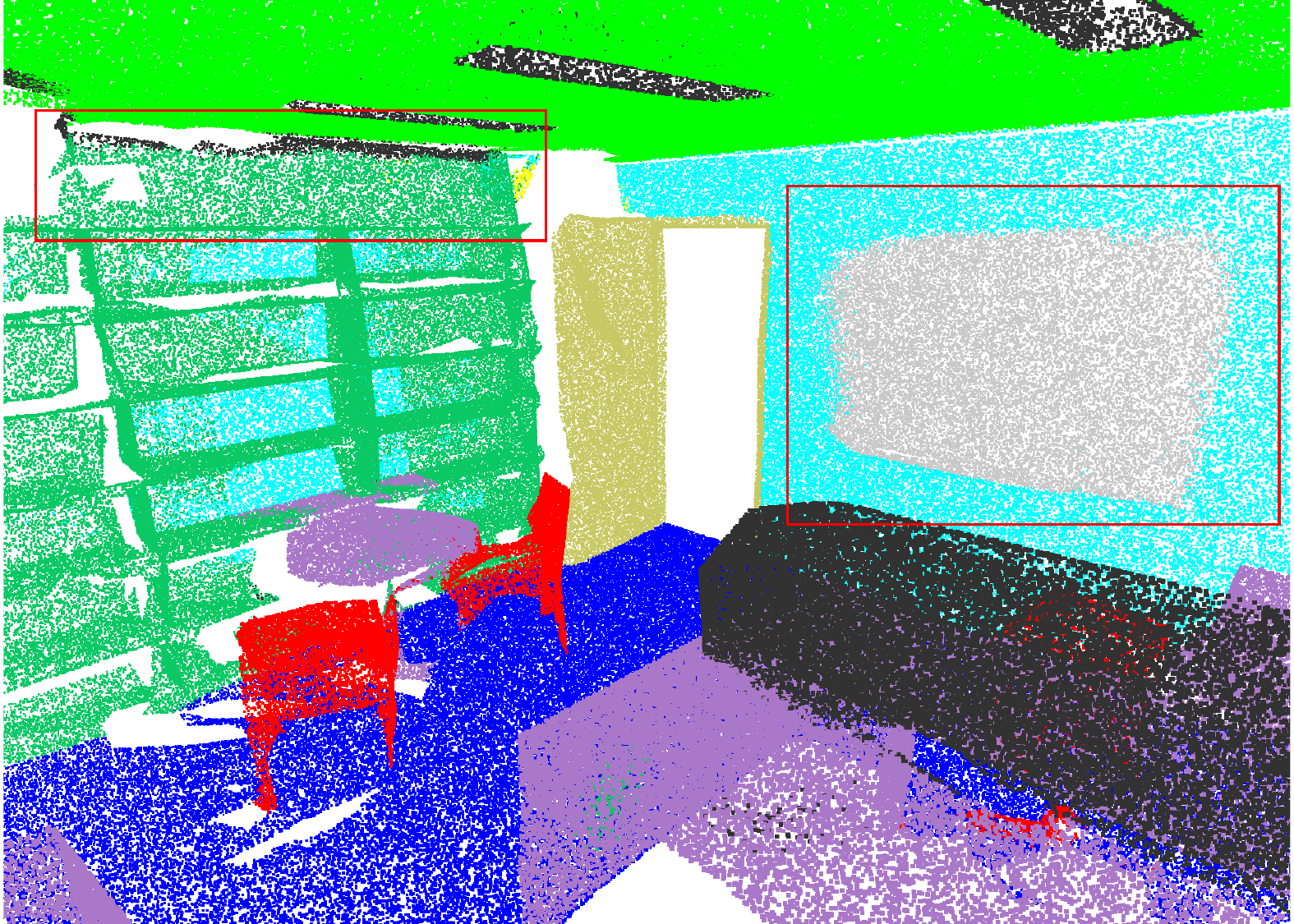}
    \end{minipage}
    \hfill
    \begin{minipage}{0.22\textwidth}
        \centering
        \includegraphics[width=\textwidth]{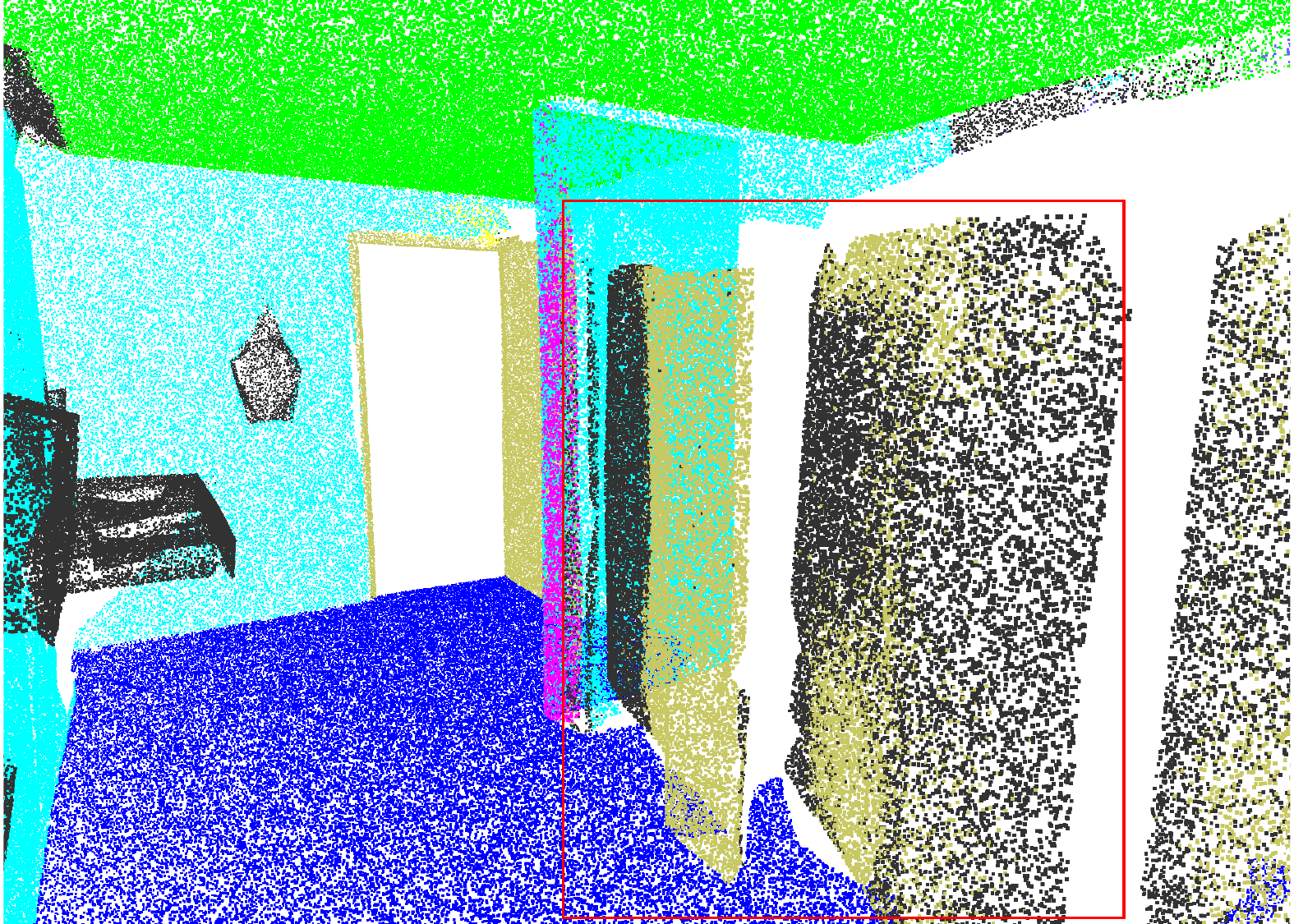}
    \end{minipage}
    \hfill

    % 换行
    \vspace{0.2em}

    % 第三行左侧的竖排标签
    \begin{minipage}{0.09\textwidth}
        \centering
        IDPT
    \end{minipage}
    \hfill
    % 第三行图片
    \begin{minipage}{0.22\textwidth}
        \centering
        \includegraphics[width=\textwidth]{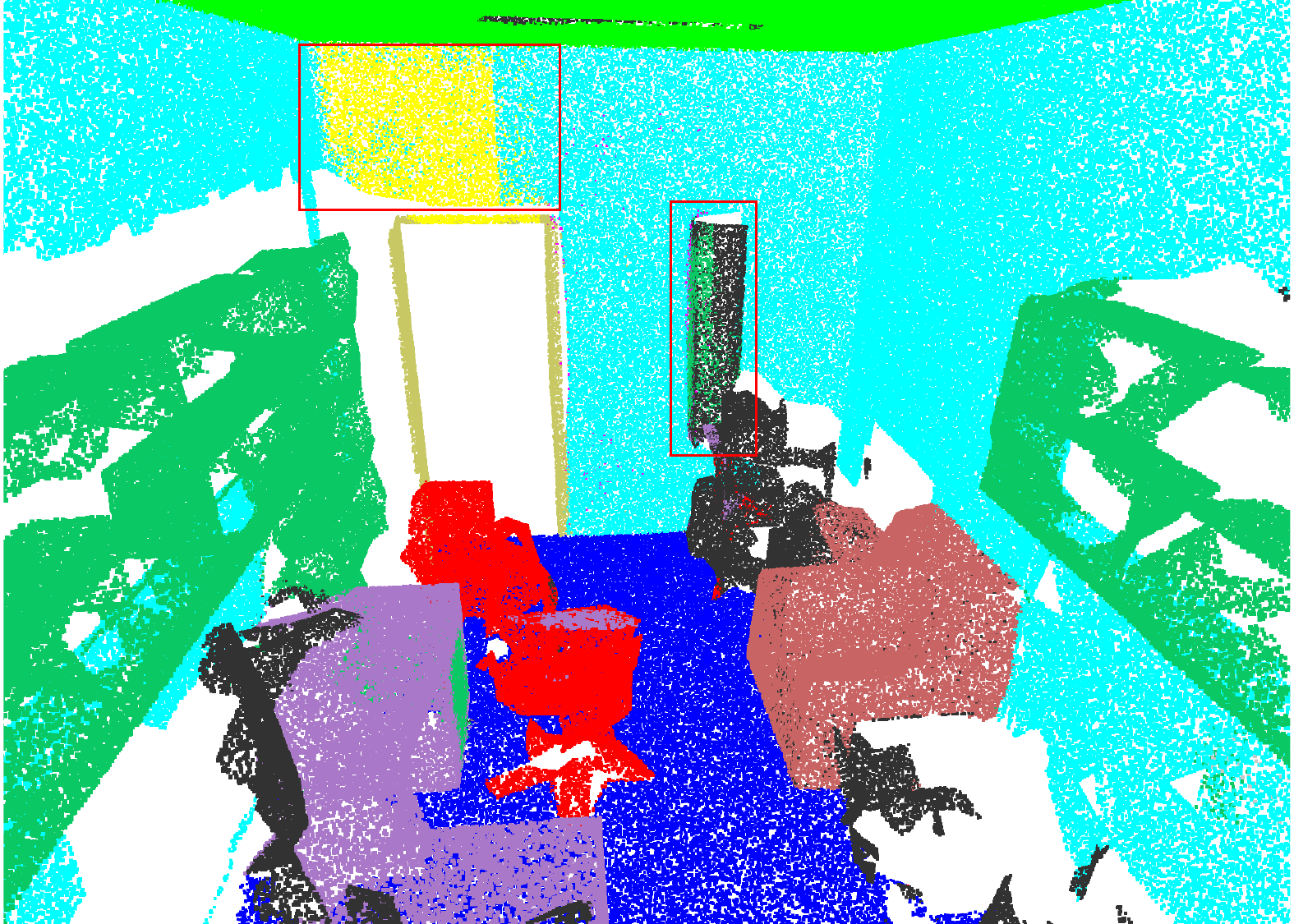}
    \end{minipage}
    \hfill
    \begin{minipage}{0.22\textwidth}
        \centering
        \includegraphics[width=\textwidth]{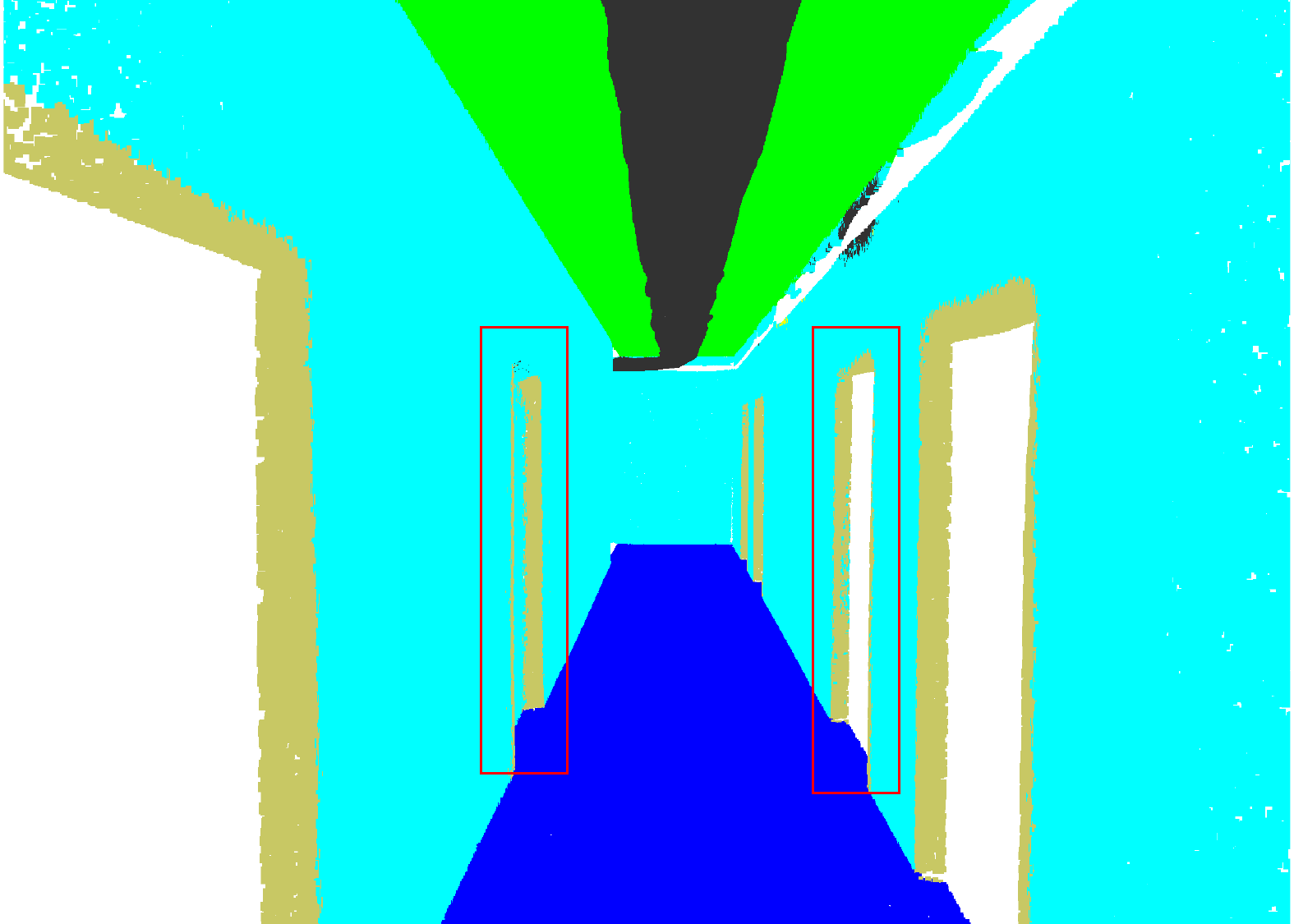}
    \end{minipage}
    \hfill
    \begin{minipage}{0.22\textwidth}
        \centering
        \includegraphics[width=\textwidth]{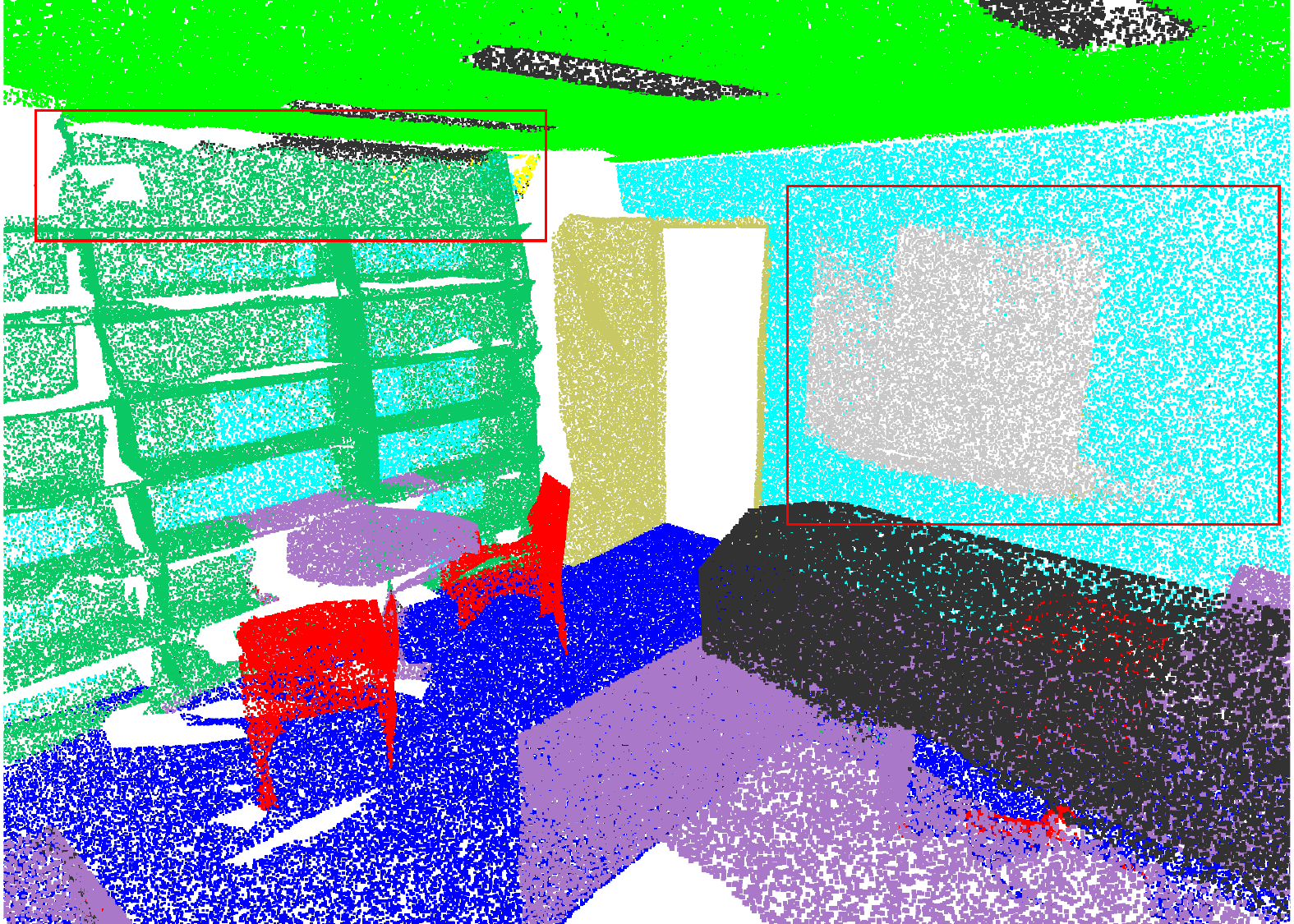}
    \end{minipage}
    \hfill
    \begin{minipage}{0.22\textwidth}
        \centering
        \includegraphics[width=\textwidth]{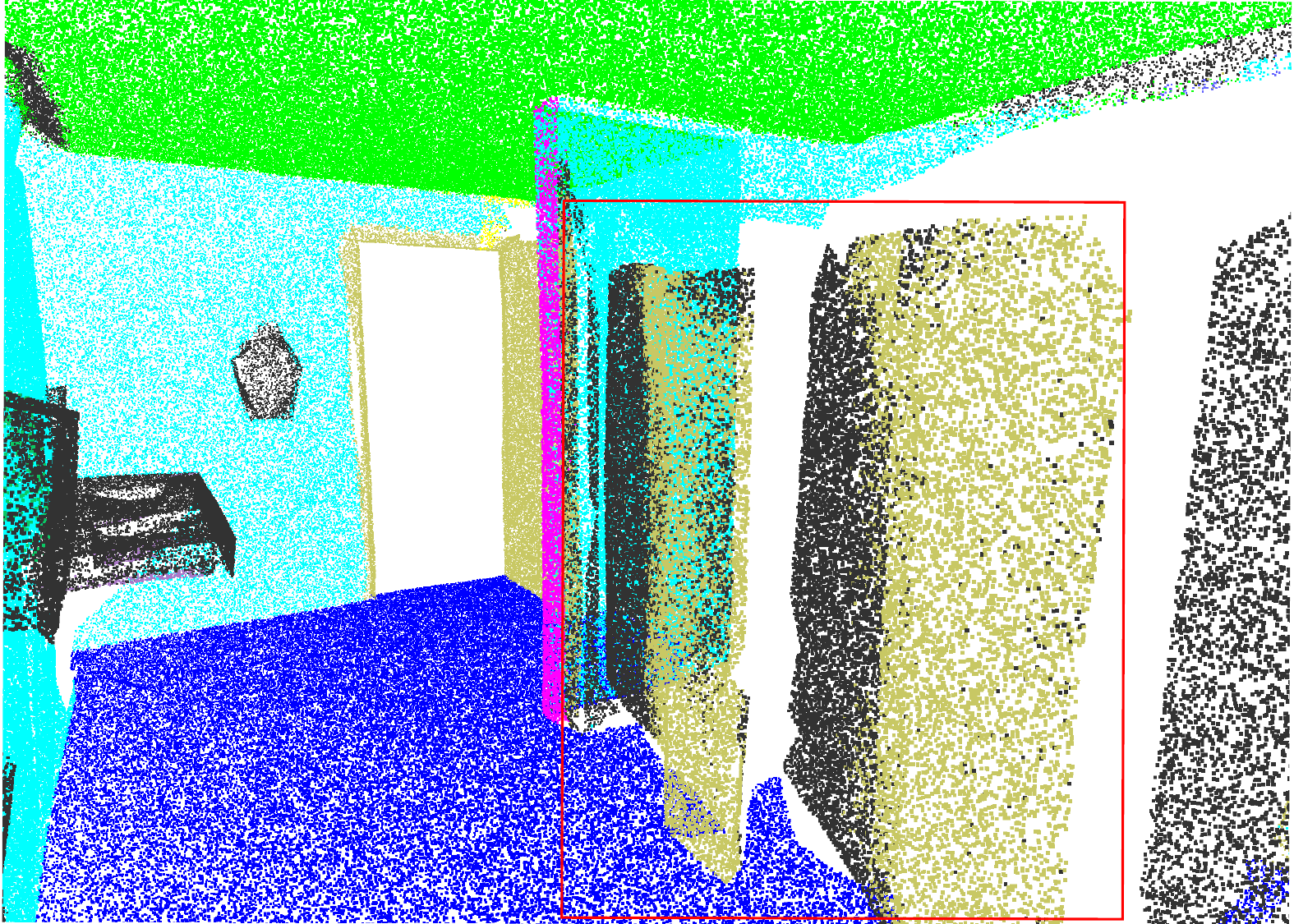}
    \end{minipage}
    \hfill

    % 换行
    \vspace{0.2em}

    % 第四行左侧的竖排标签
    \begin{minipage}{0.09\textwidth}
        \centering
        PPT
    \end{minipage}
    \hfill
    % 第四行图片
    \begin{minipage}{0.22\textwidth}
        \centering
        \includegraphics[width=\textwidth]{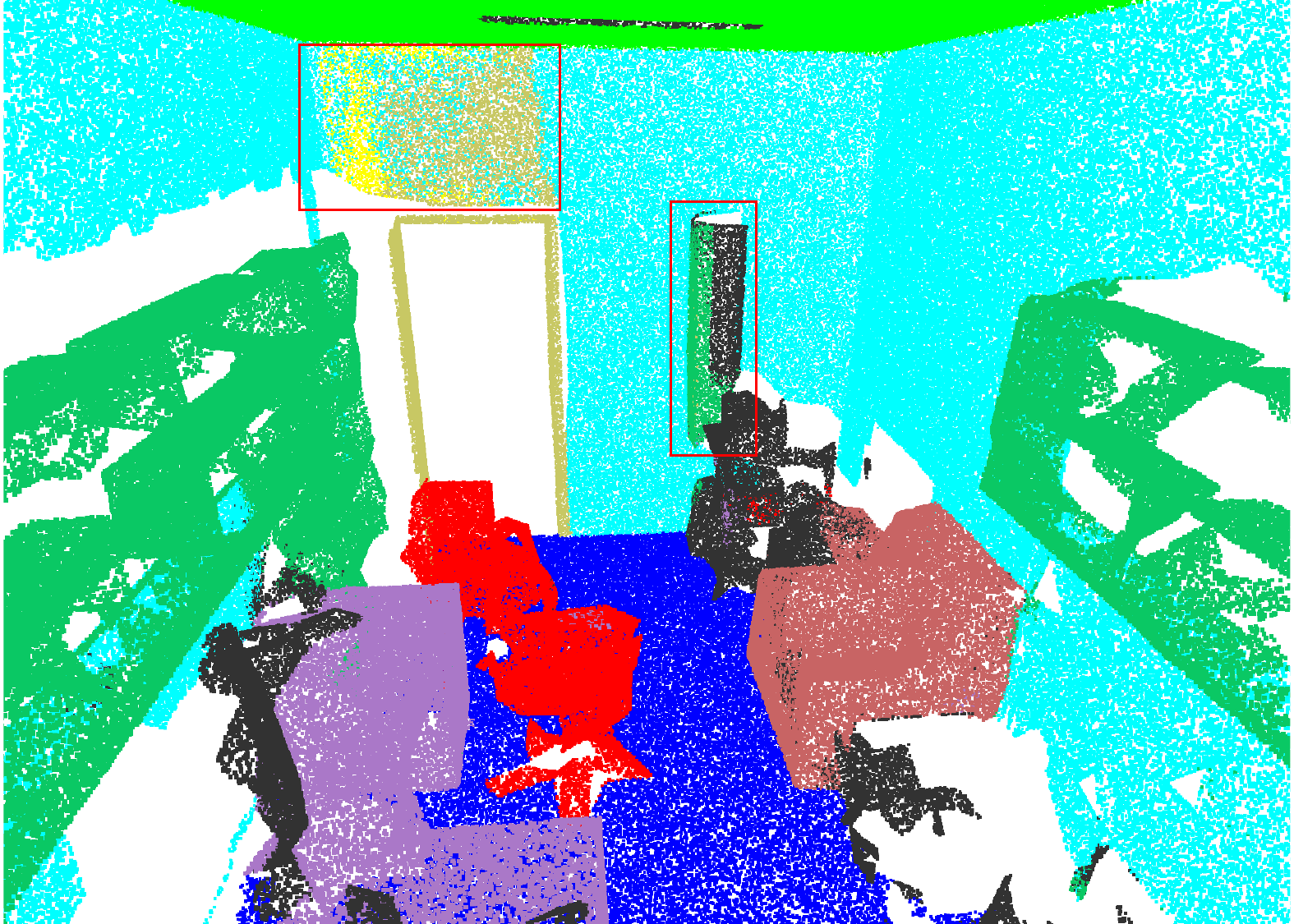}
    \end{minipage}
    \hfill
     \begin{minipage}{0.22\textwidth}
        \centering
        \includegraphics[width=\textwidth]{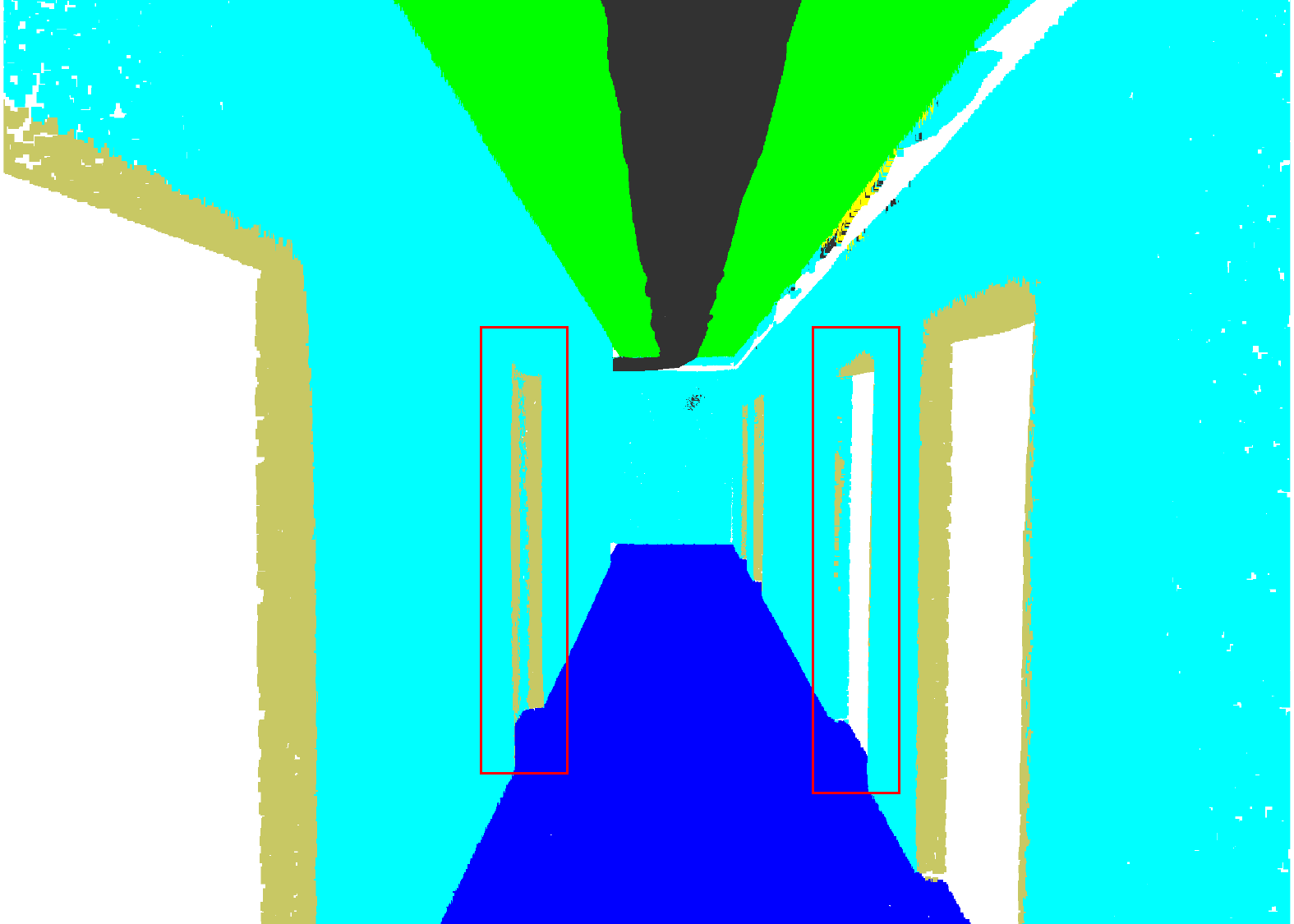}
    \end{minipage}
    \hfill
    \begin{minipage}{0.22\textwidth}
        \centering
        \includegraphics[width=\textwidth]{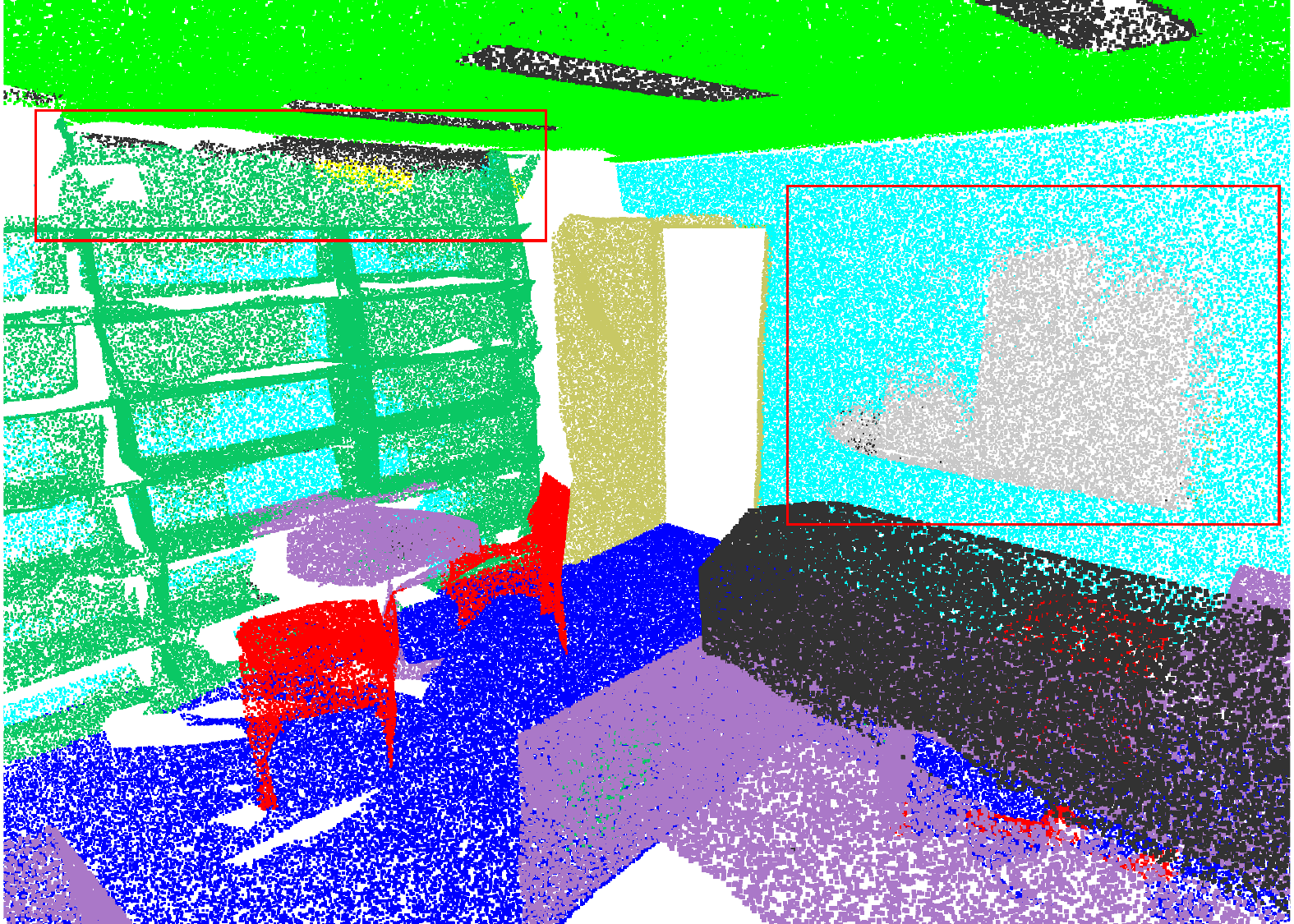}
    \end{minipage}
    \hfill
    \begin{minipage}{0.22\textwidth}
        \centering
        \includegraphics[width=\textwidth]{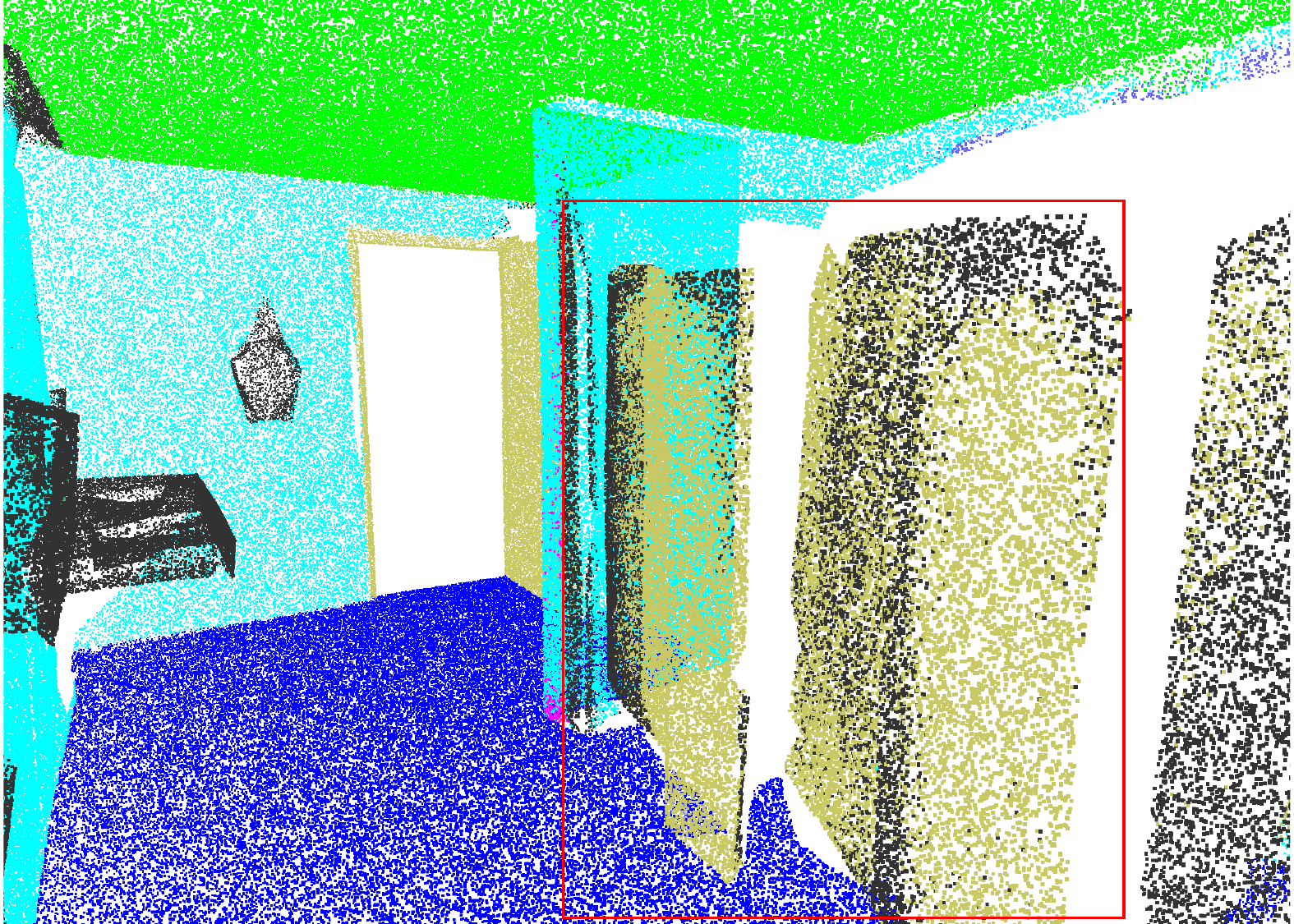}
    \end{minipage}
    \hfill

    % 换行
    \vspace{0.2em}

    % 第五行左侧的竖排标签
    \begin{minipage}{0.09\textwidth}
        \centering
        PointGST
    \end{minipage}
    \hfill
    % 第五行图片
    \begin{minipage}{0.22\textwidth}
        \centering
        \includegraphics[width=\textwidth]{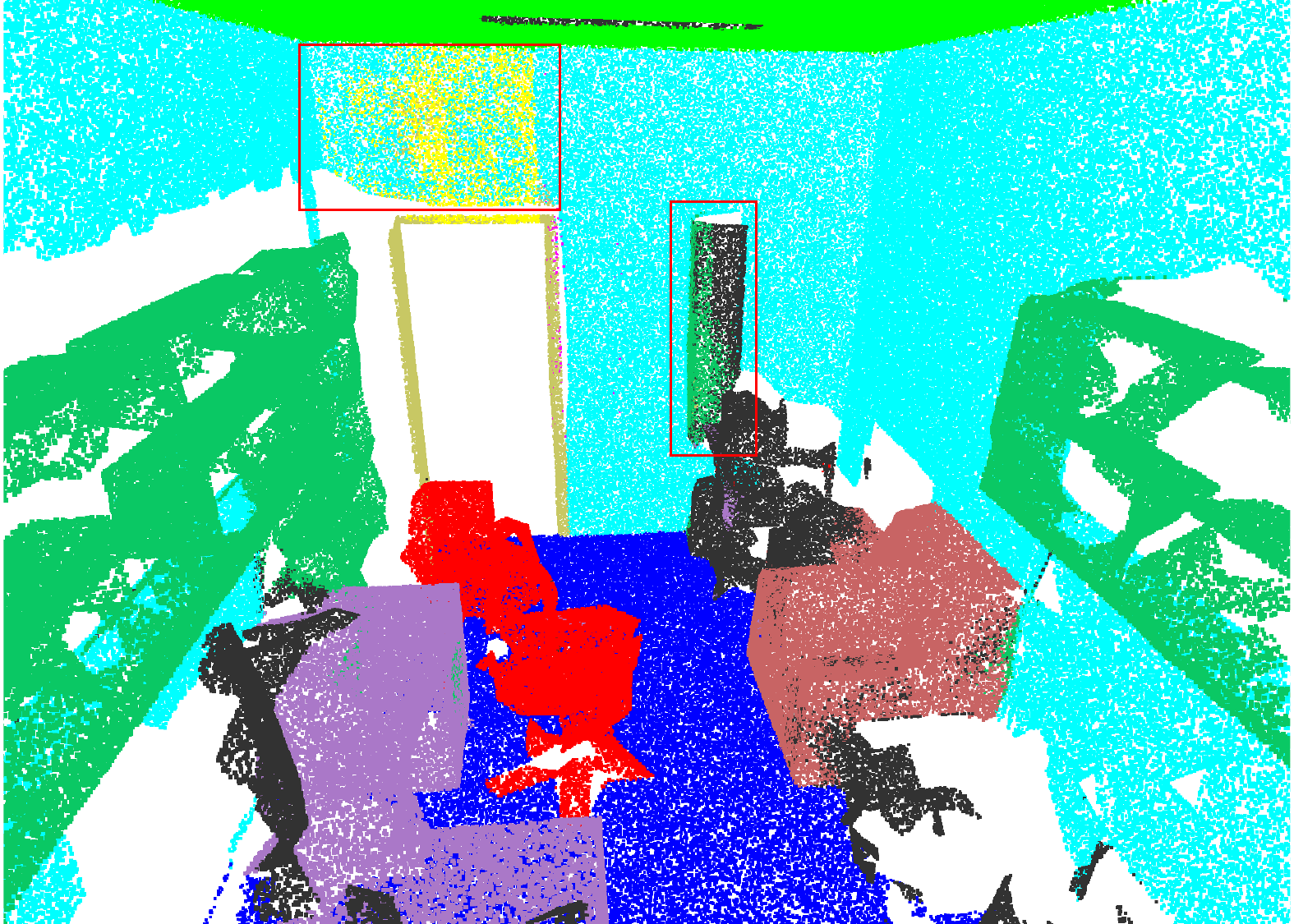}
    \end{minipage}
    \hfill
    \begin{minipage}{0.22\textwidth}
        \centering
        \includegraphics[width=\textwidth]{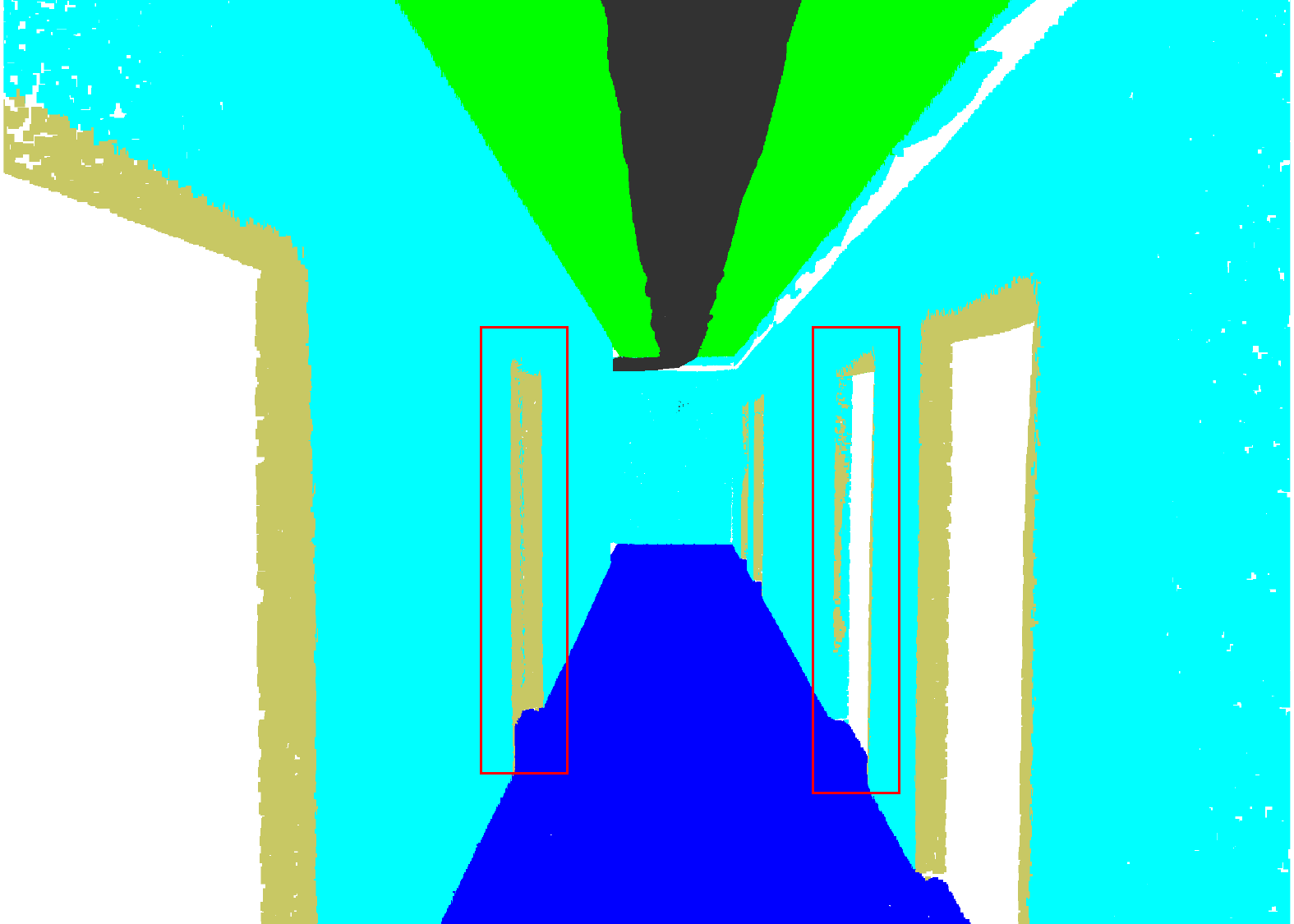}
    \end{minipage}
    \hfill
    \begin{minipage}{0.22\textwidth}
        \centering
        \includegraphics[width=\textwidth]{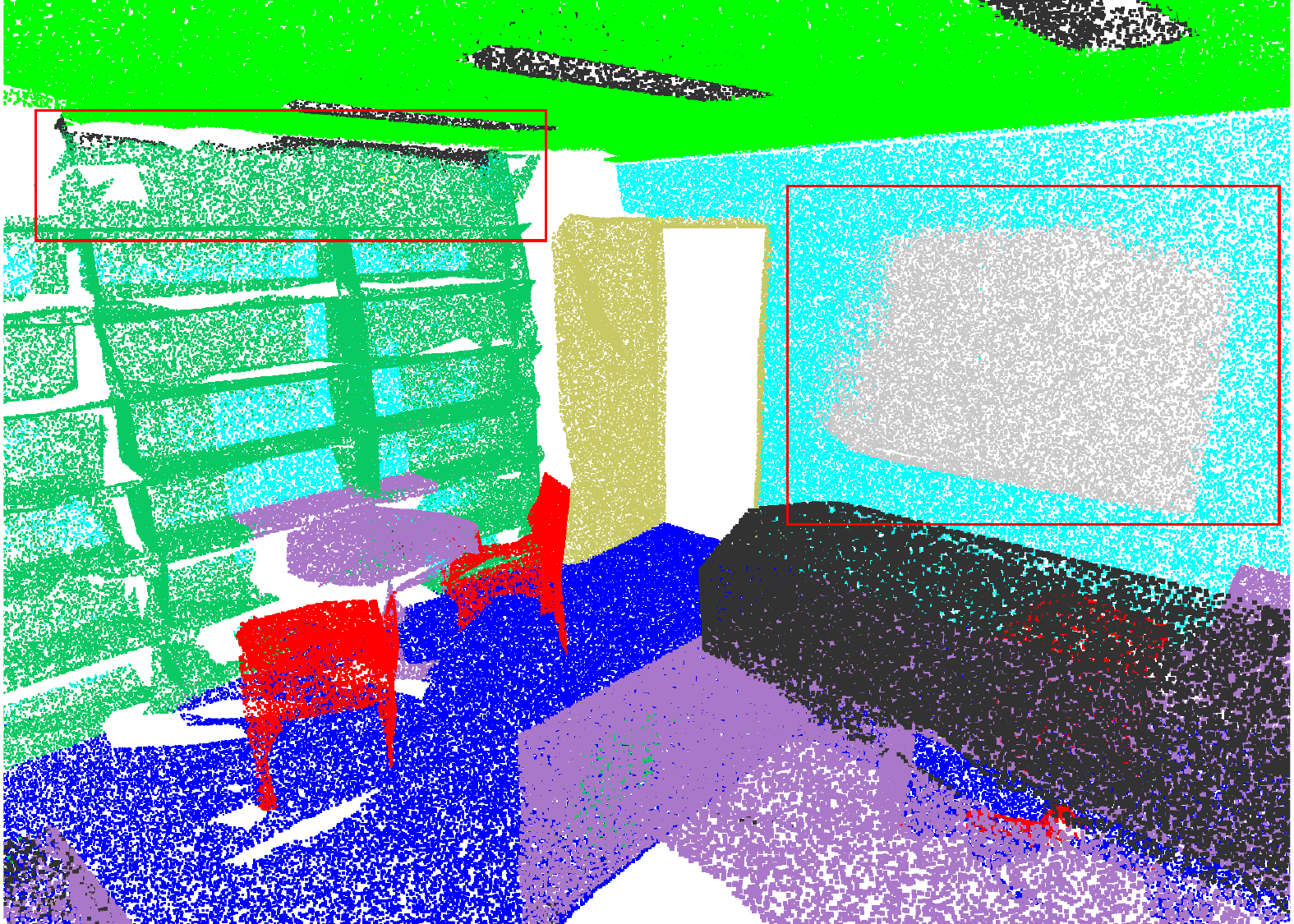}
    \end{minipage}
    \hfill
    \begin{minipage}{0.22\textwidth}
        \centering
        \includegraphics[width=\textwidth]{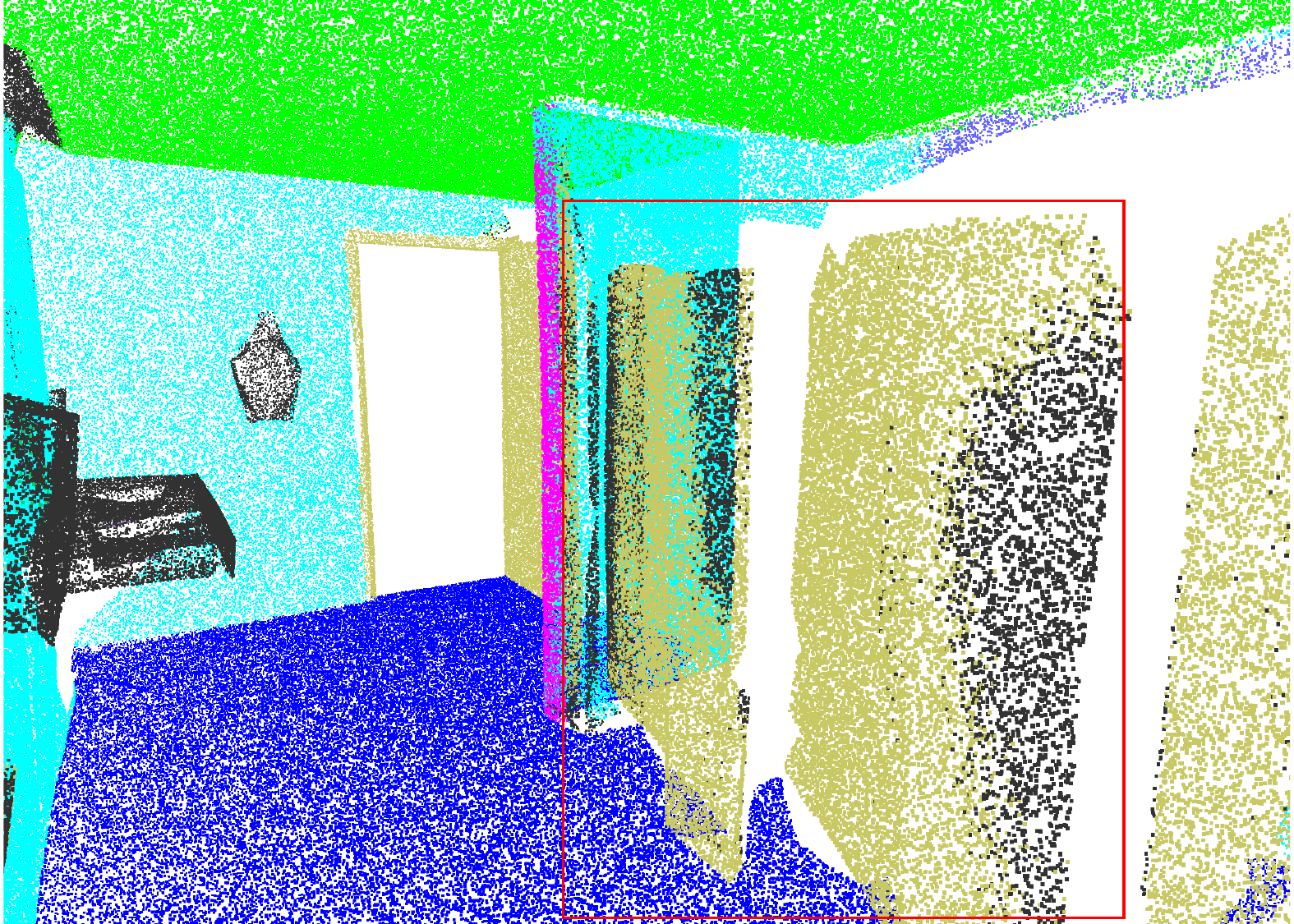}
    \end{minipage}
    \hfill

    % 换行
    \vspace{0.2em}

    % 第六行左侧的竖排标签
    \begin{minipage}{0.09\textwidth}
        \centering
        PLT (Ours)
    \end{minipage}
    \hfill
    % 第六行图片
    \begin{minipage}{0.2\textwidth}
        \centering
        \includegraphics[width=\textwidth]{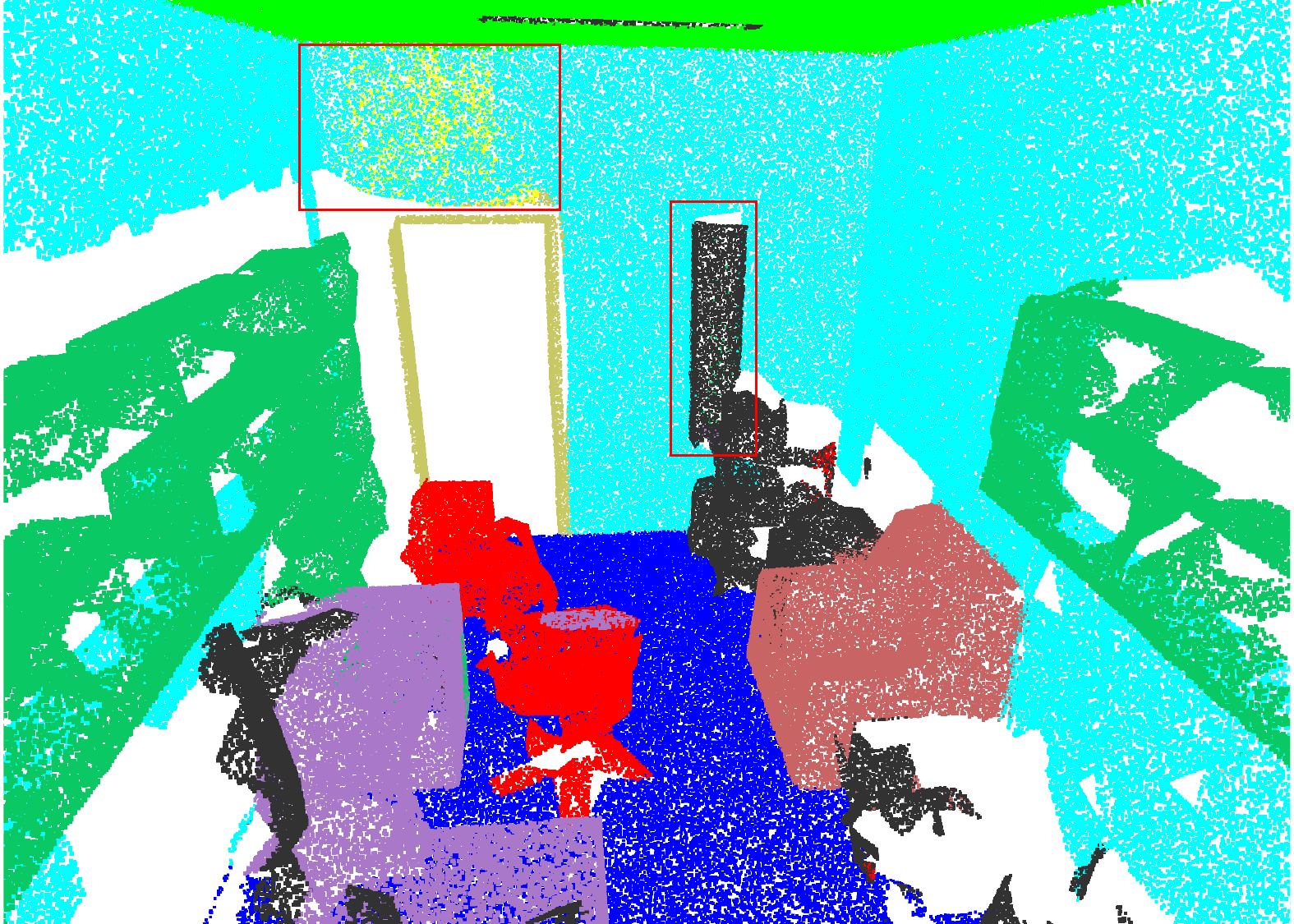}
    \end{minipage}
    \hfill
    \begin{minipage}{0.22\textwidth}
        \centering
        \includegraphics[width=\textwidth]{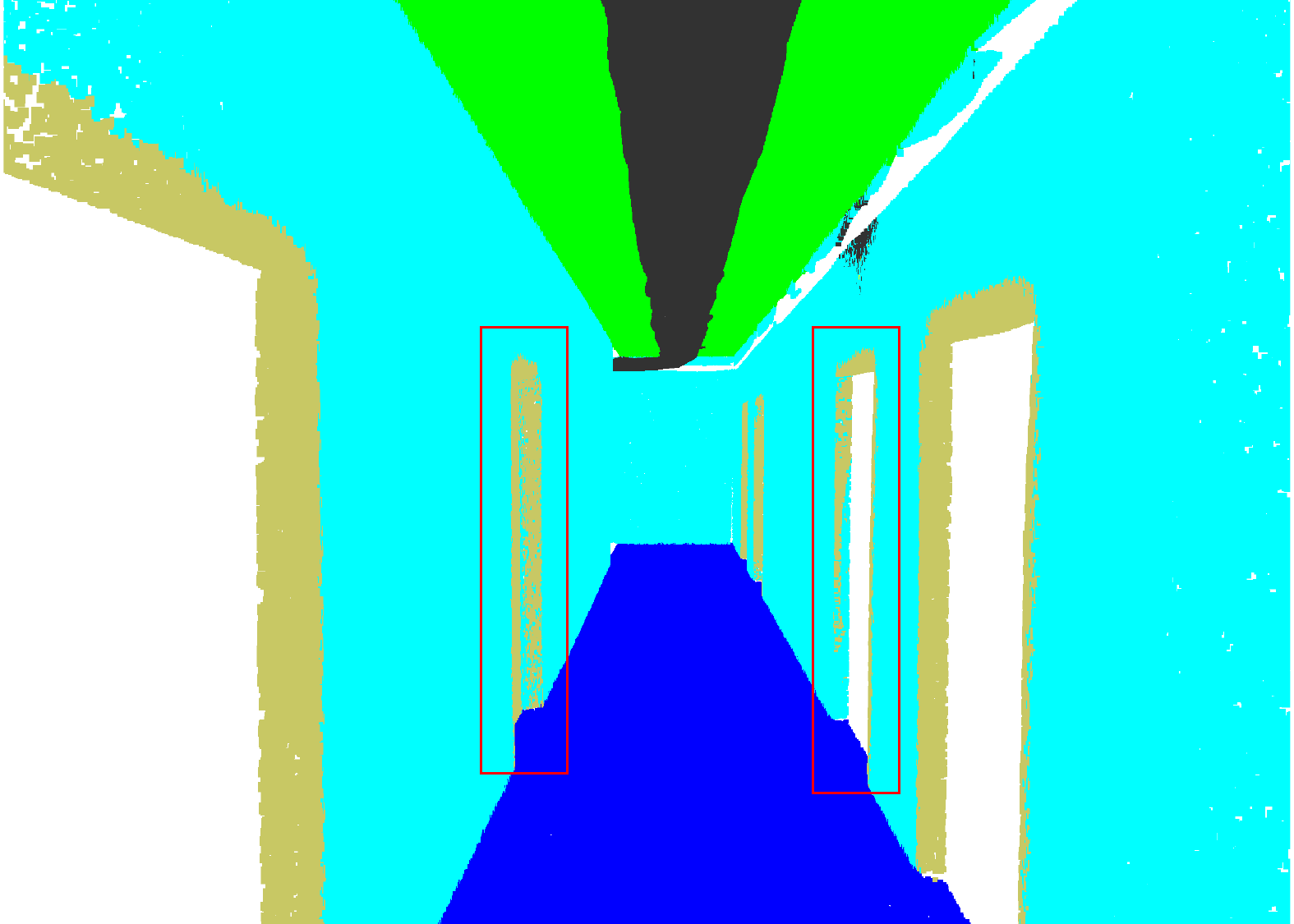}
    \end{minipage}
    \hfill
    \begin{minipage}{0.22\textwidth}
        \centering
        \includegraphics[width=\textwidth]{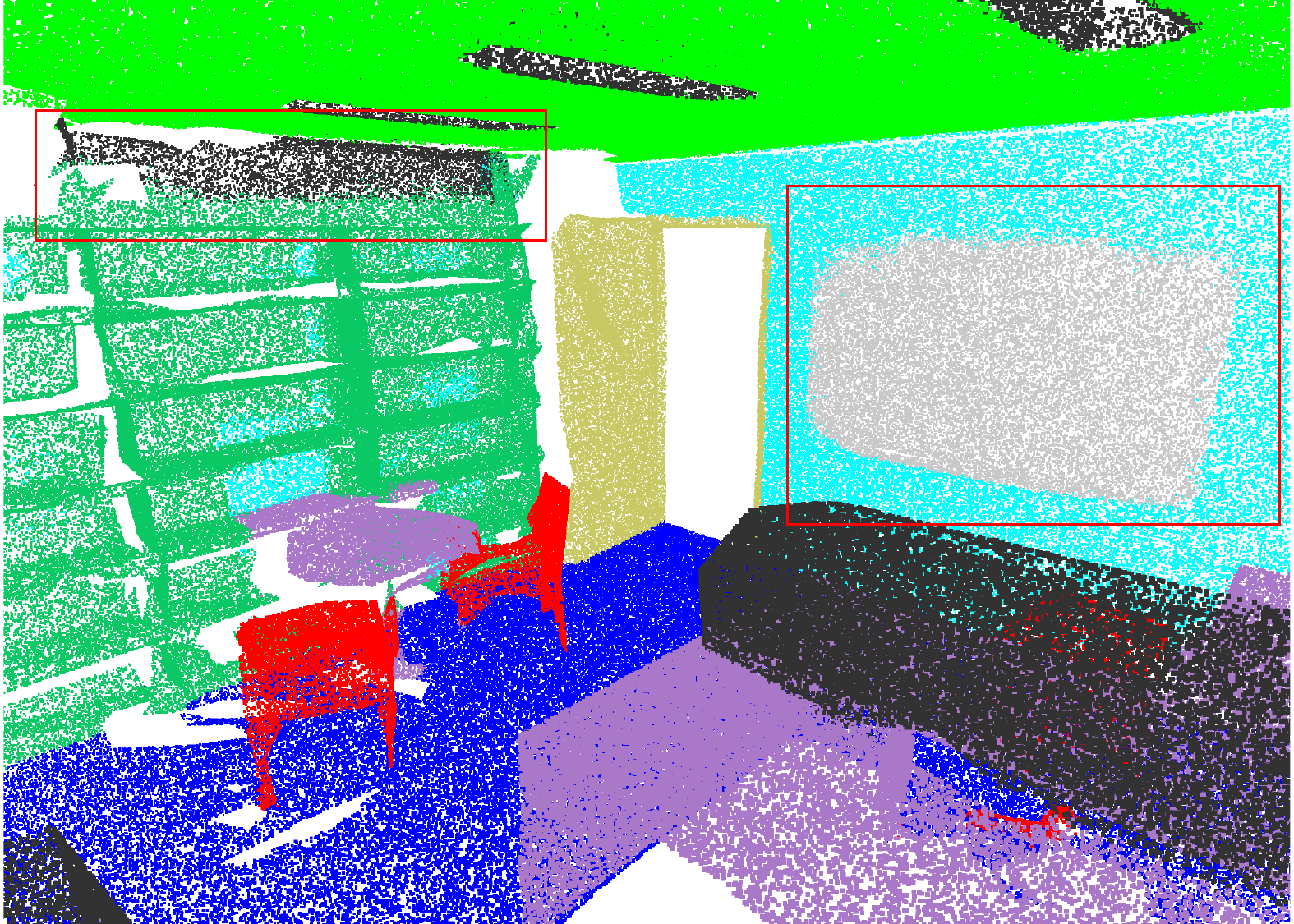}
    \end{minipage}
    \hfill
    \begin{minipage}{0.22\textwidth}
        \centering
        \includegraphics[width=\textwidth]{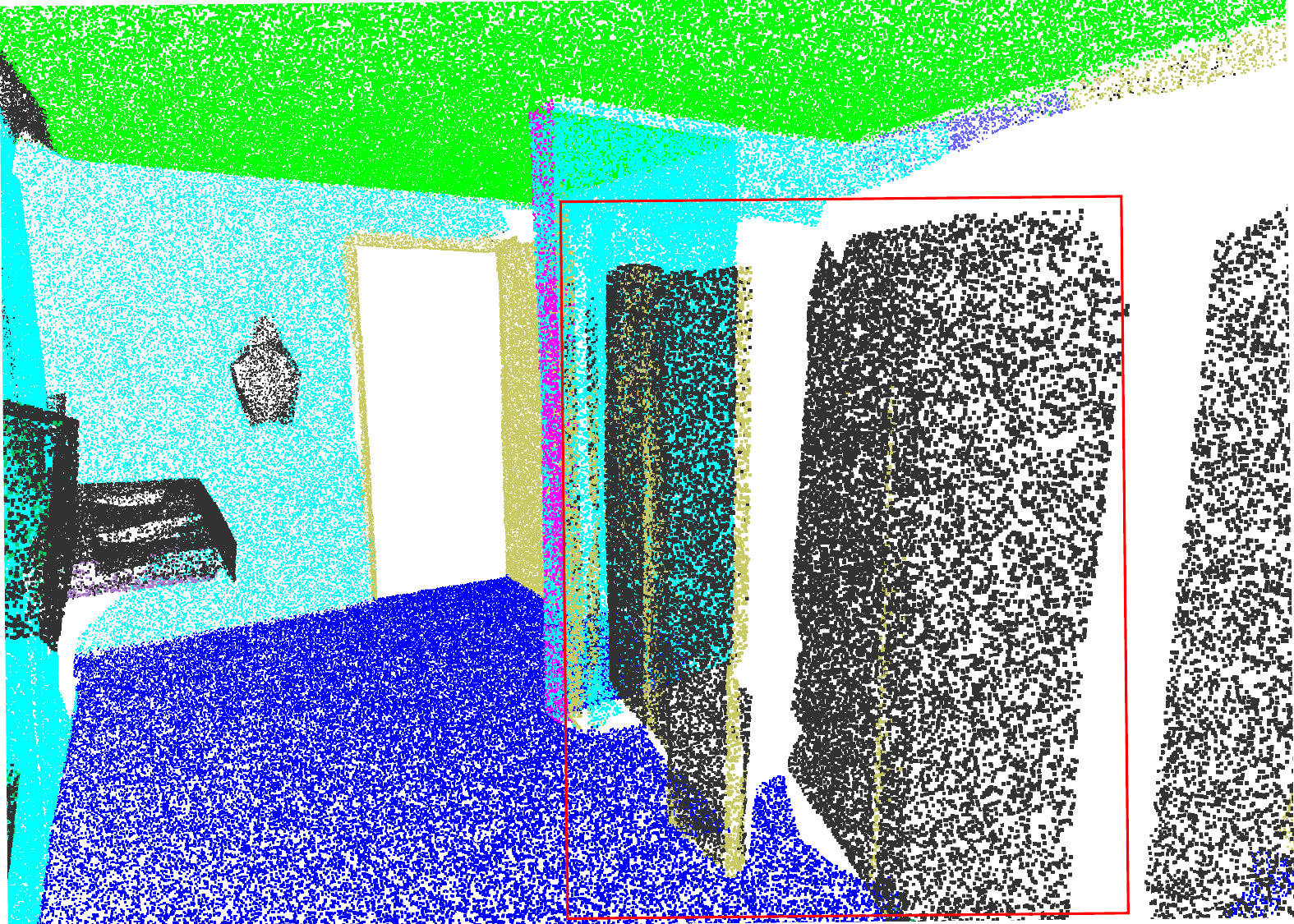}
    \end{minipage}
    \hfill

    % 换行
    \vspace{0.2em}

    % 第七行左侧的竖排标签
    \begin{minipage}{0.09\textwidth}
        \centering
        GT
    \end{minipage}
    \hfill
    % 第七行图片
    \begin{minipage}{0.22\textwidth}
        \centering
        \includegraphics[width=\textwidth]{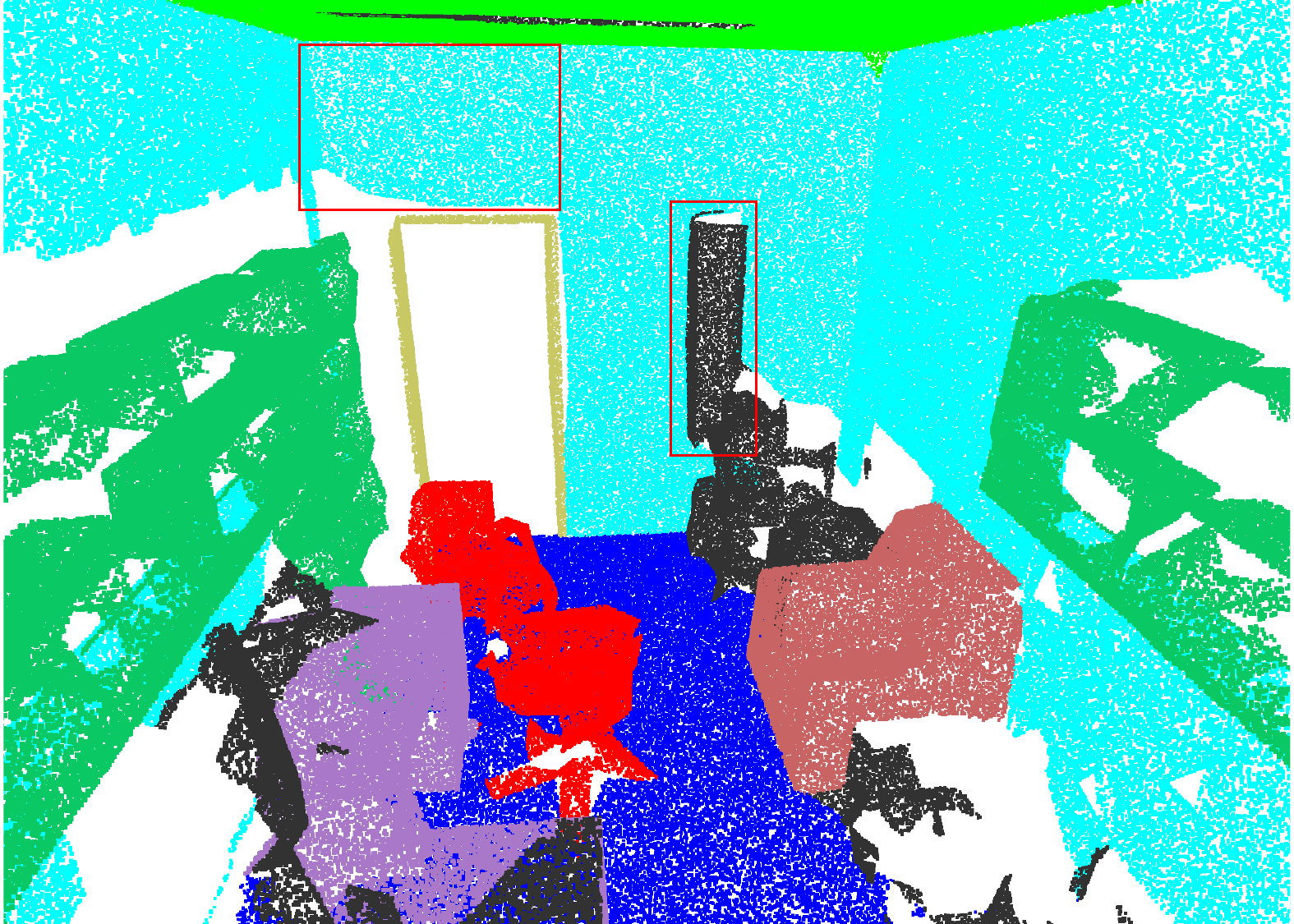}
    \end{minipage}
    \hfill
    \begin{minipage}{0.22\textwidth}
        \centering
        \includegraphics[width=\textwidth]{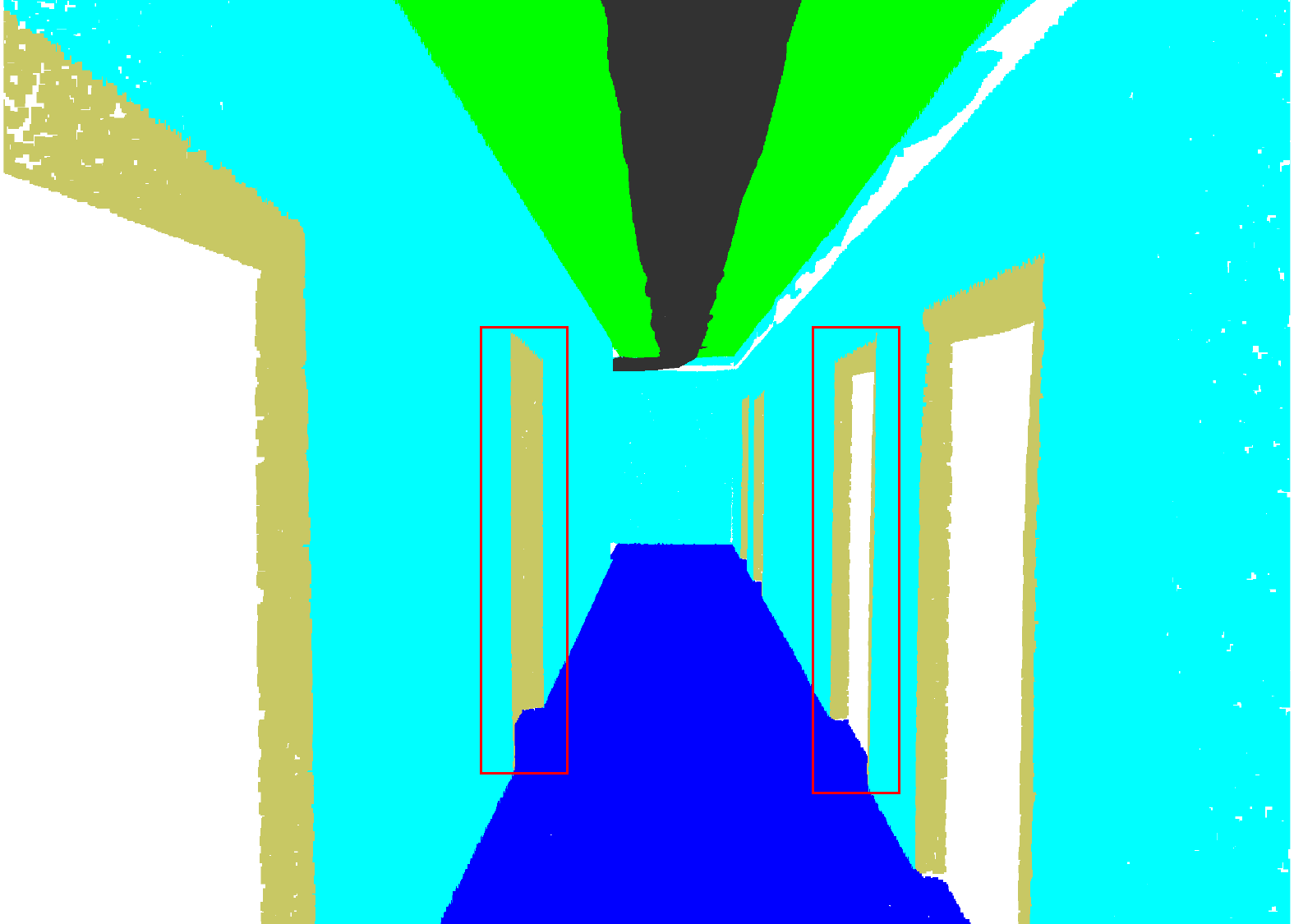}
    \end{minipage}
    \hfill
    \begin{minipage}{0.22\textwidth}
        \centering
        \includegraphics[width=\textwidth]{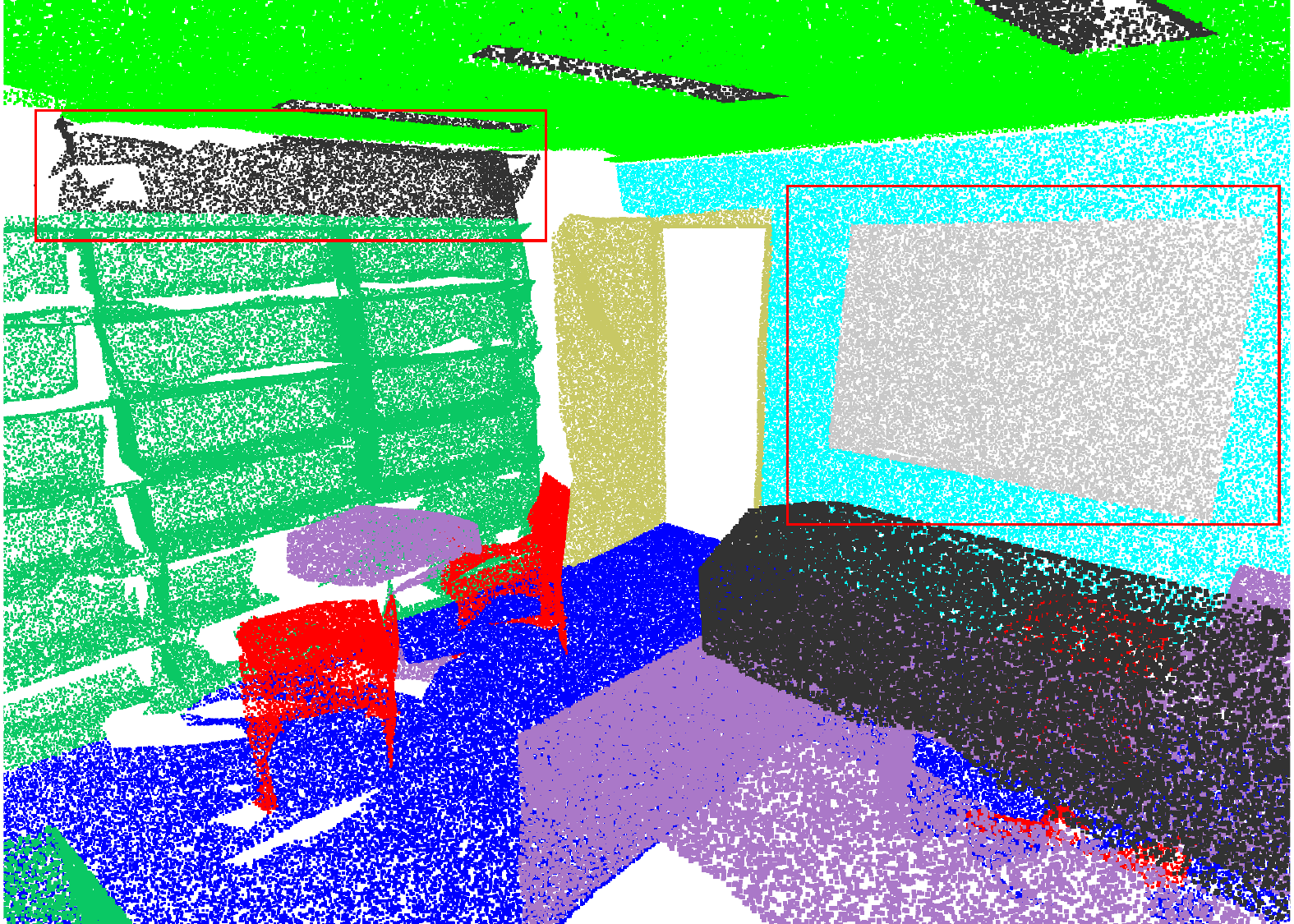}
    \end{minipage}
    \hfill
    \begin{minipage}{0.22\textwidth}
        \centering
        \includegraphics[width=\textwidth]{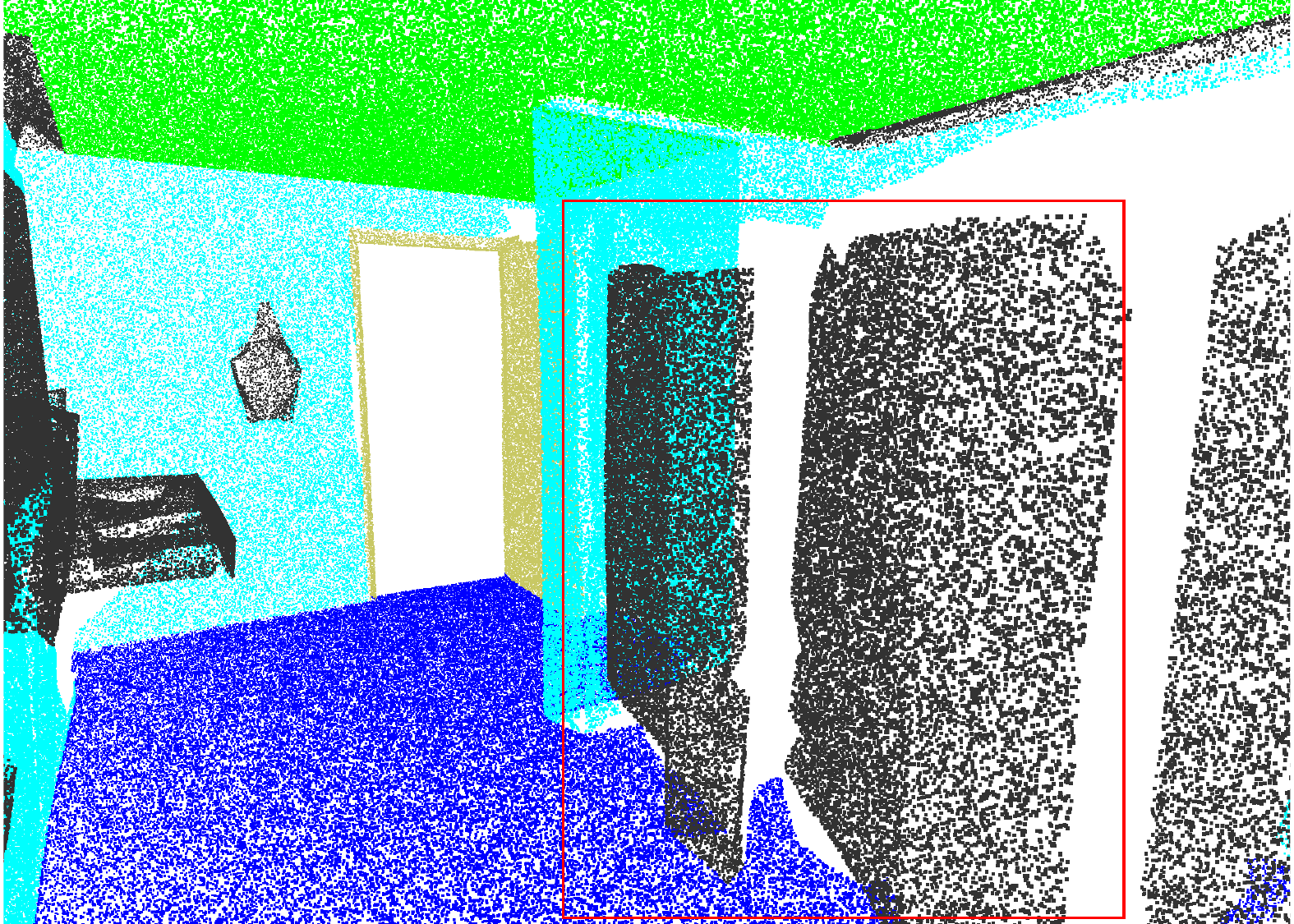}
    \end{minipage}
    \hfill
    
    %下方的标签
    \vspace{0.2em}
    \begin{minipage}{0.09\textwidth} % 左侧空白区域
        \color{white}{12}
    \end{minipage}
    \hfill
    \begin{minipage}{0.22\textwidth} % 第2列标题
        \centering
        office\_9
    \end{minipage}
    \hfill
    \begin{minipage}{0.22\textwidth} % 第1列标题
        \centering
        hallway\_10
    \end{minipage}
    \hfill
    \begin{minipage}{0.22\textwidth} % 第3列标题
        \centering
        office\_35
    \end{minipage}
    \hfill
    \begin{minipage}{0.22\textwidth} % 第3列标题
        \centering
        wc\_2
    \end{minipage}
    \hfill
    \caption{The visualization of qualitative results for point cloud semantic segmentation on Area 5 of the S3DIS dataset~\cite{armeni20163d}. %We compare our PLT with previous methods on Area 5 of the S3DIS dataset~\cite{armeni20163d}.
    }
    \label{fig:s3dis_1}

\end{figure*}